\newif\ifconfver
\confvertrue        

\newif\ifonecoltab
\onecoltabtrue        

\newif\ifplainver  
\plainvertrue

\newif\ifplainver  
\plainvertrue

\ifplainver
    \confverfalse                         
\fi

\ifconfver
     \documentclass[10pt,journal]{IEEEtran}
\else
    \ifplainver
        \documentclass[10pt]{article}
\usepackage{cite,palatino}
\usepackage{fullpage}
    \else
        \documentclass[12pt,draftcls,onecolumn]{IEEEtran}
    \fi
\fi

\usepackage{calc,amsfonts,amssymb,amsmath,bm,url,color,theorem,graphicx,cite,epstopdf,nicefrac,bbold}
\usepackage{psfrag,subfigure,float}
\usepackage[algoruled,linesnumbered]{algorithm2e}
\usepackage{multirow}
\usepackage[normalem]{ulem}
\usepackage{stmaryrd}
\usepackage{xcolor}
\usepackage{psfrag,framed,mdframed}
\usepackage{comment}
\usepackage{mdframed}
\usepackage{threeparttable}

\usepackage{bibunits}

\definecolor{orange}{RGB}{255,107,0}

\newcommand{\Q}{\boldsymbol{Q}}
\newcommand{\X}{\boldsymbol{X}}

\newcommand{\A}{\boldsymbol{A}}

\renewcommand{\S}{\boldsymbol{S}}
\newcommand{\s}{\boldsymbol{s}}

\newcommand{\x}{\boldsymbol{x}}

\newcommand{\w}{\boldsymbol{w}}

\renewcommand{\a}{\boldsymbol{a}}

\newcommand{\T}{{\!\top\!}}

\DeclareMathOperator*{\minimize}{\textrm{minimize}}

\usepackage{xcolor}
\usepackage{psfrag,framed}
\usepackage{lipsum}
\usepackage{chapterbib}
\PassOptionsToPackage{normalem}{ulem}
\usepackage{ulem}
\definecolor{shadecolor}{RGB}{220,220,220}

\newtheorem{Fact}{Fact}
\newtheorem{Lemma}{Lemma}
\newtheorem{Prop}{Proposition}
\newtheorem{Theorem}{Theorem}
\newtheorem{Def}{Definition}

\theorembodyfont{\rmfamily}

\newtheorem{Assumption}{Assumption}

\begin{document}

\begin{bibunit}[IEEEtran]

\newcommand{\papertitle}{\bf
Identifiability-Guaranteed Simplex-Structured Post-Nonlinear Mixture Learning via Autoencoder
}

\newcommand{\paperabstract}{%
This work focuses on the problem of unraveling nonlinearly mixed latent components in an unsupervised manner. The latent components are assumed to reside in the probability simplex, and are transformed by an unknown post-nonlinear mixing system. This problem finds various applications in signal and data analytics, e.g., nonlinear hyperspectral unmixing, image embedding, and nonlinear clustering. Linear mixture learning problems are already ill-posed, as identifiability of the target latent components is hard to establish in general. With unknown nonlinearity involved, the problem is even more challenging. Prior work offered a function equation-based formulation for provable latent component identification. However, the identifiability conditions are somewhat stringent and unrealistic. In addition, the identifiability analysis is based on the infinite sample (i.e., population) case, while the understanding for practical finite sample cases has been elusive. Moreover, the algorithm in the prior work trades model expressiveness with computational convenience, which often hinders the learning performance. Our contribution is threefold. First, new identifiability conditions are derived under largely relaxed assumptions. Second, comprehensive sample complexity results are presented---which are the first of the kind. Third, a constrained autoencoder-based algorithmic framework is proposed for implementation, which effectively circumvents the challenges in the existing algorithm. Synthetic and real experiments corroborate our theoretical analyses.
}


\ifplainver

    \date{\today}

    \title{\papertitle}

    \author{
    Qi Lyu and Xiao Fu\\
    School of Electrical Engineering and Computer Science\\
    Oregon State University\\
    Email: (lyuqi, xiao.fu)@oregonstate.edu
    }

	\date{}

    \maketitle
    
    \begin{abstract}
    \paperabstract
    \end{abstract}

\else
    \title{\papertitle}

    \ifconfver \else {\linespread{1.1} \rm \fi

\author{Qi Lyu and Xiao Fu
	
	\thanks{

    	This work is supported in part by the National Science Foundation (NSF)-Intel MLWiNS Program under Project CNS-2003082 and NSF ECCS-1808159, and in part by the Army Research Office (ARO) under Project ARO W911NF-21-1-0227.
		
		Q. Lyu and X. Fu are with the School of Electrical Engineering and Computer Science, Oregon State University, Corvallis, OR 97331, United States. email: (lyuqi, xiao.fu)@oregonstate.edu

	}
}

\maketitle

\ifconfver \else
\begin{center} \vspace*{-2\baselineskip}
\end{center}
\fi



    \ifconfver \else \IEEEpeerreviewmaketitle} \fi

 \fi

\ifconfver \else
    \ifplainver \else
        \newpage
\fi \fi
\section{Introduction}
Unsupervised mixture learning (UML) aims at unraveling the aggregated and entangled underlying latent components from ambient data, without using any training samples. This task is also known as {\it blind source separation} (BSS) and {\it factor analysis} in the literature \cite{Common2010}. UML has a long history in the signal processing and machine learning communities; see, e.g., the early seminal work of {\it independent component analysis} (ICA) \cite{Common2010}. Many important applications can be considered as a UML problem, e.g., audio/speech separation \cite{fu2015blind}, EEG signal denoising \cite{li2009joint}, image representation learning \cite{lee1999learning}, hyperspectral unmixing \cite{Ma2013}, and topic mining \cite{fu2016robust}, just to name a few.

One of the arguably most important aspects in UML/BSS is the so-called {\it identifiability} problem---is it possible to identify the mixed latent components from the mixtures in an unsupervised manner? The UML problem is often ill-posed, since an arbitrary number of solutions exist in general; see, e.g., discussions in \cite{Common2010,fu2018nonnegative}. 
To establish identifiability, one may exploit prior knowledge of the mixing process and/or the latent components. Various frameworks were proposed for unraveling {\it linearly} mixed latent components by exploiting their properties, e.g., statistical independence, nonnegativity, boundedness, sparsity, and simplex structure---which leads to many well-known unsupervised learning models, i.e., ICA \cite{Common2010}, {\it nonnegative matrix factorization} (NMF) \cite{fu2018nonnegative}, {\it bounded component analysis} (BCA) \cite{cruces2010bca}, {\it sparse component analysis} (SCA) \cite{zibulevsky2001blind}, and {\it simplex-structured matrix factorization} (SSMF) \cite{fu2016robust,Gillis2012,fu2015blind}. These structures often stem from physical meaning of their respective engineering problems. 
For example, the simplex structure that is of interest in this work is well-motivated in applications such as topic mining, hyperspectral unmixing, community detection, crowdsourced data labeling, and image data representation learning \cite{fu2016robust,Gillis2012,huang2019detecting,zhou2011minimum,ibrahim2019crowdsourcing}.

Identifiability research of linear mixtures is relatively well established. However, unknown nonlinear distortions happen ubiquitously in practice; see examples in hyperspectral imaging, audio processing, wireless communications, and brain imaging \cite{dobigeon2014nonlinear,ziehe2000artifact,oveisi2012nonlinear}. Naturally, establishing latent component identifiability in the presence of unknown nonlinear transformations is much more challenging relative to classic linear UML cases. Early works tackled the identifiability problem from a nonlinear ICA (nICA) viewpoint. In \cite{hyvarinen1999nonlinear}, Hyv{\"a}rinen {\it et al.} showed that in general, even strong assumptions like statistical independence of the latent components are not sufficient to establish identifiability of the nonlinear mixture model. In \cite{taleb1999source,achard2005identifiability}, the structure of the nonlinear distortions were used to resolve the problem, leading to the so-called {\it post-nonlinear mixture} (PNM) ICA framework. In recent years, nICA has attracted renewed attention due to its connection to unsupervised deep learning \cite{hyvarinen2016unsupervised,hyvarinen2019nonlinear}.

Beyond ICA, other classic UML models (e.g., NMF and SSMF) have rarely been extended to the nonlinear regime. Considering latent component properties {without assuming statistical} independence is of great interest, since {this condition} among {the} latent components is not a mild assumption.
{Without resorting to statistical independence, some works assumed the nonlinear distortions fall into certain {\it known categories}, e.g., bi-linear, linear-quadratic, and polynomial distortion functions in \cite{deville2019separability,fantinato2019second,altmann2014unsupervised}, respectively.
Nonetheless, identifiability study of UML under {\it unknown} nonlinear distortions beyond nICA has been largely elusive.
}
A couple of exceptions include the nonlinear multiview analysis model in \cite{lyu2020nonlinear} and a {\it simplex-constrained post-nonlinear mixture} (SC-PNM) model in \cite{yang2020learning}. In particular, Yang {\it et al.} offered a model identification criterion and showed that the unknown nonlinear distortions can be provably removed \cite{yang2020learning}---under the assumption that the latent components are generated from the probability simplex.
Yang {\it et al.}'s work offered an approachable angle for learning the underlying components in the SC-PNM model. However, there are a number of challenges in theory and practice. First, the model identifiability condition in \cite{yang2020learning} is stringent---and some key conditions are neither enforceable nor checkable. Second, the identifiability analysis in \cite{yang2020learning} (and in all nonlinear ICA works such as those in \cite{hyvarinen2016unsupervised,achard2005identifiability,taleb1999source,hyvarinen2019nonlinear}) are based on the assumption that infinite data samples are available (i.e., the population case). Performance analysis under the finite sample case has been elusive, yet is of great interest.
Third, the implementation in \cite{yang2020learning} uses a positive neural network to model the target nonlinear transformations. Using such a special neural network is a compromise between computational convenience and model expressiveness, which loses generality and often fails to produce sensible results in practice.

\noindent
{\bf Contributions.} This work advances the understanding to the SC-PNM model learning problem in both theory and implementation. Our detailed contributions are as follows:

\noindent
$\bullet$ {\bf Deepened Identifiability Analysis.} In terms of identifiability theory, we offer a set of new sufficient conditions under which the unknown nonlinear transformations in the SC-PNM model can be provably removed. Unlike the conditions in \cite{yang2020learning}, our conditions do not involve unrealistic non-enforceable conditions. 
A number of other stringent conditions in \cite{yang2020learning}, e.g., nonnegativity of the mixing system, are also relaxed.

\noindent
$\bullet$ {\bf Finite Sample Analysis.} Beyond identifiability analysis under infinite data, we also consider performance characterization under the finite sample case.
Leveraging a link between our learning criterion and classic generalization theories, we show that the unknown nonlinear transformations can be removed up to some `residue' (measured by a certain metric) in the order of $O(N^{-\frac{1}{4}})$, where $N$ is the number of samples. To our best knowledge, this is the first finite sample performance analysis for nonlinear UML problems.

\noindent
$\bullet$ {\bf Neural Autoencoder-Based Algorithm.} We propose a carefully constructed constrained neural autoencoder for implementing the learning criterion. Unlike the implementation in \cite{yang2020learning}, our design retains the expressiveness of the neural networks when enforcing constraints in the learning criterion.
We show that finding a set of feasible points of our formulation provably removes the unknown nonlinear distortions in SC-PNM models. We offer a pragmatic and effective Lagrangian multiplier-based algorithmic framework for tackling the autoencoder learning problem. Our framework can easily incorporate popular neural network optimizers (e.g., \texttt{Adam} \cite{kingma2014adam}), and thus exhibits substantially improved efficiency relative to the Gauss-Newton method in \cite{yang2020learning}. We validate the theoretical claims and algorithm on various simulated and real datasets.

\smallskip
Part of the work was accepted by EUSIPCO 2020 \cite{lyu2020dependent}. The journal version additionally includes the detailed identifiability analysis, the newly derived finite sample analyses, 
the constrained autoencoder based implementation, the characterization of its feasible solutions, the Lagrangian multiplier algorithm, and a number of new real data experiments.

\smallskip
\noindent {\bf Notation.} We will follow the established conventions in signal processing. To be specific, $x,\bm x,\bm X$ represent a scalar, vector, and matrix, respectively; {$\text{Diag}(\bm{x})$ denotes a diagonal matrix with $\bm{x}$ as its diagonal elements}; $\|\bm{x}\|_1$, $\|\bm{x}\|_2$ and $\|\bm{x}\|_\infty$ denote the $\ell_1$ norm, the Euclidean norm and the infinity norm, respectively; $^\top$, $^\dagger$ and $^\perp$ denote the transpose, Moore-Penrose pseudo-inverse operations and {orthogonal complement}, respectively;
$\bm{P}_{\bm{X}}$ denotes the orthogonal projector onto the range space of $\bm{X}$;
$\circledast$ denotes the Hadamard product; the shorthand notation $f$ is used to denote a function $f(\cdot):\mathbb{R}\rightarrow \mathbb{R}$;
$f'$, $f''$ and $f^{(n)}$ denote the first-order,  second-order and $n$th order derivatives of the function $f$, respectively; $f\circ g$ denotes the function composition operation; $\bm{1}$ denotes an all-one vector with a proper length; 
$[K]$ denotes the integer set $\{1,2,\ldots,K\}$; 
$\sigma_{\min}(\bm{X})$ denotes the smallest singular value of matrix $\bm{X}$;
$\mathbb{E}[\cdot]$ and $\mathbb{V}[\cdot]$ denote expectation and variance of its argument, respectively;
${\rm int}{\cal X}$ means the interior of the set ${\cal X}$.

\section{Background}
In this section, we briefly introduce the pertinent background of this work.
\subsection{Simplex-Constrained Linear Mixture Model}
\begin{figure}[t!]
    \centering
    \includegraphics[width=.7\linewidth]{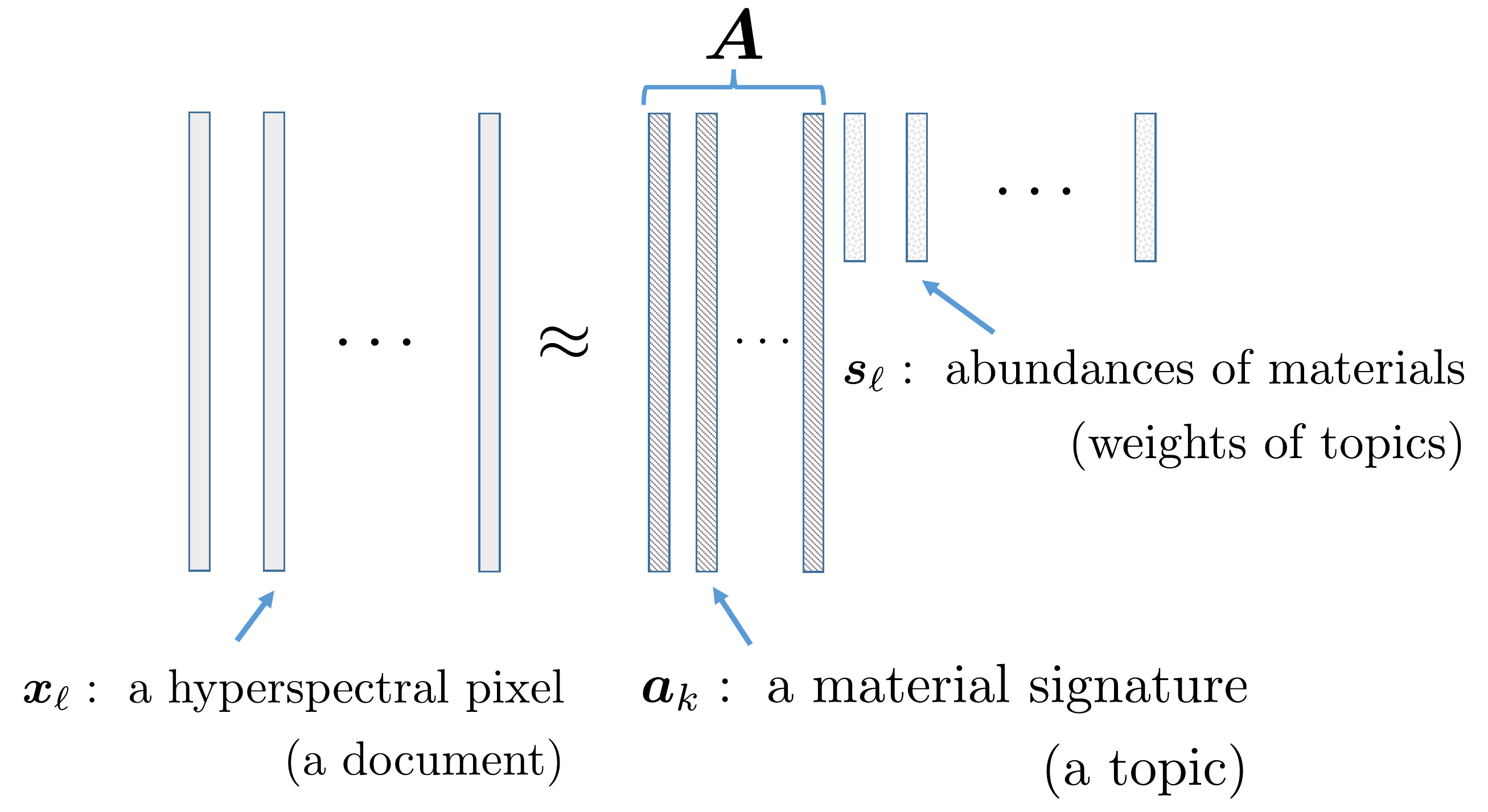}
    \caption{SC-LMM and its applications in hyperspectral imaging and topic mining; adapted from \cite{fu2016robust}.}
    \label{fig:sclmm}
    \vspace{-.25cm}
\end{figure}

The {\it simplex-constrained linear mixture model} (SC-LMM) often arises in signal and data analytics.
Consider a set of acquired signal/data samples $\x_\ell\in\mathbb{R}^M$ for $\ell=1,\ldots,N$. Under the ideal noiseless LMM, we have
\begin{align}\label{eq:LMM}
    \bm{x}_\ell = \bm{A}\bm{s}_\ell,~ \ell=1,\ldots,N,
\end{align}
where $\A\in\mathbb{R}^{M\times K}$ is referred to as the mixing system,
$\s_\ell=[s_{1,\ell},\ldots,s_{K,\ell}]^\T\in\mathbb{R}^K$ is a vector that holds the latent components $s_{1,\ell},\ldots,s_{K,\ell}$, and $\ell$ is the sample index. In SC-LMM, it is assumed that
\begin{equation}\label{eq:simplex}
    ~\s_\ell\in \bm \varDelta_K,~\bm \varDelta_K =\{\s\in\mathbb{R}^K|\bm 1^\T\s=1,\s\geq\bm 0\};
\end{equation}
i.e., the latent component vector $\bm s_\ell$ resides in the probability simplex.
In general, if $\bm x_\ell$ can be associated with different clusters (whose centroids are represented by $\bm a_k$'s) with probability or weight $s_{k,\ell}$, the SC-LMM is considered reasonable.
For example, in hyperspectral unmixing (HU) \cite{Ma2013}, $\x_\ell$ is a hyperspectral pixel, $\A=[\a_1,\ldots,\a_K]$ collects $K$ spectral signatures of materials contained in the pixel, and $s_{1,\ell},\ldots,s_{K,\ell}$ are the abundances of the materials.
Many other applications can be approximated by SC-LMM, e.g., image representation learning \cite{zhou2011minimum}, community detection \cite{huang2019detecting}, topic modeling \cite{fu2016robust}, soft data clustering \cite{fu2016robust}, just to name a few; see an illustration in Fig.~\ref{fig:sclmm}. 

LMM learning algorithms aim at recovering both $\A$ and $\s_\ell$. Note that this is usually an ill-posed problem: Given $\x_\ell$, the LMM representation is non-unique. Specifically, consider an arbitrary nonsingular $\Q\in\mathbb{R}^{K\times K}$, one can always represent $\x_\ell$ as
$\x_\ell =\A\Q\Q^{-1}\s_\ell=\widetilde{\A}\widetilde{\s}_\ell, $
where $\widetilde{\A}=\A\Q$ and $\widetilde{\s}_\ell=\Q^{-1}\s_\ell$---making the latent components not uniquely identifiable.
Establishing identifiability of $\A$ and $\s_\ell$ is often an art---different signal structures or prior knowledge has to be exploited to serve the purpose. As mentioned, ICA \cite{Common2010}, NMF \cite{fu2018nonnegative}, and SCA \cite{zibulevsky2001blind} exploit statistical independence, nonnegativity, and sparsity of the latent components in $\bm s_\ell$, respectively, to come up with identifiability-guaranteed LMM learning approaches. Identifiability of SC-LMM has also been extensively studied; see \cite{fu2018nonnegative,Ma2013}.

\subsection{Post-Nonlinear Mixture Model Learning}
Despite of the popularity, LMMs are considered over-simplified in terms of faithfully capturing the complex nature of real-world signals and data. In particular, {\it unknown} nonlinear effects are widely observed in different domains such as hyperspectral imaging \cite{dobigeon2014nonlinear} and brain signals processing \cite{oveisi2009eeg,lyu2020nonlinear}.
If not accounted for, nonlinear effects could severely degrade the performance of latent component identification.

\begin{figure}[t!]
    \centering
	\subfigure{\includegraphics[page=1, clip, trim=5.5cm 4cm 5.5cm 4cm, width=0.7\textwidth]{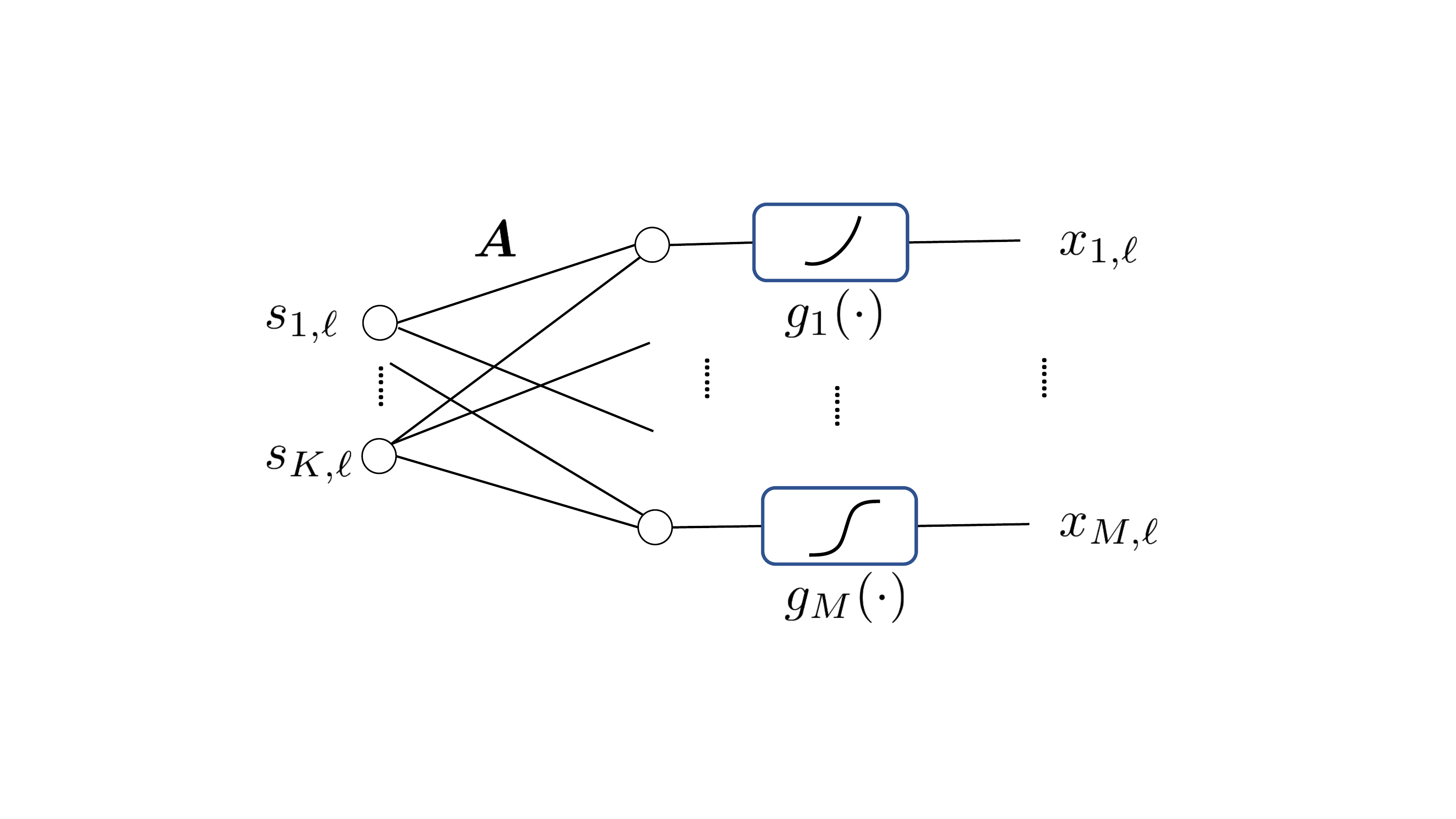}}
    \caption{Illustration of PNM, where $\A$, $\s_\ell$ and $g_m(\cdot)$ are all unknown.}\label{fig:generative}
    \vspace{-.25cm}
\end{figure}

To take the nonlinear distortions into consideration, one way is to employ the so-called {\it post-nonlinear mixture} (PNM) model \cite{taleb1999source,achard2005identifiability,oja1997nonlinear,ziehe2003blind,lyu2020nonlinear}. Under PNM, the data model is expressed as follows:
\begin{align}\label{eq:PNM}
x_{m,\ell} = g_m\left(  \sum_{k=1}^K\bm a_k s_{k,\ell} \right),\quad   \ell=1,2,\ldots,N,
\end{align}
where $g_m(\cdot):\mathbb{R} \rightarrow \mathbb{R}$ is a scalar-to-scalar unknown nonlinear continuous invertible function; see Fig.~\ref{fig:generative}. The model can also be expressed as 
$\bm x_\ell=\bm g(\A\s_\ell)$, where $\bm{g}(\cdot)=[g_1(\cdot),\ldots,g_M(\cdot)]^\T$.

The PNM model is a natural extension of LMM, and has a wide range of applications, e.g., low power/microwave wireless communications \cite{larson1998radio}, chemical sensor design \cite{bermejo2006isfet} and integrated Hall sensor arrays \cite{paraschiv2002source}. {In principle, if the data acquisition process is believed to have unknown nonlinear distortions on the sensor end (represented by $g_m(\cdot)$), such a modeling is considered appropriate. In addition, for data analytics problems that have relied on the LMM based representation learning (e.g., image embedding \cite{lee1999learning}), using PNM may improve the generality of the model, thereby offering benefits in terms of downstream tasks' performance \cite{lyu2020nonlinear}.}

Identifiability of the latent components under PNM has been studied for a number of special cases.
It was shown in \cite{taleb1999source,achard2005identifiability} that if $\s_{k,\ell}$ for $k=1,\ldots,K$ are statistically independent, an nICA framework can provably extract $\bm s_\ell$ up to certain ambiguities. In fact, the vast majority of PNM learning works were developed under the nICA framework; see, e.g., \cite{oja1997nonlinear,ziehe2003blind,taleb1999source,achard2005identifiability}. We should mention that there are other nonlinear mixture models in the literature beyond PNM; see \cite{hyvarinen2019nonlinear,gresele2020incomplete}, which are also developed as extensions of ICA.

The statistical independence assumption is considered stringent in many applications (e.g., in hyperspectral unmixing \cite{nascimento2005does} or any application where the simplex constraint in \eqref{eq:simplex} holds).
New PNM learning frameworks have been proposed to circumvent this assumption. The work in \cite{lyu2020nonlinear} uses multiple `views' of the data entities (e.g., audio and image representation of the same entity `cat') to remove unknown nonlinear distortions without using statistical independence.
When only one view of the data is available, the work in \cite{yang2020learning} studies the model identifiability of the PNM under the simplex constraint, i.e.,
\begin{equation}\label{eq:SCLMM}
     \x_\ell = \bm{g}(\bm{A}\bm{s}_\ell),~\s_\ell \in \bm \varDelta_K,~\forall \ell\in [N],  
\end{equation}
which will be referred to as the SC-PNM model.
The SC-PNM model is well-motivated, since unknown nonlinear distortions are often observed in applications such as hyperspectral imaging and image/text analysis \cite{dobigeon2014nonlinear}, where the constraint \eqref{eq:simplex} on the latent components is believed to be appropriate. 
Learning the SC-PNM model is our interest in this work.

\subsection{Prior Work in \cite{yang2020learning}}
To learn the latent components from SC-PNM, Yang {\it et al.} proposed to learn a nonlinear function $\bm{f}:\mathbb{R}^{M}\rightarrow\mathbb{R}^M$ such that it can `cancel' or inversely identify the distortion function $\bm{g}(\cdot)=[g_1(\cdot),\ldots,g_M(\cdot)]^\T$ using the following criterion \cite{yang2020learning}:
\begin{mdframed}
\begin{subequations}\label{eq:population}
\begin{align}
    \text{find}&~\bm{f}=[f_1,\ldots,f_M]^\T,\\
({\sf P})\quad\quad    \text{subject~to}&~\bm{1}^\top \bm{f}(\bm{x})=1,\ \forall \x\in{\cal X}\label{eq:sumto1_constraint}\\
    &~f_m:\mathbb{R}\rightarrow \mathbb{R}:~{\rm invertible} \label{eq:invert_constraint}
\end{align}
\end{subequations}
\end{mdframed}
where $f_m$ is a scalar-to-scalar nonlinear function and
\[  {\cal X}=\{ \x\in\mathbb{R}^M|\x = \bm g(\A\s),~\forall \s\in\mathbb{R}^K,~\s\in\bm \varDelta_K \} \]
is the domain where all data samples are generated from. 
In the above, the subscript `$\ell$' that is used as the sample index is eliminated since the criterion assumes the continuous domain ${\cal X}$ is available---and the constraint \eqref{eq:sumto1_constraint} is enforced over the entire domain.
In \cite[Theorem 2]{yang2020learning}, it was shown that when $M>K$, there always exists at least one feasible solution of \eqref{eq:population}.
Here, we first show that feasible solutions also exist for the $M=K$ case under some mild conditions:
\begin{Lemma}[Feasibility]\label{lem:feas}
    Assume that $M\geq K$ and that $\A$ is drawn from any joint absolutely continuous distribution. Then, almost surely, there exists at least a feasible solution of \eqref{eq:population} under the generative model in \eqref{eq:SCLMM}.
\end{Lemma}
\noindent
{\bf Proof:}
Let us first consider the $M=K$ case.
By construction, let $\bm f=\bm D\bm g^{-1}$ where $\bm D={\rm Diag}(\bm \tau)$ such that $\bm A^\T\bm \tau=\bm 1$. 
Such an $\bm f$ satisfies \eqref{eq:sumto1_constraint}.
Furthermore, if $\bm \tau$ does not have zero elements,
then $\bm f$ is invertible. Our proof boils down to showing that $\bm A^\T\bm \tau = \bm 1$ admits a dense solution. Note that for square random matrix $\bm{A}$, we have $\bm{\tau}=\bm{A}^{-\top}\bm{1}$. 

Since $\A$ is continuously distributed, and matrix inversion is an invertible continuous operator, $\A^{-\T}$ also follows a joint absolutely continuous distribution.
Let us denote $\bm p_k^\T$ be the $k$th row of $\A^{-\T}$. Then, ${\sf Prob}\{ \bm p_k^\T\bm 1=0 \}=0$ since $\bm p_k^\T$ follows a joint absolutely continuous distribution. Consequently, $\bm{\tau}=\bm{A}^{-\top}\bm{1}$ is a dense vector with probability one.

For the $M>K$ case, we show that a random $\bm A$ satisfies the `incoherent' condition given in \cite[Proposition 1]{yang2020learning}. To be specific, one can see that $\bm e_m\in{\sf range}(\A)$ with probability zero, if $\A$ follows any joint absolutely continuous distribution. Hence, $\bm A^\T\bm \tau=\bm 1$ holds with a dense $\bm \tau$ almost surely by invoking \cite[Proposition 1]{yang2020learning}. \hfill $\square$

\bigskip

Lemma~\ref{lem:feas} asserts that when $M\geq K$ (other than $M>K$ in \cite{yang2020learning}), at least one feasible solution of \eqref{eq:population} exists.
The reason why we could derive a (slightly) better lower bound of $M$ is because
the feasibility analysis in \cite{yang2020learning} is based on a worst-case argument, while we consider the generic case.
\color{black}

In \cite{yang2020learning}, the following theorem was shown:
\begin{Theorem}\cite{yang2020learning}\label{thm:thm_of_yang}
	Consider the SC-PNM model in \eqref{eq:SCLMM}. 
	Assume 1) that $\A\geq \bm 0$ is tall (i.e., $M>K$), full rank, and incoherent (see definition in \cite{yang2020learning}), 2) and that the solution of \eqref{eq:population} (denoted by $\widehat{f}_m$ for $m\in[M]$) makes the composition $\widehat{h}_m=\widehat{f}_m\circ g_m$ convex or concave for $m\in[M]$. Then, $\widehat{h}_m=\widehat{f}_m\circ g_m$ has to be an affine function; i.e., any $f_m$ that is a solution of \eqref{eq:population} satisfies
	$    \widehat{\bm h}_m(y) = \widehat{f}_m \circ g_m(y) =c_m  y + d_m,~m\in[M]$,
where $c_m\neq 0$ and $d_m$ are constants. {In addition, if $\sum_{m=1}^M d_m \neq 1$, we have
$$ \widehat{\bm h}(\A\s) = \widehat{\bm f} \circ \bm g(\A\s) =\widehat{\bm A}\s,$$
where $\widehat{\bm A}= \bm D\bm A$ and $\bm D$ is a full-rank diagonal matrix;
i.e., $\widehat{\bm h}(\A\s)$ is a linear function of $\A\s$ for $\s\in{\rm int}\bm \varDelta_K$.}
\end{Theorem}
Theorem~\ref{thm:thm_of_yang} means that if \eqref{eq:sumto1_constraint} can be solved with Conditions 1)-2) satisfied, the unknown nonlinear transformation $\bm g$ can be removed from $\x=\bm g(\A\bm s)$ and $\bm f(\bm x)=\widehat{\A}\s$, where $\widehat{\A}$ is a row-scaled version of $\A$.
Hence, $\bm y=\bm f(\bm x)$ becomes SC-LMM---and identifying the latent components from SC-LMM has been well-studied (see, e.g., \cite{fu2018nonnegative,MVES,fu2016robust,Ma2013}), as mentioned in previous sections.

\subsection{Remaining Challenges}
Theorem~\ref{thm:thm_of_yang} has offered certain theoretical justifications for the nonlinear model learning criterion in \eqref{eq:population}. However, an array of challenges remain.

\subsubsection{Stringent Conditions} Theorem~\ref{thm:thm_of_yang} was derived under rather restrictive conditions. Condition 1) means that $\A$ has to be nonnegative, which may not be satisfied in many applications, e.g., blind channel equalization in wireless communications. Condition 2) is perhaps even more challenging to satisfy or check. Since $\widehat{f}_m$ is a function to be learned and $g_m$ is unknown, the convexity/concavity of the composition $\widehat{h}_m=\widehat{f}_m\circ g_m$ is neither in control of the designers/learners nor enforceable by designing regularization/constraints.

\subsubsection{Lack of Understanding For Finite Sample Cases} The proof in Theorem~\ref{thm:thm_of_yang} is based on the assumption that the constraints in \eqref{eq:population} hold over the entire ${\cal X}$. Note that ${\cal X}$ is a continuous domain, which means that an uncountably infinite number of $\bm x$'s exist in the domain. In practice, only a finite number of samples $\x_\ell\in{\cal X}$ are available.  
It is unclear how the learning performance scales with the number of samples.
In general, finite sample analysis for nonlinear mixture learning is of great interest, but has not been addressed.

\subsubsection{Implementation Issues} The implementation of \eqref{eq:population} in \cite{yang2020learning} was based on parameterizing $f_m$'s using positive neural networks (NNs), i.e., NNs that only admit positive network weights. This design is driven by the fact that NNs are universal function approximators. Hence, it is natural to use NNs for representing the target unknown nonlinear functions. The positivity is added to enforce the invertibility of the learned $f_m$'s [cf. \eqref{eq:invert_constraint}]. 
However, adding positivity constraints to the network weights may hinder the universal approximation capacity---which often makes the performance unsatisfactory, as one will see in the experiments.
Optimization involving positivity-constrained NN is also not trivial. The work in \cite{yang2020learning} employed a Gauss-Newton based method with dog-leg trust regions, which is not scalable.

\section{Nonlinearity Removal: A Deeper Look}
In this work, we start with offering a new model identification theorem under the learning criterion in \eqref{eq:population}, without using the stringent conditions in Theorem~\ref{thm:thm_of_yang}. Then, we will characterize the criterion under finite sample cases. We will also propose a primal-dual algorithmic framework that effectively circumvents the challenges in the positive NN-based implementation in \cite{yang2020learning}.

We begin with showing that the stringent conditions in 1) and 2) of Theorem~\ref{thm:thm_of_yang} can be relaxed.
To proceed, we first show the following lemma:
\begin{Lemma}\label{lem:lemma_uu}
Assume $K\geq 3$.
Consider $\bm{s} =[s_1,\ldots,s_K]^\T \in {\rm int }\bm \varDelta_K$. Then, $\frac{\partial s_i}{\partial s_j}=0$ for $i\neq j$ where $i,j=1,\ldots,K-1$.
\end{Lemma}

\noindent
{\bf Proof:}
First, for $\bm{s}_\ell\in {\rm int }\bm \varDelta_K$, we only have $K-1$ free variables, i.e., without loss of generality, $s_i$ for $i=1,\ldots, K-1$. Assume that $i,j\in[K-1]$ and $i\neq j$. For any fixed $\bar{s}_i$, $s_j$ can take any possible values within a nonempty continuous domain (e.g., if $s_i=0.5$ then the domain of $s_j$ is $(0,0.5)$ regardless of other components). 
Hence, if one treats $s_i$ as a function of $s_j$, $\frac{\partial s_i}{\partial s_j}=0$ always holds within $\s\in{\rm int}\bm \varDelta_K$. \hfill $\square$

\bigskip

Equipped with Lemma~\ref{lem:feas} and Lemma~\ref{lem:lemma_uu}, we show our first main theorem:
\begin{Theorem}[Nonlinearity Removal]\label{thm:main_population}
	Under the model in \eqref{eq:SCLMM}, assume that the criterion \eqref{eq:population} is solved.
	In addition, assume that $\bm{A}\in\mathbb{R}^{M\times K}$ is drawn from any joint absolutely continuous distribution,
	{and that the learned $\widehat{h}_m=\widehat{f}_m\circ g_m$ is twice differentiable for all $m\in[M]$.}
	Suppose that 
	\begin{equation}\label{eq:Mbound}
	    3\leq K\leq	M\leq  \frac{K(K-1)}{2}.
	\end{equation}
	Then, almost surely, any $\widehat{f}_m$ that is a solution of \eqref{eq:population} satisfies
    $\widehat{h}_m(y) = \widehat{f}_m \circ g_m(y) =c_m  y + d_m$, $\forall m\in[M],$
	where $c_m\neq 0$ and $d_m$ are constants. {Furthermore, if $\sum_{m=1}^M d_m \neq 1$, then, almost surely, we have
$$ \widehat{\bm h}(\A\s) = \widehat{\bm f} \circ \bm g(\A\s) =\widehat{\bm A}\s,$$
where $\widehat{\bm A}= \bm D\bm A$ and $\bm D$ is a full-rank diagonal matrix;
i.e., $\widehat{\bm h}(\A\s)$ is a linear function of $\A\s$ for $\s\in{\rm int}\bm \varDelta_K$. }
\end{Theorem}

\noindent
{\bf Proof:}
    Consider $\bm{s}\in{\rm int } \bm \varDelta_K$. With Lemma \ref{lem:lemma_uu}, we have $\frac{\partial s_i}{\partial s_j}=0$ where $i,j=1,\ldots,K-1$ and $i\neq j$. 
Hence, by solving problem ~\eqref{eq:population}, we have the following equation:
\begin{align*}
    \sum_{i=1}^M \widehat{h}_i\left(a_{i,1}s_1+a_{i,2}s_2+\ldots+a_{i,K}\left(1-\sum_{j=1}^{K-1}s_j\right)\right) = 1,
\end{align*}
where we have used $s_K=1-s_1-\ldots-s_{K-1}$. By taking second-order derivatives of both sides w.r.t $s_i$ and $s_j$ for $i,j\in[K-1]$, we have
\begin{align}\label{eq:linear_system}
\bm G\bm{h}''=
\underbrace{\begin{bmatrix}
    \left(\bm{b}_{1}\circledast \bm b_1\right)^\top\\
    \ldots\\
    \left(\bm{b}_{K-1} \circledast \bm b_{K-1}\right)^\top\\
    \left(\bm{b}_{1}\circledast \bm{b}_{2}\right)^\top\\
    \ldots\\
    \left(\bm{b}_{K-2}\circledast\bm{b}_{K-1}\right)^\top\\
\end{bmatrix}}_{{\bm G}}
\begin{bmatrix}
\widehat{h}_1''\\
\vdots\\
\widehat{h}_{M}''
\end{bmatrix}=\bm{0},
\end{align}
where $\bm{b}_i=[a_{1,i}-a_{1,K},\ a_{2,i}-a_{2,K},\ \ldots,\ a_{M,i}-a_{M,K}]^\top$ where $i=1,\ldots,K-1$ and ${\bm B}=[\bm{b}_1,\ldots,\bm{b}_{K-1}]$. Note that $\bm G$ has a size of $K(K-1)/2\times M$.

We hope to show that ${\rm rank}(\bm G)=M$. To achieve this goal, we show that there are $M$ rows of $\bm G$ that are linearly independent.
This can be shown by showing that there exists a particular case such that an $M\times M$ submatrix of $\bm G$ has full column rank. The reason is that the determinant of any $M\times M$ submatrix of $\bm G$ is a polynomial of $\bm A$, and a polynomial is nonzero almost everywhere if it is nonzero somewhere \cite{caron2005zero}. 

To this end, consider a special case where ${\bm B}$ is a Vandermonde matrix, i.e.,
$\bm{b}_i = [1, z_i, z_i^2, \ldots, z_i^{M-1}]^\top,$
and $z_i\neq z_j$.
{Such a $\bm B$ can always be constructed by letting $\bm A^\T$'s first $K-1$ rows to be a Vandermonde matrix and the last row to be all zeros.}
By picking $\widetilde{K}$ columns from the $\bm{B}$ matrix, where we require $\widetilde{K}\leq K-1$ to satisfy $M\leq \frac{\widetilde{K}(\widetilde{K}+1)}{2}$ (for simplicity, we take $M=\frac{\widetilde{K}(\widetilde{K}+1)}{2}$ for the rest of proof). 
Hence, the corresponding rows in the matrix of interest have the following form:
\begin{align}\label{eq:zpoly}
    \begin{bmatrix}
        1 & z_{1}^2  & \ldots & z_{1}^{2(M-1)}\\
        \ldots&\ldots&&\ldots \\
        1 & z_{\widetilde{K}}^2 & \ldots & z_{\widetilde{K}}^{2(M-1)}\\
        1 & z_{1}z_{2} & \ldots & (z_{1}z_{2})^{M-1}\\
        \ldots&\ldots&&\ldots\\
        1 & z_{\widetilde{K}-1}z_{\widetilde{K}} & \ldots & (z_{\widetilde{K}-1}z_{\widetilde{K}})^{M-1} 
    \end{bmatrix}
\end{align}

Note that one can always construct such a sequence---e.g., $z_{1}=1, z_{2}=1.1, z_{3}=1.11, \ldots$ such that the matrix in \eqref{eq:zpoly} is full rank. 
This means that the linear combination of this second order homogeneous polynomials is not identically zero, which implies that it is non-zero almost everywhere \cite{caron2005zero}. Hence, the matrix ${\bm G}$ has full column rank almost surely which further means that $\widehat{h}_m''=0$ for all $m\in[M]$. 

{Note that $\widehat{h}_m''=0$ indicates that $h_m$ is affine. Since $f_m$ and $g_m$ for all $m\in[M]$ are both invertible, $\widehat{h}_m(y)=c_m y+d_m$ with $c_m\neq 0$---otherwise, $h_m$ is not invertible. Then, following arguments in \cite[Remark 1]{yang2020learning}, one can show that $\bm h(\A\s)$ is linear in $\bm A\bm s$ if $\sum_{m=1}^Md_m\neq 1$.} \hfill $\square$

\bigskip

Comparing Theorems~\ref{thm:thm_of_yang} and \ref{thm:main_population}, one can see that the conditions in 1) and 2) in Theorem~\ref{thm:thm_of_yang} are no longer needed. Relaxing the nonnegativity of $\A$ makes the method applicable to many more problems where the mixing system can have negative entries (e.g., speech separation and wireless communications). Removing Condition 2) is even more important, since this condition is not checkable or enforceable. 
{Specifically, instead of asking for $\widehat{h}_m=\widehat{f}_m\circ g_m$ to be all convex or all concave as in Theorem~\ref{thm:thm_of_yang}, the conditions in Theorem~\ref{thm:main_population} only need $\widehat{h}_m=\widehat{f}_m\circ g_m$ to be twice differentiable. 
This may be {\it enforced} through construction and regularization.
For example, if one uses a neural network to represent the learning function $\widehat{f}_m$ (as we will do in this work), the differentiability can be promoted via bounding the network weights and using differentiable activation functions.}

Notably, Theorem~\ref{thm:main_population} holds if $M\leq \frac{K(K-1)}{2}$ [cf. Eq.~\eqref{eq:Mbound}]. This, at first glance, seems uncommon. 
By the `conventional wisdom' from LMM, having more `channels' (i.e., larger $M$) means more degrees of freedom, and normally better performance. 
However, in PNM learning under the criterion in \eqref{eq:population}, having more output channels is not necessarily helpful, since more channels means that more unknown nonlinear functions $g_m(\cdot)$'s need to be learned; more precisely, more $h_m=f_m\circ g_m$'s that need to be tied to the solution of a linear system [see Eq.~\eqref{eq:linear_system}]---and this increases the difficulty, which is a sharp contrast to the LMM case.     
Formally, we use the following theorem to show that $M\leq \frac{K(K-1)}{2}$ is necessary under the learning criterion \eqref{eq:population}:
\begin{Theorem}\label{thm:necessary}
    Under the same model of Theorem~\ref{thm:main_population}, assume that $M> \frac{K(K-1)}{2}.$ Then, there exist solutions $f_m$'s that satisfy the constraints in \eqref{eq:population} but $h_m=f_m\circ g_m$'s are not affine.
\end{Theorem}

Theorem~\ref{thm:necessary} puts our discussions into context: In SC-PNM learning, one should avoid directly applying the criterion in \eqref{eq:population} onto the cases where $M\gg K$. Nonetheless, this issue is more of an artifact of the criterion used in \eqref{eq:population}, and
can be easily fixed by modifying the criterion. For the $M>K$ cases, one may change the constraints in \eqref{eq:population} to be segment by segment:
\begin{equation}\label{eq:selectchannel}
    \bm 1^\T\bm f\left([\x]_{(p-1)K+1:pK}\right)=1,~p\in[M/K],
\end{equation}
if $M/K$ is an integer. If $M/K$ is not an integer, then overlap between the segments can be used.

\section{Sample Complexity Analysis}
The last section was developed under the assumption that the function equation $\bm 1^\T\bm f(\bm x)=1$ holds over the continuous domain ${\cal X}$ [cf. Eq.~\eqref{eq:population}].
In this section, we turn our attention to more practical settings where only finite samples are available. Consider the finite sample version of \eqref{eq:population}:
\begin{mdframed}
\begin{subequations}\label{eq:sample}
    \begin{align}
    \text{find}&~\bm{f}=[f_1,\ldots,f_M]^\T,\\
    (\widehat{\sf P})~~ \text{subject~to}&~\bm{1}^\top \bm{f}(\bm{x}_\ell)=1,~\forall\ell\in[N] \label{eq:sample_eq}\\
    &~f_m:\mathbb{R}\rightarrow \mathbb{R}:~{\rm invertible},~f_m\in{\cal F}, \nonumber
    \end{align}
\end{subequations}
\end{mdframed}
where ${\cal F}$ is the function class where every $f_m$ is chosen from. For example, ${\cal F}$ may represent classes such as polynomial functions, kernels, and neural networks.
In practice, the criterion in \eqref{eq:population} can only be carried out via tackling criterion~\eqref{eq:sample}. Hence, understanding key characterizations of \eqref{eq:sample}, e.g., the trade-off between sample complexity and the effectiveness of nonlinearity removal, is critical.

\subsection{Finite Function Class}
We first show the case where ${\cal F}$ contains a finite number of atoms, i.e., the finite function class. To this end, we begin with the following definition and assumptions:
{\begin{Def}\label{def:g_inverse}
    Assume $g\in{\cal G}$ is an invertible continuous nonlinear function. We define its inverse class ${\cal G}^{-1}$ as a function class that contains all the $u$'s satisfying
    $u\circ g(y) =  y$, $\forall g\in{\cal G}$.
\end{Def}}

\begin{Assumption}[Realizability]\label{as:exact_finite}
Assume that $g_m\in{\cal G}$, $f_m\in {\cal F}$, and ${\cal G}^{-1}\subseteq {\cal F}$.
\end{Assumption}
\begin{Assumption}[Finite Class]\label{as:finite}
Assume that $|{\cal F}|\leq d_{\cal F}$ {where $d_{\cal F}$ is an upper bound of the cardinality of ${\cal F}$.}
\end{Assumption}

\begin{Assumption}[Boundedness]\label{as:bound}
    Assume that the fourth-order derivatives of ${f}_m\in{\cal F}$ and $g_m\in{\cal G}$ exist. In addition, $|f_m^{(n)}(\cdot)|$ and $|g_m^{(n)}(\cdot)|$ are bounded for all $n\in[4]$ and $m\in[M]$. Also assume that $\A$ has bounded elements. 
\end{Assumption}
Assumption~\ref{as:exact_finite} corresponds to the `realizable case' in statistical learning theory---i.e., a solution does exists in the function class used for learning.
Assumption~\ref{as:finite}
formalizes the finite function class postulate. Assumption~\ref{as:bound} restricts our analysis to the function class ${\cal F}$ and the unknown nonlinear distortions that admit bounded $n$th-order derivatives (with $n$ up to 4), as a regularity condition.

Using the above assumptions, we show the following: 
\begin{Theorem}[Finite Class Sample Complexity]\label{thm:approx_err_finite_class}
    Under the generative model \eqref{eq:SCLMM}, assume that Assumptions~\ref{as:exact_finite}-\ref{as:bound} hold, 
    {that $\bm x_\ell$ for $\ell\in[N]$ are i.i.d. samples from ${\cal X}$ according to a certain distribution ${\cal D}$,}
    and that the criterion in \eqref{eq:sample} is solved. Denote any solution of \eqref{eq:sample} as $\widehat{\bm f}=[\widehat{f}_1,\ldots,\widehat{f}_M]^\T$ and $\widehat{h}_m=\widehat{f}_m\circ g_m$ for $m\in[M]$.
    Then, with probability of at least $1-\delta$, {if $$\gamma=\Omega((\nicefrac{C^2\log(2d_{\mathcal{F}}/\delta)}{N C_\phi^4})^{1/16}),$$} we have
        \begin{equation}
        \mathbb{E}\left[\left\|\widehat{\bm h}''(\bm A\s)\right\|^2_2\right]=
        O \left(\frac{\left(\log\left(\frac{2d_{\cal F}}{\delta}\right)\right)^{1/4}}{N^{1/4}\sigma_{\min}^2({\bm G})}\right), 
    \end{equation}
    {for any $\bm s\in\bm \varDelta_K$ such that $1-\gamma\geq s_{i}\geq \gamma$ for all $i\in[K]$,}
    where ${\bm G}$ is a function of the mixing system $\bm A$ that is defined in Theorem \ref{thm:main_population} [cf. Eq.~\eqref{eq:linear_system}] and $\widehat{\bm h}''=[\widehat{h}_1'',\ldots,\widehat{h}_M'']^\T$, {with $C$ and $C_\phi$ defined in Lemmas \ref{lem:hoeffding} and \ref{lemma:bounded_4th}, respectively}.
\end{Theorem}

The parameter $\gamma$ confines the applicability to the interior of $\bm \varDelta_K$---when $N$ grows, $\gamma$ approaches zero, making the applicability of the theorem gradually cover the entire ${\rm int}\bm \varDelta_K$.
Note that the performance metric here is the `size' of $\widehat{\bm h}''$, since $\widehat{\bm h}''=\bm 0$ means that the learned composition $\widehat{h}_m=\widehat{f}_m\circ g_m$ is affine, and thus $\|\widehat{\bm h}''\|_2^2\approx 0$ indicates that an approximate solution has been attained.

Theorem~\ref{thm:approx_err_finite_class} uses the assumption that ${\cal F}$ is finite. 
Considering finite classes is meaningful, as continuous functions $f_m:\mathbb{R}\rightarrow\mathbb{R}$ are always approximated by (high-precision) discrete representations in digital systems.
For example, if the function values of $f_m$ are represented by a 64-bit double-precision floating-point format consisting of $q$ real-valued parameters, then the number of $f_m(\cdot)$'s is from a finite class ${\cal F}$ consisting of at most ${2^{64q}}$ functions, which is not unrealistic considering the $\log$ operation is involved.

\subsection{Neural Networks}
One may also be interested in analyzing the sample complexity if specific function classes (in particular, neural networks) are used. In this subsection, we make the following assumption on ${\cal F}$:

\begin{Assumption}[Neural Network Structure]\label{as:nn}
    Assume that $f_m$ is parameterized by a two-layer neural network with $R$ neurons and $z$-Lipschitz nonlinear activation function $\zeta(\cdot):\mathbb{R}\rightarrow \mathbb{R}$ {with $\zeta(0)=0$}\footnote{Many popular activation functions used in modern neural networks satisfy $\zeta(0)=0$, e.g., tanh, rectified linear unit (ReLU), centered sigmoid, Gaussian error linear unit (GELU), and exponential linear unit (ELU) \cite{wan2013regularization}.}, i.e.,
   \begin{equation}\label{eq:NNclass}
       {\cal F}=\{f_m|f_m(x)=\bm w_2^\T\bm \zeta(\bm w_1 x),\|\bm w_i\|_2\leq B,~i=1,2\}, 
   \end{equation} 
    where $\bm \zeta (\bm y)=[\zeta(y_1),\ldots,\zeta(y_R)]^\T$, $\bm w_i\in\mathbb{R}^{R}$ for $i=1,2$.
    
\end{Assumption}
Given these assumptions, we first show the following theorem:

\begin{Theorem}[Neural Network Sample Complexity]\label{thm:approx_err_nn_class}
    {Under the generative model \eqref{eq:SCLMM},
    assume that Assumptions~\ref{as:exact_finite}, \ref{as:bound} and \ref{as:nn} hold.
    Suppose {that $\bm x_\ell$ for $\ell\in[N]$ are i.i.d. samples from ${\cal X}$ according to a certain distribution ${\cal D}$.} 
    Denote any solution of \eqref{eq:sample} as $\widehat{\bm f}=[\widehat{f}_1,\ldots,\widehat{f}_M]^\T$ and $\widehat{h}_m=\widehat{f}_m\circ g_m$ for $m\in[M]$. Then, with probability of at least $1-\delta$, {if $\gamma=\Omega((\nicefrac{MzB^2C_x}{C_\phi})^{1/4}(\sqrt{\nicefrac{R}{N}}+\sqrt{\nicefrac{\log(1/\delta)}{N}})^{1/8})$}, it holds that
    \begin{equation}
        \mathbb{E}\left[\left\|\widehat{\bm h}''(\A\bm s)\right\|_2^2\right]= O\left( \frac{M\left(\sqrt{R}+\sqrt{\log(1/\delta)}\right)^{1/2}}{\sigma_{\min}^2({\bm G}) N^{1/4}}  \right) .
    \end{equation}}
    for any $\bm s\in\bm \varDelta_K$ such that $1-\gamma\geq s_{i}\geq \gamma$ for all $i\in[K]$, {with $C_\phi$ and $C_x$ defined in Lemmas \ref{lemma:bounded_4th} and \ref{lem:rc_nn}, respectively}.
\end{Theorem}
Note that by increasing $R$, the employed neural networks are more expressive, and thus Assumption~\ref{as:nn} may have a better chance to be satisfied. On the other hand, increasing $R$ makes the sample complexity increases. This balance is more articulated in the next subsection.

\subsection{In The Presence of Function Mismatches}
For both Theorem~\ref{thm:approx_err_finite_class} and Theorem~\ref{thm:approx_err_nn_class}, a key assumption is that ${\cal G}^{-1}\subset {\cal F}$ for all $m$ (i.e., Assumption~\ref{as:exact_finite}). In practice, since ${\cal F}$ is picked by the learner, and thus it is likely $g_m^{-1}\notin{\cal F}$. 
To address this case, we show the following theorem:

\begin{Theorem}\label{thm:function_mismatch}
   Assume that for any $f\in{\cal F}$, there exists $u\in{\cal G}^{-1}$ such that
    \begin{equation*}
        \sup_{\x\in{\cal X}}~|f(x_m)-u(x_m)|<\nu,~\forall m\in[M].
    \end{equation*}
    Then, under the same generative model of $\x$ as in Theorems~\ref{thm:approx_err_finite_class}-\ref{thm:approx_err_nn_class}, with probability of at least $1-\delta$, there exists an $\widehat{f}\in{\cal F}$ that violates \eqref{eq:sample_eq} up to $O(M\nu)$ on average. In addition, the following holds for any $\bm s\in\bm \varDelta_K$ such that $1-\gamma\geq s_{i}\geq \gamma$ for all $i\in[K]$:
    
   \noindent
   1) if $|{\cal F}|\leq d_{\cal F}$ {and $\gamma=\Omega((\nicefrac{C^2\log(2d_{\mathcal{F}}/\delta)}{N C_\phi^4})^{1/16})$}, then 
    \begin{align*}
        \mathbb{E}\left[\left\|\widehat{\bm h}''(\A\s)\right\|_2^2\right]&=O\left(\frac{1}{\sigma_{\min}^2({\bm G})}  \left(\frac{\log\left(\frac{2d_{\cal F}}{\delta}\right)}{N}+M^2\nu^2\right)^{1/4} \right);
    \end{align*}
    2) moreover, if ${\cal F}$ is the neural network class from Assumption~\ref{as:nn} {and $\gamma=\Omega((\nicefrac{MzB^2C_x}{C_\phi})^{1/4}(\sqrt{\nicefrac{R}{N}}+\sqrt{\nicefrac{\log(1/\delta)}{N}})^{1/8})$}, then
    \begin{align*}
    \mathbb{E}\left[\left\|\widehat{\bm{h}}''(\A\s)\right\|_2^2\right]
    = O\left( \frac{{M}\left(\sqrt{R}+\sqrt{\log(1/\delta)}\right)^{1/2}}{\sigma_{\min}^2({\bm G}) N^{1/4}}+\frac{{M\nu}}{\sigma_{\min}^2({\bm G})}  \right).
       \end{align*}
\end{Theorem}
\color{black}

The proof is relegated to Appendix~\ref{appdx:function_mismatch}.
Theorem~\ref{thm:function_mismatch} reveals a trade-off between the expressiveness/complexity of the picked function class ${\cal F}$ and the nonlinearity removal `residue' induced by the function mismatch error $\nu$. For example, for the neural network case, we can always decrease the error $\nu$ by increasing the width or depth of the employed neural network. However, this inevitably increases the Rademacher complexity {\cite{bartlett2002rademacher}} of ${\cal F}$. In practice, these two aspects need to be carefully balanced. {In addition, note that function mismatch could be a result of the presence of noise. Hence, Theorem~\ref{thm:function_mismatch} also sheds some light on the noise robustness of the proposed approach.}

\smallskip

{
We hope to remark that the conditions in Theorems~\ref{thm:approx_err_finite_class}-\ref{thm:function_mismatch} are often unknown {\it a priori}. This means that it may not be straightforward to leverage these results for practical purposes, e.g., neural network structure design and hyperparameter selection. 
Nonetheless, Theorems~\ref{thm:approx_err_finite_class}-\ref{thm:function_mismatch} are the first to explain the reason why post-nonlinear mixture model learning is viable even if only finite samples are available. 
This set of results also helps understand how the learning performance scales with some key problem parameters, e.g., the sample size $N$ and the complexity of ${\cal G}$ that contains $\bm g$.
}

\color{black}
\section{Algorithm Design}
Theorems~\ref{thm:approx_err_finite_class}-\ref{thm:function_mismatch} assert that solving the formulation in \eqref{eq:sample} removes the nonlinear distortion $\bm g$. However, implementing the criterion is nontrivial. 
In this section, we develop an algorithm to realize the learning criterion in \eqref{eq:sample}.
\subsection{Challenge: Enforcing Invertibility}
One particular challenge comes from the constraint that $f_m$ is invertible, which is critical for avoiding trivial solutions, as alluded in the proof of Theorem~\ref{thm:main_population}; also see discussions in \cite{yang2020learning}.
In \cite{yang2020learning}, the implementation of \eqref{eq:sample} is by recasting the problem into a constrained optimization problem:
\begin{align}\label{eq:bo_reformulation}
\minimize_{f_m \in {\cal F}_+,\forall m}~\frac{1}{N}\sum_{\ell=1}^N(1-\bm 1^\T \bm f(\bm x_\ell))^2,
\end{align}
where ${\cal F}_+ = \{ f(x) ~|~f(x)=\bm w_2^\T{\bm \zeta}(\bm w_1x),~\w_1,\w_2>0  \}$.
The idea of adding $\bm w_i>\bm 0$ is to ensure that the learned $f_m$'s are invertible---It is not hard to see that positive neural networks [cf. Eq.~\eqref{eq:NNclass}] equipped with invertible activation functions are always invertible functions.
There are a couple of challenges.
First, the reason why NNs are employed for parametrizing $f_m$ is that NNs are universal approximators. However, it is unclear if the universal approximation property can still be retained after adding the positivity constraints. Second, in terms of optimization, the reformulation in \eqref{eq:bo_reformulation} is not necessarily easy. Note that many classic first-order optimization approaches, e.g., projected gradient, works over convex closed sets. However, this reformulation admits an open set constraint, which is normally handled by some more complex algorithms, e.g., the interior point methods. The work in \cite{yang2020learning} ended up employing a Gauss-Newton method with trust-region based updates for handling \eqref{eq:bo_reformulation}, which faces scalablity challenges.

\subsection{Proposed: Neural Autoencoder Based Reformulation}
We propose the following reformulation of \eqref{eq:sample}:
\begin{mdframed}
\begin{subequations}\label{eq:auto}
\begin{align}
(\widetilde{\sf P})\quad    \minimize_{\bm \theta}&~\frac{1}{N}\sum_{\ell=1}^N\left\|\bm q\left( \bm f(\x_\ell)\right) -\x_\ell \right\|_2^2\\
    {\rm subject~to}&~ \bm 1^\T\bm f(\bm x_\ell)-1 =0,\ \forall \ell\in[N],
\end{align}
\end{subequations}
\end{mdframed}

Our formulation uses a reconstruction NN $\bm q(\cdot)=[q_1(\cdot),\ldots,q_M(\cdot)]^\T$ where $q_m(\cdot):\mathbb{R}\rightarrow \mathbb{R}$ together with $\bm f$ such that
\[ f_m,q_m  \in {\cal F}= \{ f(x):\mathbb{R}\rightarrow \mathbb{R} ~|~f(x)=\bm w_2^\T{\bm \zeta}(\bm w_1x) \}, \]
and $\bm \theta$ in \eqref{eq:auto} collects all the network parameters from $f_m,q_m$ for all $m$.
Unlike the network in \cite{yang2020learning}, the proposed method does not impose any constraints on $\bm \theta$. Instead, the invertibility is promoted by introducing the reconstruction network $q_m$ for $f_m$. Note that if $q_m\circ f_m(x_m)=x_m$ for all $\bm x$ and all $m$, it means that $f_m(x)$ is invertible over ${\cal X}$. This simple idea allows us to work with conventional NNs (instead of positive NNs)---so that the universal function approximation property of $f_m$ is not hindered. 

\begin{figure}[t!]
    \centering
	\subfigure{\includegraphics[page=2, clip, trim=3cm 2.5cm 3cm 2.5cm, width=0.7\textwidth]{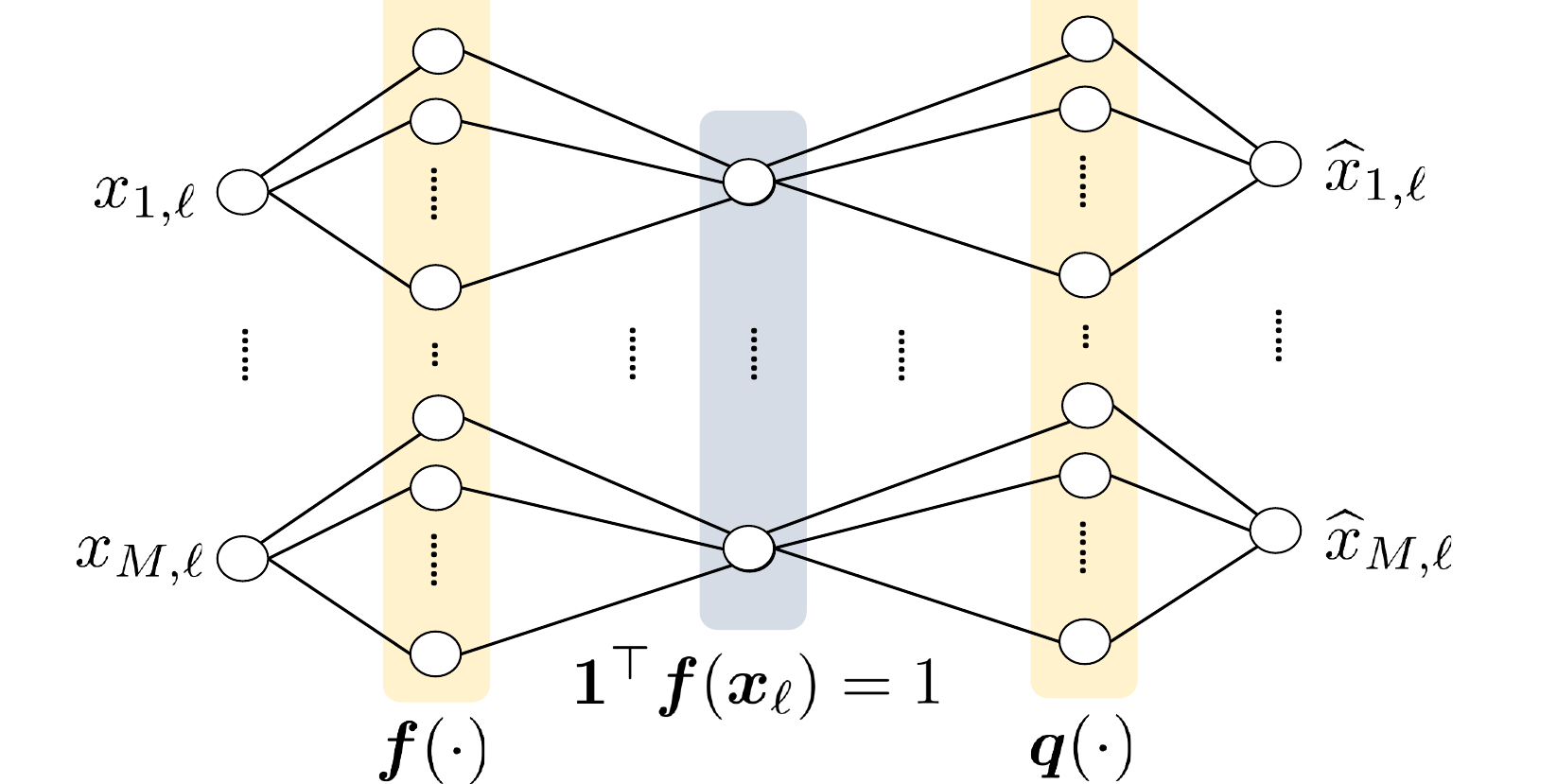}}
    \caption{The `constrained autoencoder' neural network structure of the proposed formulation.}\label{fig:network}
\end{figure}

A remark is that the nonlinear dimensionality reduction plus reconstruction idea is a classic NN architecture for unsupervised learning, i.e., the neural autoencoder \cite{hinton1994autoencoders}. There, $\bm f$ and $\bm q$ are often fully connected neural networks (FCN) or convolutional neural networks (CNN), and constraints are not considered.
Our formulation can be understood as a specially constructed constrained autoencoder. In our case, the `bottleneck layer output' is constrained to be sum-to-one. In addition, $\bm f$ and $\bm q$ consist of individual neural networks for their elements; see Fig.~\ref{fig:network}.

Solving the problem in \eqref{eq:auto} is hard.
Nonetheless, we find that it is not really necessary to solve the formulated problem in \eqref{eq:auto} to optimality---finding a feasible solution often suffices to achieve our goal. Note that finding a feasible solution of Problem~\eqref{eq:auto} is also not an easy task, but practical effective algorithms exist for this purpose, as one will see.

To proceed, we show the following proposition that connects a set of feasible points of \eqref{eq:auto} and the desired solution. To see this, let us denote $v_{N,\bm \theta}$ as the objective value of $(\widetilde{\sf P})$ under $N$ samples and a solution $\bm \theta$. In our definition, $v_{\infty,\bm \theta}$ corresponds to the population case where the sample average in \eqref{eq:auto} is replaced by expectation.

\begin{Prop}\label{prop:kkt}
    Assume that {the generative model in Theorem~\ref{thm:main_population} holds, that} $M=K$ and that a feasible point of \eqref{eq:auto} is attained at $\bm \theta$.
    In addition, the corresponding objective value ${v}_{\infty,\bm \theta}$ satisfies
${v}_{\infty,\bm \theta} <\sum_{k=1}^K\sigma^2_k,$
     where $\sigma_k^2=\mathbb{V}[x_k]$ is the variance of $x_k$. Then, the learned $\widehat{f}_{m}$ for $m=1,\ldots,M$ are invertible and $\widehat{f}_m\circ g_m$ is affine with probability one.
\end{Prop}
\noindent
{\bf Proof:}
In the population case, consider the first-order derivatives of $\phi(\bm s)=\bm{1}^\top{\bm h}(\A\s)=1$. Putting $\frac{\partial \phi(\bm s)}{\partial s_i}$ for $i\in[K-1]$ together, we have a system of linear equations:
\[ \bm B^\T \bm h'=\bm 0, \]
where $\bm h'=[h_1',\ldots,h_K']^\T$, and $\bm{B}=[\bm{b}_1,\ldots,\bm{b}_{K-1}]\in\mathbb{R}^{K\times (K-1)}$ is defined in Theorem~\ref{thm:main_population} with $\bm{b}_i=[a_{1,i}-a_{1,K},\ a_{2,i}-a_{2,K},\ \ldots,\ a_{K,i}-a_{K,K}]^\top$ where $i=1,\ldots,K-1$. 

Note that when $h'_i=0$ for a particular $i$, all the other $h'_j$ with $j\neq i$ have to be zeros. This is because any $(K-1)\times(K-1)$ submatrix of $\bm B$ is nonsingular with probability one, due to the {assumption} that the $\bm{A}$ matrix {follows any joint absolutely continuous distribution; see the proof technique in Theorem~\ref{thm:main_population} for showing ${\bm G}$ to be full rank.} Therefore, when any $h'_i=0$, which means $h_i$ is not invertible, all the other $h_j$'s are also not invertible. 
Thus, the $h_i$'s are either all non-invertible or they are all invertible.

{Note that $v_{\infty,\bm \theta}$ corresponds to the population case.
Hence, following the proof of Theorem \ref{thm:main_population}, one can show that $h''_m=0$ always holds.} 
Therefore, if $h_m$ is not invertible, it has to be a constant. As a result, the only possible non-invertible case is that all $h_m$'s are constant functions.
{Therefore, $q_k\circ f_k\circ g_k(\cdot)=q_k\circ h_k(\cdot)$ has to be a constant, no matter what is $q_k$---since the input to $q_k$ is a constant.
Let us denote $c_k=q_k\circ{f}_k(x_k)$ as the constant.} 
Then, we have
\begin{align*}
    &\mathbb{E}_{\bm x\sim {\cal D}}\left[\|\bm{q}(\bm{f}(\bm{x}))-\bm{x}\|_2^2\right]= \mathbb{E}\left[\sum_{k=1}^K(x_{k}-\mu_k+\mu_k-c_{k})^2 \right]\\
    & = \sum_{k=1}^{K}\mathbb{E} \left[({x}_{k}-\mu_k)^2+2({x}_{k}-\mu_k)(\mu_k-c_{k})+(\mu_k-c_{k})^2 \right]\\
    &=\sum_{k=1}^{K}\mathbb{E}\left[ ({x}_{k}-\mu_k)^2\right]+\mathbb{E}\left[(\mu_k-c_k)^2\right]\geq \sum_{k=1}^K\sigma^2_k,
\end{align*}
where $\mu_k=\mathbb{E}[x_k]$ and $\sigma^2_k=\mathbb{V}[x_k]$ are the mean and variance of ${x}_{k}$, respectively.
Note that the above is a necessary condition of all $h_k$'s being zeros.
Therefore, if $v_{\infty,\bm \theta}<\sum_{k=1}^K\sigma^2_k$ is satisfied by a feasible point $\bm \theta$,
then all the learned functions parametrized by $\bm \theta$ are invertible and the composite functions $f_m\circ g_m$'s are all affine. \hfill $\square$

\bigskip

Proposition~\ref{prop:kkt} indicates that if one attains a feasible solution of the expectation version of Problem~\eqref{eq:sample} with a sufficiently small objective value in the {\it population} case,
then the learned $\bm f$ is invertible and $\bm f\circ\bm g$ is affine. 
This also has practical implications: $\sigma_k^2$ can be estimated by sample variance, and the conditions in Proposition~\ref{prop:kkt} can be checked when a solution $\bm \theta$ is obtained.
Again, in practice, one can only work with finite samples (i.e., $N<\infty$).
Nonetheless, the next proposition shows that when $N$ is large enough, any feasible solution of \eqref{eq:sample} is also {an approximately feasible} solution of the population case. In addition, $v_{N,\bm \theta}$ is close to $v_{\infty,\bm \theta}$.

\begin{Prop}\label{prop:samplekkt}
    Assume that Assumptions~\ref{as:exact_finite} and \ref{as:nn} hold. Then, when 
    \begin{align*}
        N= \max\left\{ \Omega\left(\frac{M^4z^4B^8C^4_x\left(\sqrt{R}+\sqrt{{\log(1/\delta)}}\right)^2}{\varepsilon^2}\right),\ \right.
        \left. \Omega\left(\frac{M^2z^8B^{16}C_x^4\left(\sqrt{R}+\sqrt{\log(1/\delta)}\right)^2}{\varepsilon^2}\right)\right\},
    \end{align*}
    the following holds with probability of at least $1-\delta$: 
    \begin{align}\label{eq:sample_kkt}
    \left| \mathbb{E}[{\cal C}_{\bm \theta^\ast}(\x)] \right|^2 &\leq \varepsilon,~
    v_{\infty,\bm \theta^\ast} \leq v_{N,\bm \theta^\ast} + \varepsilon,
    \end{align}
  where ${\cal C}_{\bm \theta}(\bm x)=\bm 1^\T\bm f(\bm x)-1$ with $\x\in {\cal X}$.
    
\end{Prop}

The proof of Proposition~\ref{prop:samplekkt} can be found in Appendix~\ref{appdx:samplekkt}. Note that the first inequality in \eqref{eq:sample_kkt} is the basic ingredient for showing that $\mathbb{E}[\|\widehat{\bm h}''\|_2^2]\approx 0$ in the proofs of Theorems~\ref{thm:approx_err_finite_class}-\ref{thm:function_mismatch}. Hence, \eqref{eq:sample_kkt} also implies that with a sufficient number of samples, the nonlinear distortions can be approximately removed.

\subsection{Augmented Lagrangian Multiplier Method}
Finding a feasible point of Problem~\eqref{eq:sample} is nontrivial. 
Note that any Karush--Kuhn--Tucker (KKT) point is feasible. Hence, one can leverage effective KKT-point searching algorithms from nonlinear programming.
To proceed, we consider the standard augmented Lagrangian $ {\cal L}(\bm \theta,\bm \lambda)$ that can be expressed as follows:
\begin{equation*}\label{eq:lagrangian}
\begin{aligned}
    \frac{1}{N}\sum_{\ell=1}^N{\cal J}_{\bm \theta}(\bm x_\ell) + \frac{1}{N} \sum_{\ell=1}^N{\lambda_\ell} {\cal C}_{\bm \theta}(\bm x_\ell)+\frac{\rho}{2N} \sum_{\ell=1}^N \left| {\cal C}_{\bm \theta}(\bm x_\ell)  \right|^2,
\end{aligned}
\end{equation*}
where ${\cal J}_{\bm \theta}(\bm x_\ell)=\left\|\bm q\left( \bm f(\x_\ell)\right) -\x_\ell \right\|_2^2$, ${\cal C}_{\bm \theta}(\bm x_\ell)=\bm 1^\T\bm f(\bm x_\ell)-1$ and $\bm{\lambda}=[\lambda_1,\cdots,\lambda_N]$ collects the dual variables associated with all the constraints.
Here $\rho>0$ reflects the `importance' of the equality constraint.
The algorithm updates $\bm \theta$ and $\bm \lambda$ using the following rule:
\begin{subequations}\label{eq:alg}
\begin{align}
    \bm \theta^{t+1} &\leftarrow \arg\min_{\bm \theta} ~{\cal L}(\bm \theta,\bm \lambda^t)~\text{(inexact min.)},  \label{eq:theta_update}\\
    \lambda_\ell^{t+1}&\leftarrow \lambda_\ell^t + \rho^t {\cal C}_{\bm \theta^{t+1}}(\bm x_\ell),~    \rho^{t+1}\leftarrow \kappa\rho^t,
\end{align}
\end{subequations}
where {$\kappa>1$ is a pre-specified parameter.}
Note that in \eqref{eq:theta_update}, `inexact minimization' means that one needs not solve the subproblem exactly.
The proposed algorithm here falls into the category of augmented Lagrangian multiplier methods \cite{bertsekas1998nonlinear}. Applying classic convergence results for this class of algorithms, one can show that
\begin{Fact}\label{fact:convergence}
Assume that ${\cal L}(\bm \theta,\bm \lambda)$ is differentiable with respect to $\bm \theta$, that each update of $\bm \theta$ in \eqref{eq:theta_update} satisfies
$  \| \nabla {\cal L}(\bm \theta^{t+1},\bm \lambda^t) \|_2^2\leq \epsilon^{t} , $
and that $\epsilon^t\rightarrow 0$.
Also assume that $\bm \lambda^t<\infty$ for all $t$.
Then, every limit point of the proposed algorithm is a KKT point.
\end{Fact}
\noindent
{\bf Proof:}
The proof is by simply invoking \cite[Proposition 5.5.2]{bertsekas1998nonlinear}. \hfill $\square$

\bigskip

Per Fact~\ref{fact:convergence}, the primal update for $\bm \theta$ should reach an $\epsilon$-stationary point of the primal subproblem.
There are many options for this purpose. In our case, the primal update involves NN-parametrized $\bm f$ and $\bm q$, and thus using gradient-based methods are natural---since back-propagation based gradient computation of functions involving NNs is sophisticated and efficient.
To further improve efficiency of the primal update, one can use {\it stochastic gradient descent} (SGD)-based methods that have proven effective in neural network optimization, e.g., \texttt{Adam} and \texttt{Adagrad}. In practice, increasing $\rho^t$ to infinity may cause numerical issues, as mentioned in \cite{bertsekas1998nonlinear,shi2020penalty}.
Pragmatic remedies include increasing $\rho^t$ in a slow rate (i.e., using $\kappa\approx 1$ so that the algorithm may converge before $\rho^t$ becomes too large) or fixing $\rho^t$---which both work in practice, according to our simulations.
Another remark is that the sample complexity proofs in Theorem~\ref{thm:approx_err_nn_class} and Proposition~\ref{prop:samplekkt} both assume using neural networks whose weights are bounded. In our implementation, we do not explicitly impose such constraints since it may complicates the computation. Nonetheless, unbounded solutions are rarely observed in our extensive experiments.

\section{Numerical Results}
In this section we showcase the effectiveness of the proposed approach using both synthetic and real data experiments. The proposed method is implemented in Python (with Pytorch) on a Linux workstation with 8 cores at 3.7GHz, 32GB RAM and GeForce GTX 1080 Ti GPU.
The code and data will be made publicly available upon publication.

\subsection{Synthetic Experiment}
In the synthetic data experiments, we use the {\it nonlinear matrix factor recovery} (\texttt{NMFR}) method \cite{yang2020learning} as our main baseline for nonlinearity removal, since the method was developed under the same generative model in this work. 
We name the proposed method as {\it constrained neural autoencoder} (\texttt{CNAE}). {We also use generic dimensionality reduction tools (e.g., \texttt{PCA}, \texttt{NMF} and the general purpose autoencoder (\texttt{AE})\cite{hinton1994autoencoders}) as benchmarks in different experiments.} 
Note that both \texttt{NMFR} and \texttt{AE} are neural network-based methods. 
The \texttt{NMFR} method uses positive NNs to parameterize $f_m$'s. In our experiments, we use the setting following the default one in \cite{yang2020learning} (i.e., using 40-neuron one-hidden-layer and \texttt{tanh} activation functions to represent $f_m$) unless specified. 
Since \texttt{NMFR} uses a Gauss-Newton algorithm that is computationally expensive, increasing the number of neurons beyond the suggested setting in \cite{yang2020learning} is often computationally challenging.
The \texttt{AE}'s encoder and decoder are represented by fully connected one-hidden-layer 256-neuron neural networks, where the neurons are ReLU functions. 

For the proposed \texttt{CNAE} method, 
{we use an individual similar network structure as in \texttt{AE}.
The difference is that \texttt{CNAE} uses a single-hidden-layer fully connected network with $R$ neurons to represent {\it each} of $f_m:\mathbb{R}\rightarrow\mathbb{R}$ (and also $q_m$) for $m\in[M]$, as opposed to the whole $\bm f:\mathbb{R}^{M}\rightarrow\mathbb{R}^K$. Since the simulations serve for proof-of-concept and theorem validation, our focus is not testing multiple neural network structures. Nonetheless, we find that using a single-hidden-layer network with a moderate $R$ (e.g., 64 and 128) is generally effective under our simulation settings. For more challenging real experiments, the network architecture design may be assisted by procedures such as cross validation.}
We employ the \texttt{Adam} algorithm \cite{kingma2014adam} for updating the primal variable $\bm \theta$. The initial learning rate of \texttt{Adam} is set to be $10^{-3}$. The batch size is $1,000$, i.e., every iteration of \texttt{Adam} randomly samples 1,000 $\bm x_\ell$ to compute the gradient estimation for updating $\bm \theta$. 
We run up to 100 epochs of \texttt{Adam} for solving the primal subproblem in \eqref{eq:theta_update} (an epoch means a full sweep of all the data samples, if gradient estimations are constructed by sampling without replacement). In addition, if $(1/N)\sum_{\ell=1}^N |{\cal C}_{\bm \theta}(\bm x_\ell)|^2$ goes below $10^{-5}$, we stop the algorithm---as a feasible solution of \eqref{eq:auto} is approximately reached.
In our Lagrangian multiplier method, $\rho^t=\rho=10^2$ is used.

To verify the nonlinearity removal theorem, the theorems indicate that $\bm{F}=[\widehat{\bm{f}}(\bm{x}_1),\ldots,\widehat{\bm{f}}(\bm{x}_N)]\approx \bm D\bm A{\bm S}$, where $\bm S=[\bm s_1,\ldots,\bm s_N]$, is expected under the proposed learning criterion, where $\bm D$ is a full rank diagonal matrix.
Hence, we adopt the {\it subspace distance} ${\sf dist}({\cal S},\widehat{\cal S})=\|\bm{P}_{\bm S^\T}^{\perp}\bm{Q}_{\bm F^\T}^\T\|_2$ with ${\cal S}={\rm range}(\S^\T)$ and $\widehat{\cal S}={\rm range}(\bm{F}^\T)$ as the performance metric, where $\bm{Q}_{\bm F^\T}$ is the orthogonal basis of $\bm{F}^\top$ and $\bm{P}_{\bm S^\T}^{\perp}$ is the orthogonal projector of the orthogonal complement of ${\rm range}(\bm S^\T)$.
This metric ranges within $[0,1]$ with $0$ being the best. 
{When it comes to evaluating the performance of recovering $\bm s_\ell$, we also use another baseline, namely, the {\it minimum-volume enclosing simplex} (\texttt{MVES}) algorithm that provably identifies the latent components from SC-LMM under some conditions \cite{MVES,fu2015blind,fu2016robust}.}

For the first simulation, we let $M=K=3$. The three nonlinear {distortion} functions are: $g_1(x)=5\text{sigmoid}(x)+0.3x$, $g_2(x)=-3\text{tanh}(x)-0.2x$ and $g_3=0.4\exp(x)$, and $N=5,000$. The matrix $\A$ is drawn from the standard Gaussian distribution. 

\begin{figure}[t!]
    \centering
	\subfigure{\includegraphics[width=0.32\textwidth]{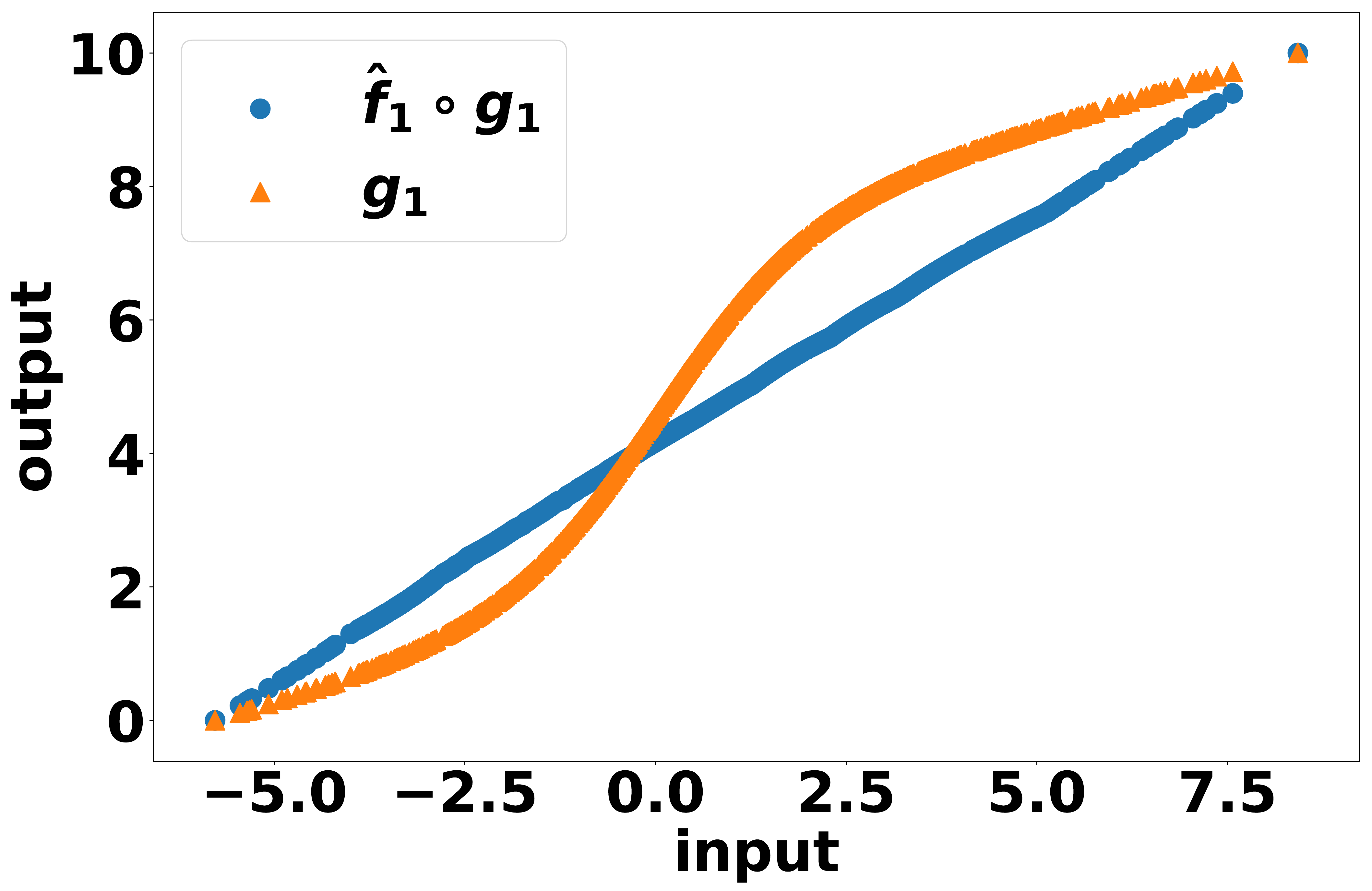}}
	\subfigure{\includegraphics[width=0.32\textwidth]{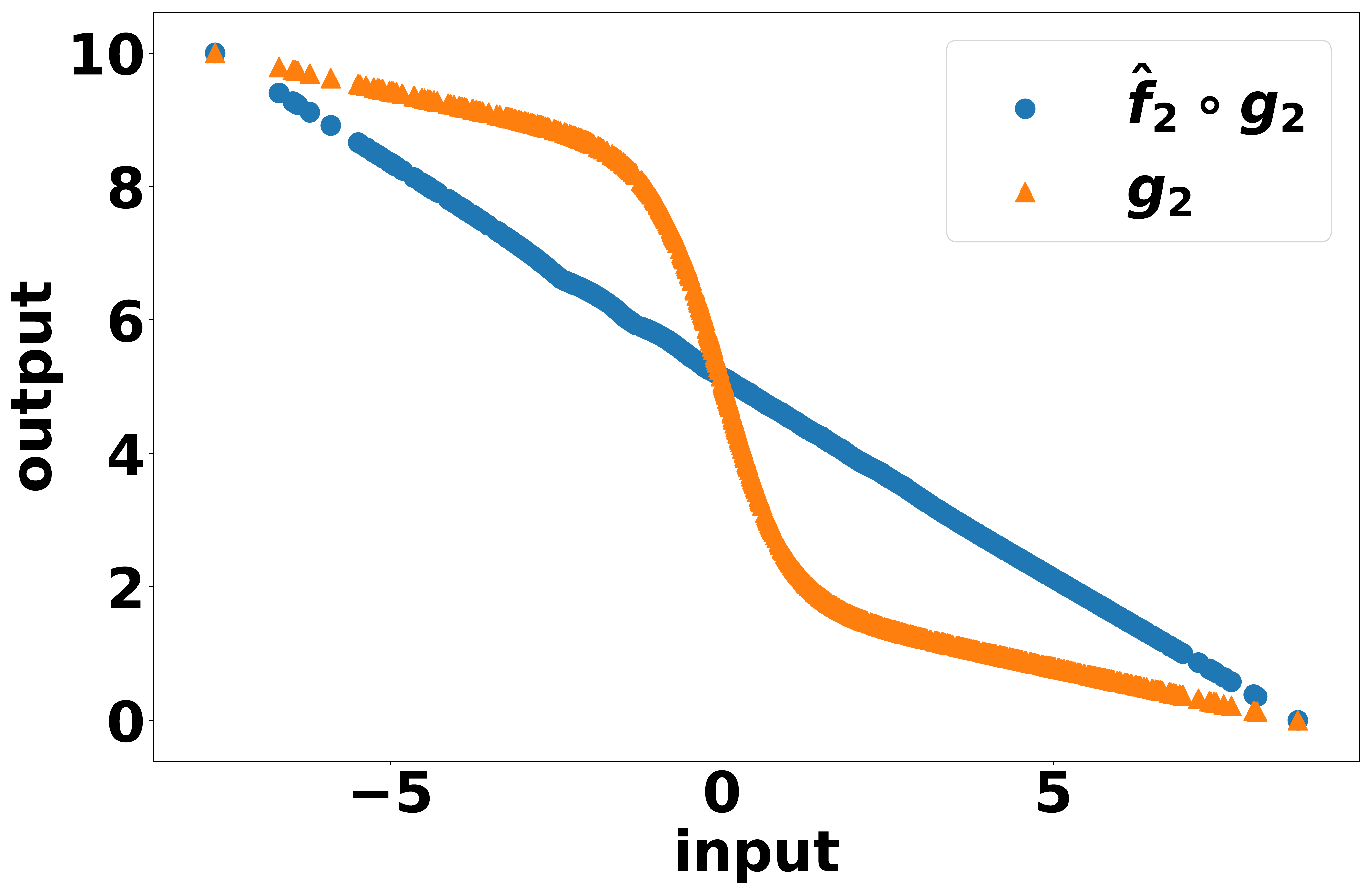}}
	\subfigure{\includegraphics[width=0.32\textwidth]{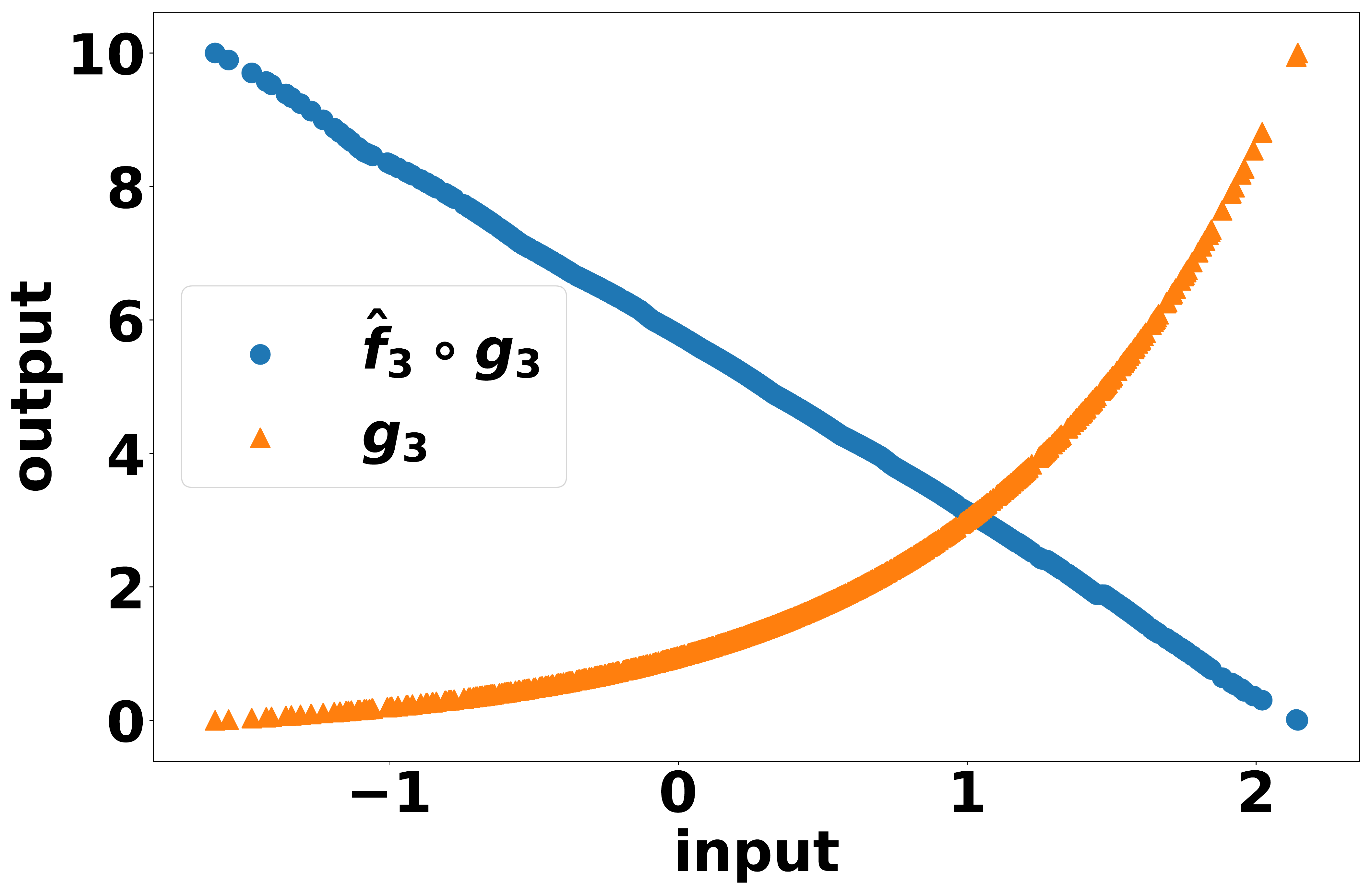}}
	
	\subfigure{\includegraphics[width=0.32\textwidth]{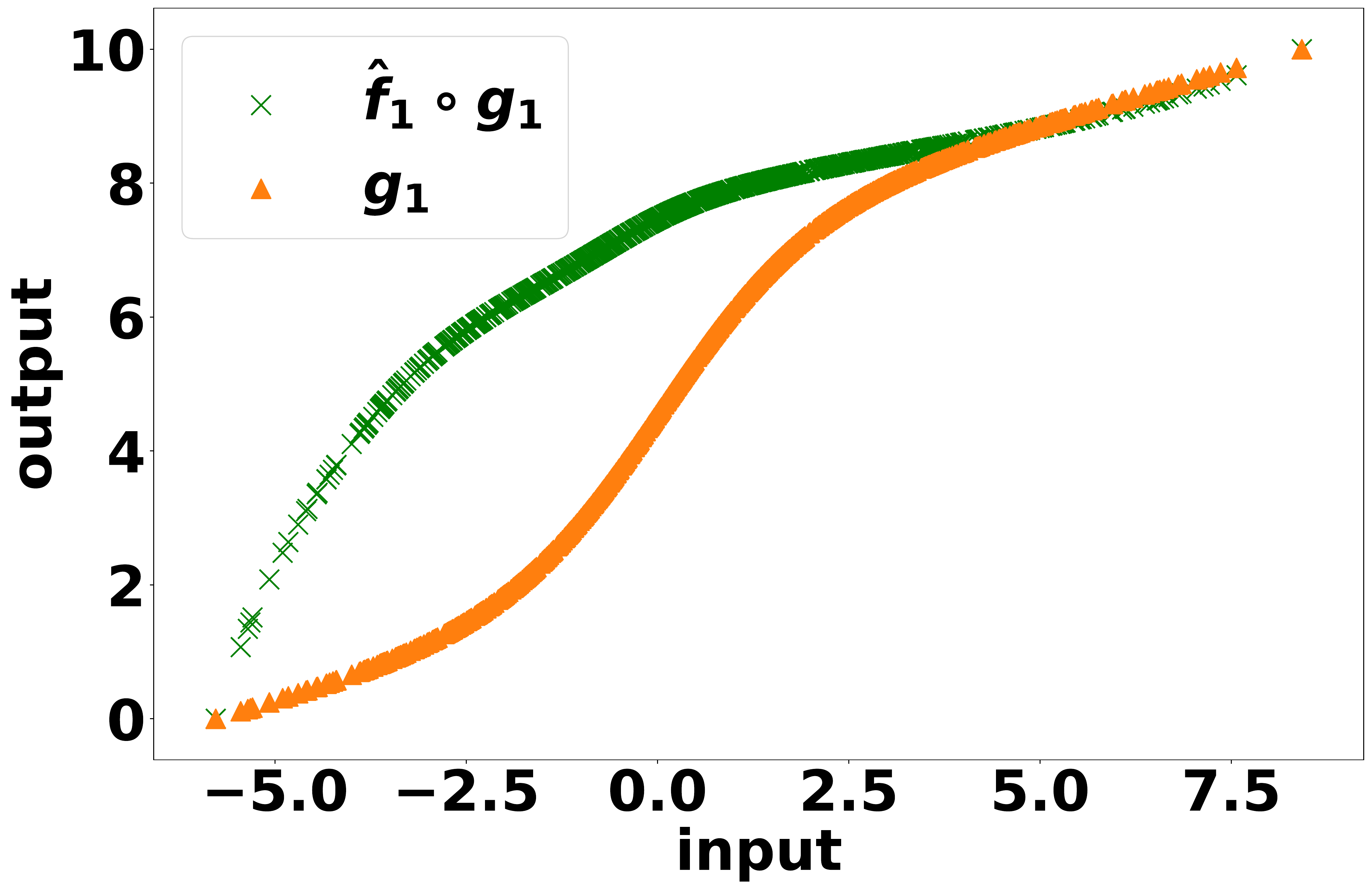}}
	\subfigure{\includegraphics[width=0.32\textwidth]{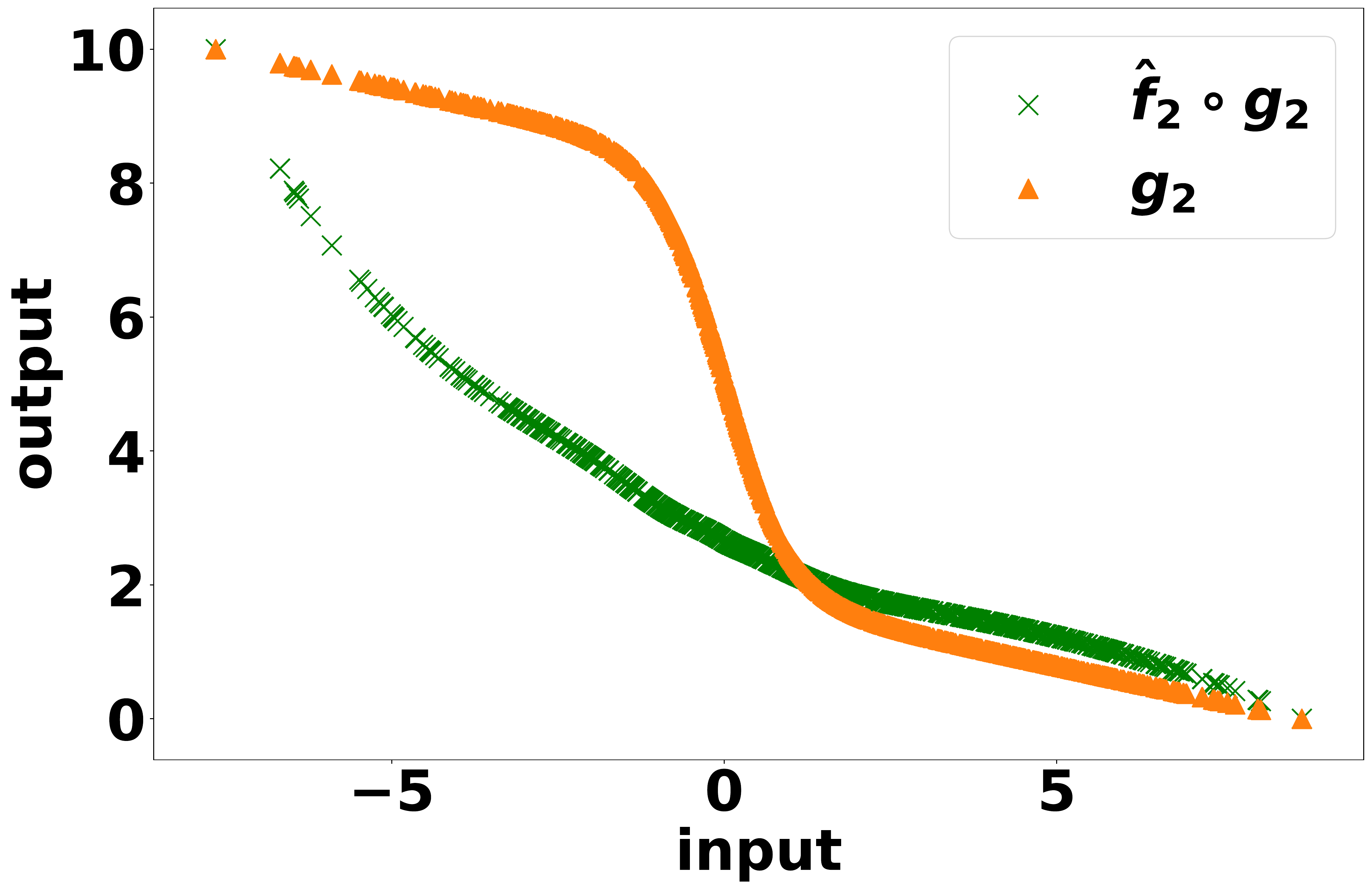}}
	\subfigure{\includegraphics[width=0.32\textwidth]{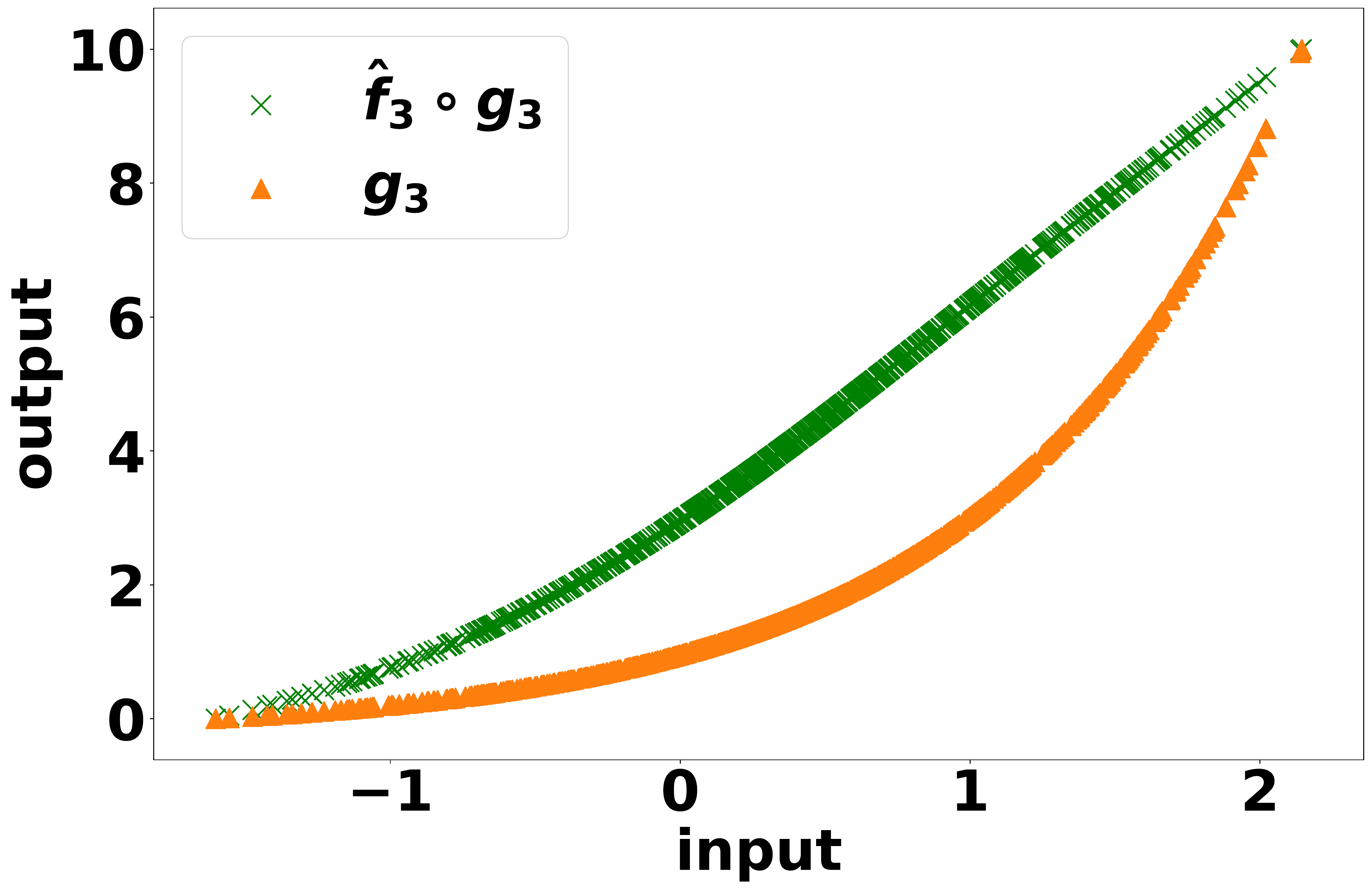}}
    \caption{Learned composition functions by the proposed \texttt{CNAE} (top) and \texttt{NMFR} in \cite{yang2020learning} (bottom); both methods use 64-neuron single-hidden-layer neural networks to model the nonlinear functions.}\label{fig:composite}
\end{figure}

Fig. \ref{fig:composite} shows that the $\widehat{f}_m\circ g_m$'s learned by the proposed method are visually affine functions. In contrast, \texttt{NMFR} fails to remove the nonlinear distortions. This may be a bit surprising at first glance, since both methods start with the same formulation in \eqref{eq:population}.
However, the implementation of \texttt{NMFR} is based on {\it positive} neural networks, which does not necessarily have the universal approximation ability---which is critical for learning complex nonlinear functions. 
Note that in this case, we deliberately used the same number of neurons (i.e., 64) for the forward networks (i.e., $f_m$'s) in both methods in order to achieve better expressiveness for \texttt{NMFR}, despite using a number of neurons beyond the default 40 substantially slows down the algorithm of \texttt{NMFR}. Nonetheless, this comparison indicates that the lack of effectiveness of \texttt{NMFR} may be more likely the result of the positive neural network used, instead of the small number of neurons that it can afford in computation.

To quantitatively assess the effectiveness of nonlinearity removal, we show the subspace distance between the range spaces recovered latent components (i.e., $\bm F^\T$) and the ground truth $\bm S^\T$. The simulations are run under various numbers of samples including $5,000$, $10,000$ and $20,000$, respectively.

The results are shown in Table~\ref{tab:subspace}, which are averaged from $10$ random trials. 
{One can see that the proposed \texttt{CNAE} method attains the lowest subspace distances in all the cases.
\texttt{MVES} apparently fails since it does not take nonlinear distortions into consideration.
\texttt{NMFR} does not work well as in the first simulation, which may be again because of its positive NN-based implementation strategy.
The generic NN-based unsupervised learning tool, namely, \texttt{AE} (which uses one-hidden-layer fully connected 256-neuron neural networks as its encoder and decoder), also does not output desired solutions. 
This shows the importance of exploiting structural prior information---i.e., the simplex-constrained post-nonlinear model in our case---to attain high-quality solutions when combating nonlinear distortions.}
In addition, one can see that the proposed method exhibits improved performance as the sample size increases, which is consistent with our analysis on sample complexity.
In terms of runtime, our method improves upon \texttt{NMFR} by large margins in all the cases. In particular, when $N=20,000$, the proposed method uses 1/6 of the time that \texttt{NMFR} uses. This is because our framework can easily incorporate efficient NN optimizers such as \texttt{Adam}.

\begin{table}[t!]
\centering
\caption{Subspace distance between the learned ranges of $\bm F$ and $\bm S^\T$ under various $N$'s; $M=K=3$. Entries: dist (running time).}\label{tab:subspace}
\begin{tabular}{c|ccc}
$N$     &  5,000 & 10,000  & 20,000  \\ \hline \hline
\texttt{MVES}  & $0.99$ (0.35s) & $0.99$ (0.15s) & $0.99$ (0.21s)   \\ 
\texttt{AE} & $0.73$ (125.9s)  &  $0.74$ (258.7s)  & $0.73$ (505.4s) \\
\texttt{NMFR} \cite{yang2020learning} & $0.69$ (559.1s)  & $0.78$ (2714.2s) &  $0.77$ (4903.7s)  \\
\texttt{CNAE} (Proposed) & $0.01$ (177.6s) &  $0.008$ (305.2s) & $0.005$ (784.7s)
\end{tabular}
\end{table}

Table~\ref{tab:mse} shows the {\it mean squared error} (MSE) of the estimated $\widehat{\bm S}$, which is defined as follows:
\begin{align*}
    \min_{\bm \pi\in\bm \Pi} \frac{1}{K}\sum_{k=1}^K \left\|\frac{\bm{S}_{k,:}}{\|\bm{S}_{k,:}\|_2}-\frac{\widehat{\bm{S}}_{\pi_k,:}}{\|\widehat{\bm{S}}_{\pi_k,:}\|_2}\right\|_2^2,
\end{align*}
where $\bm \Pi$ is the set of all permutations of $\{1,\cdots,K\}$, $\bm{S}_{k,:}$ and $\widehat{\bm{S}}_{k,:}$ are the ground truth of the $k$th row of $\bm{S}$ and the corresponding estimate, respectively. Note that the $k$th row in $\bm S$ represents the $k$th latent component.
The permutation matrix $\bm \Pi$ is used since mixture learning has an intrinsic row permutation ambiguity in the estimated $\widehat{\S}$ that cannot be removed without additional prior knowledge.
The latent components are estimated by applying the SC-LMM learning algorithm \texttt{MVES} onto $\bm F$ after the nonlinearity removal methods are used to cancel $\bm g$. {Note that the generated $\bm s_\ell$'s are {\it sufficiently scattered} (see definition in \cite{fu2018nonnegative}) in $\bm \varDelta_K$. Hence, $\bm S$ is provably identifiable (up to permutation ambiguities) from $\bm F$ by \texttt{MVES} if $\bm F$ is indeed a linear mixture of $\bm S$.}
The results are also averaged from $10$ random trials. 
From the table, one can see that \texttt{CNAE} largely outperforms the baselines, i.e., by three orders of magnitude in terms of MSE. This also asserts that the nonlinearity removal stage performed by \texttt{CNAE} is successful. 

\begin{table}[t!]
\centering
\caption{MSE between $\bm{S}$ and estimated $\widehat{\bm{S}}$. Entries: mean (std).}\label{tab:mse}
\begin{tabular}{c|ccc}
$N$ & 5,000 & 10,000 & 20,000 \\ \hline\hline
\texttt{MVES}  &  $8.7e{-2}\pm 4.6e{-3}$    & $9.2e{-2}\pm 6.8e{-3}$   & $8.9e{-2}\pm  2.0e{-3}$    \\ 
\texttt{AE}+\texttt{MVES}  & $9.6e{-2}\pm 5.5e{-3}$  &  $9.7e{-2}\pm 4.4e{-3}$ &  $7.9e{-2}\pm 6.9e{-3}$    \\ 
\texttt{NMFR} \cite{yang2020learning}+\texttt{MVES} &  $1.4e{-1}\pm 3.4e{-2}$  &  $1.1e{-1}\pm 3.3e{-2}$ & $9.2e{-2}\pm 1.2e{-2}$ \\
 \texttt{CNAE} (Proposed)+\texttt{MVES} & $4.1e{-4}\pm 3.2e{-4}$ & $4.1e{-5}\pm 1.5e{-5}$ & $1.4e{-5}\pm 9.9e{-6}$
\end{tabular}
\end{table}

\begin{figure*}[t!]
    \centering
	\colorbox{blue!5}{\subfigure{\includegraphics[width=0.23\textwidth]{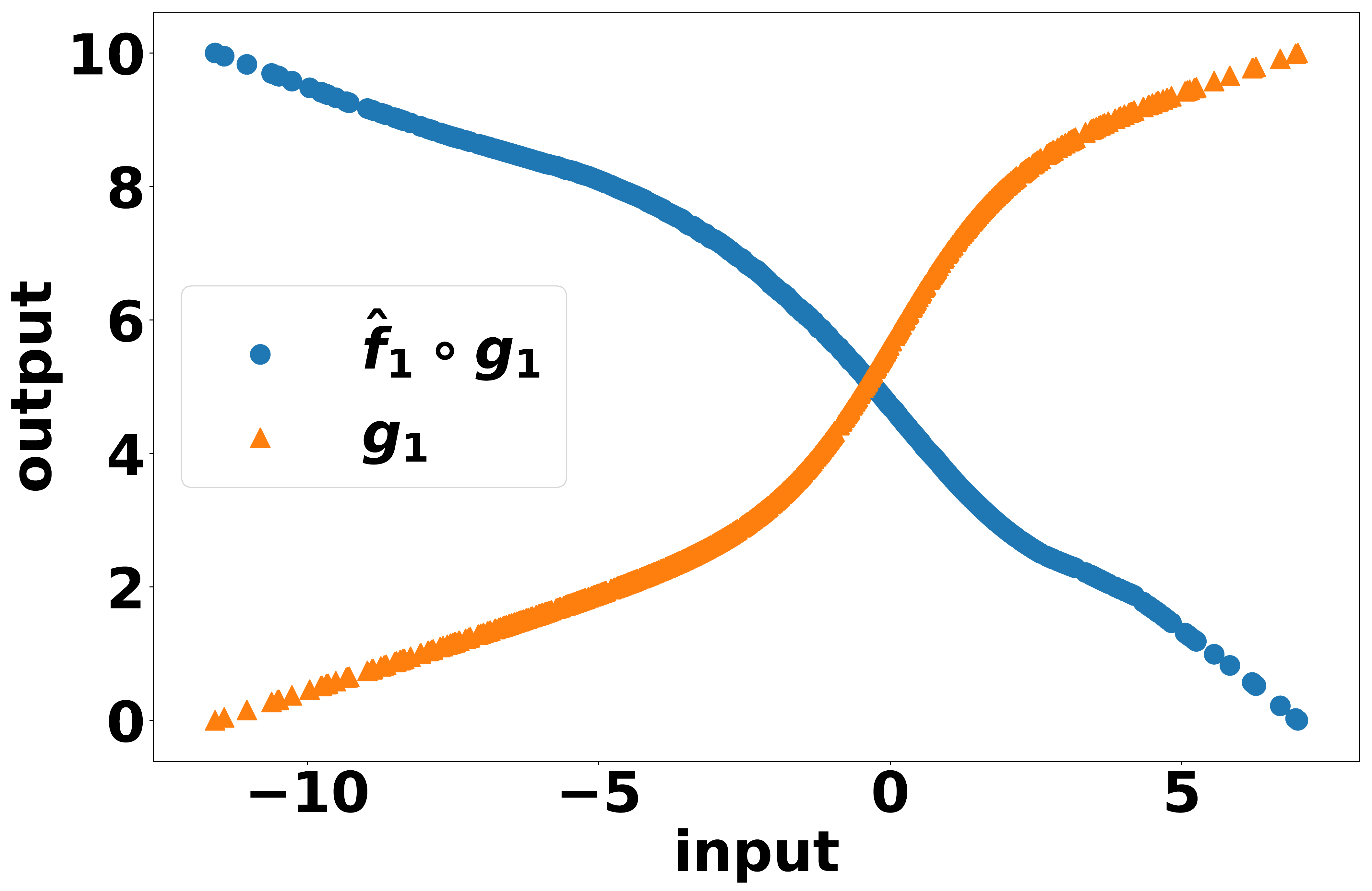}}
	\subfigure{\includegraphics[width=0.23\textwidth]{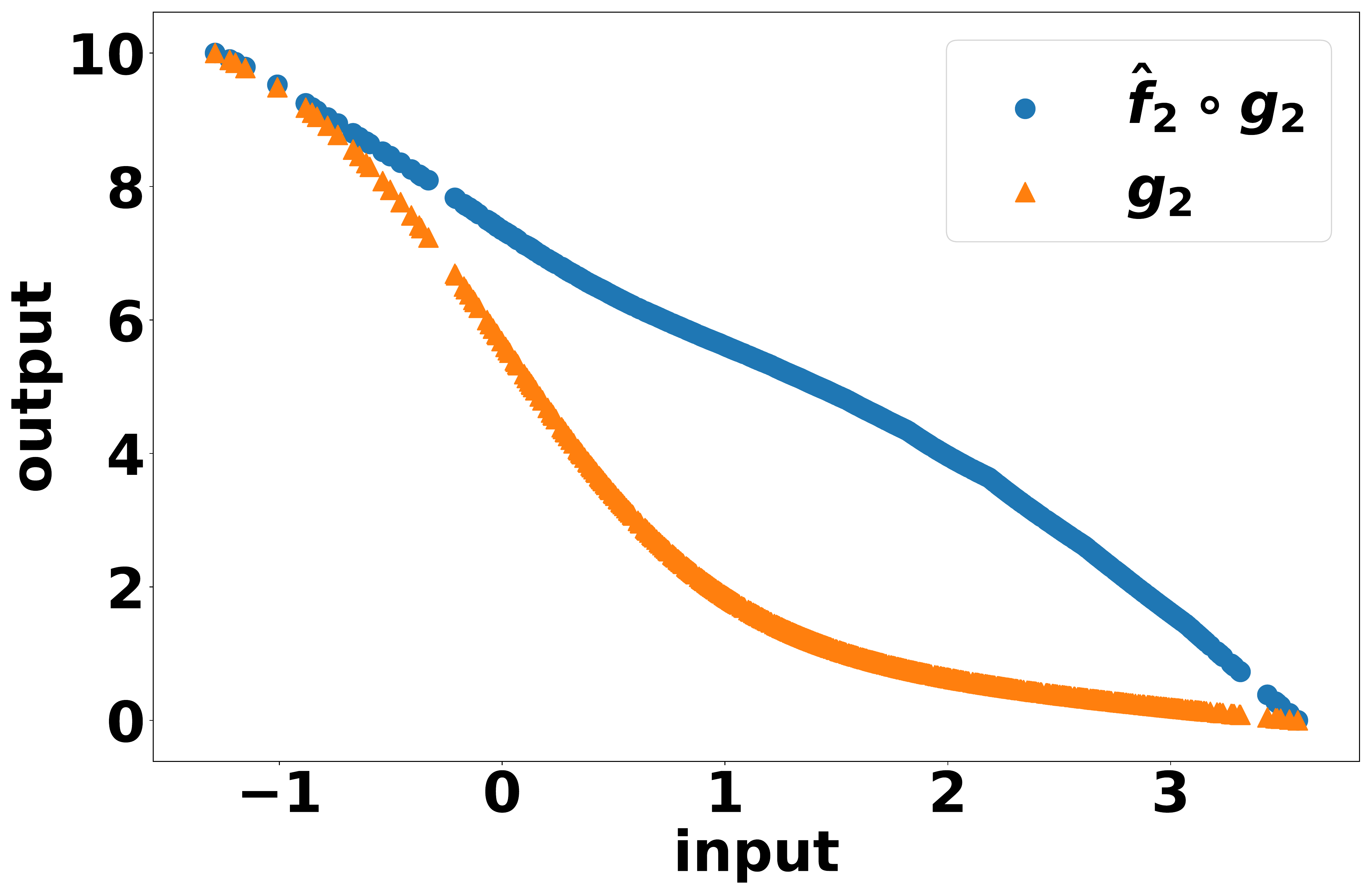}}
	\subfigure{\includegraphics[width=0.23\textwidth]{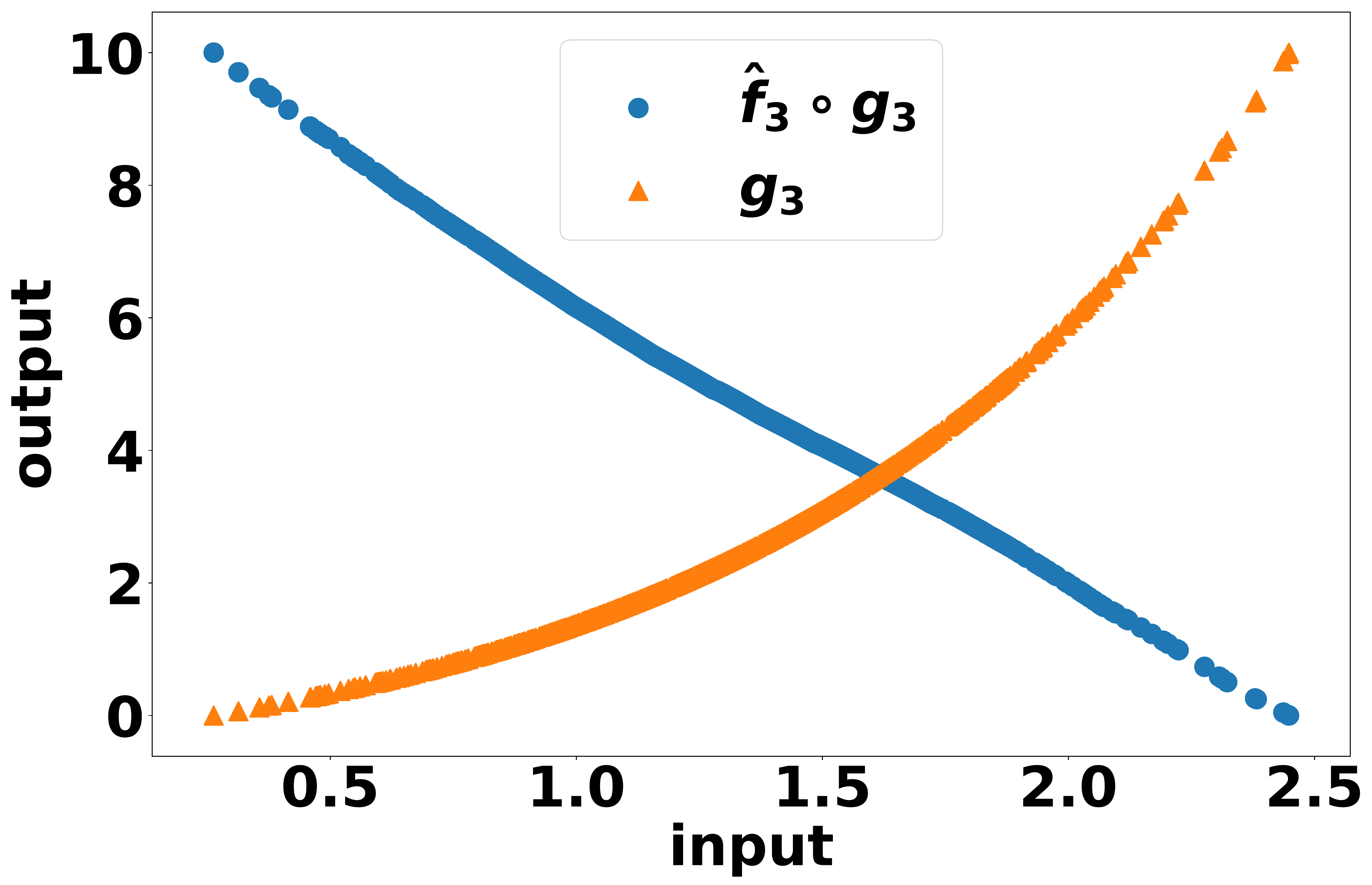}}
	\subfigure{\includegraphics[width=0.23\textwidth]{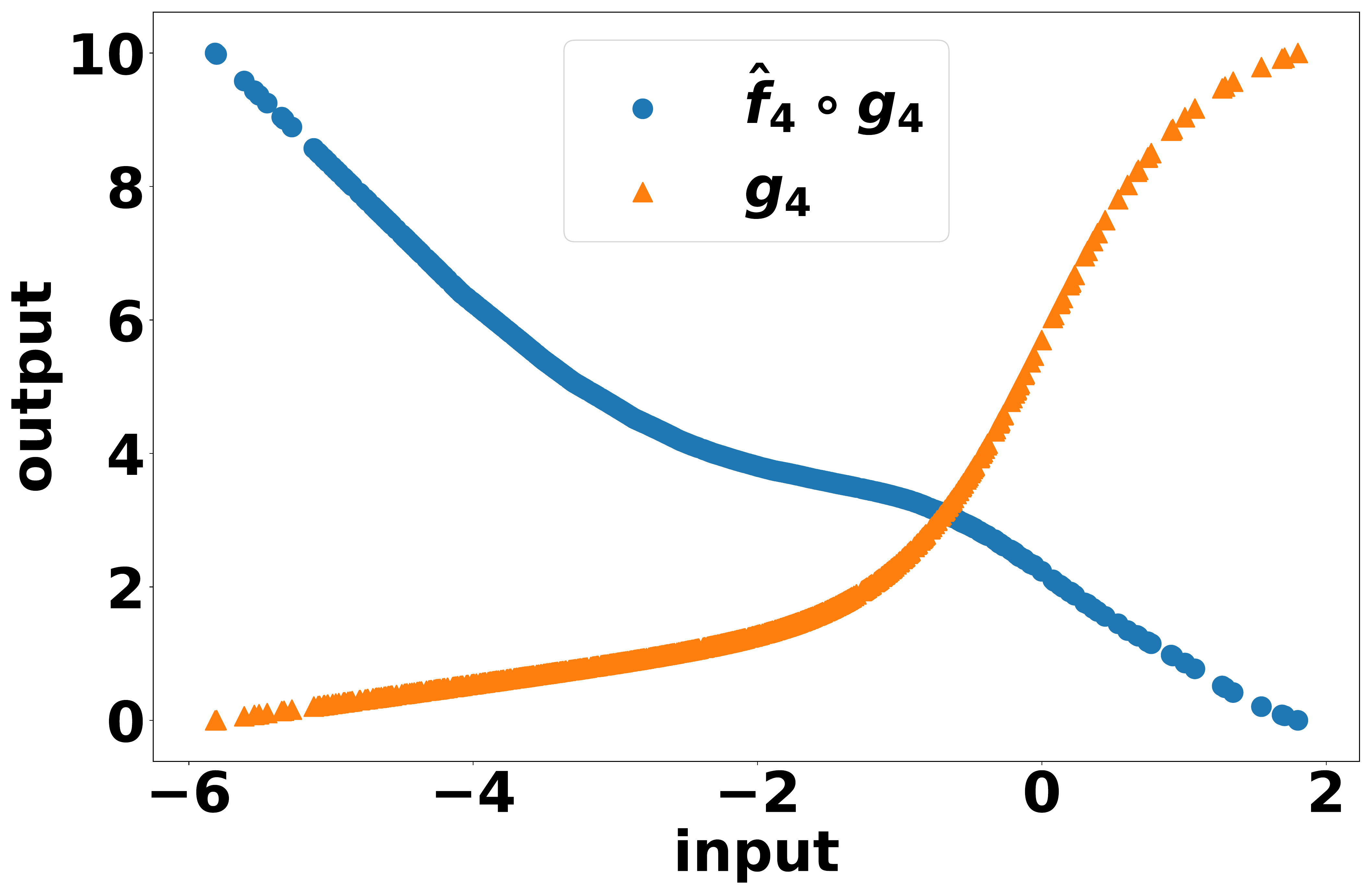}}}
	
	\colorbox{green!5}{\subfigure{\includegraphics[width=0.23\textwidth]{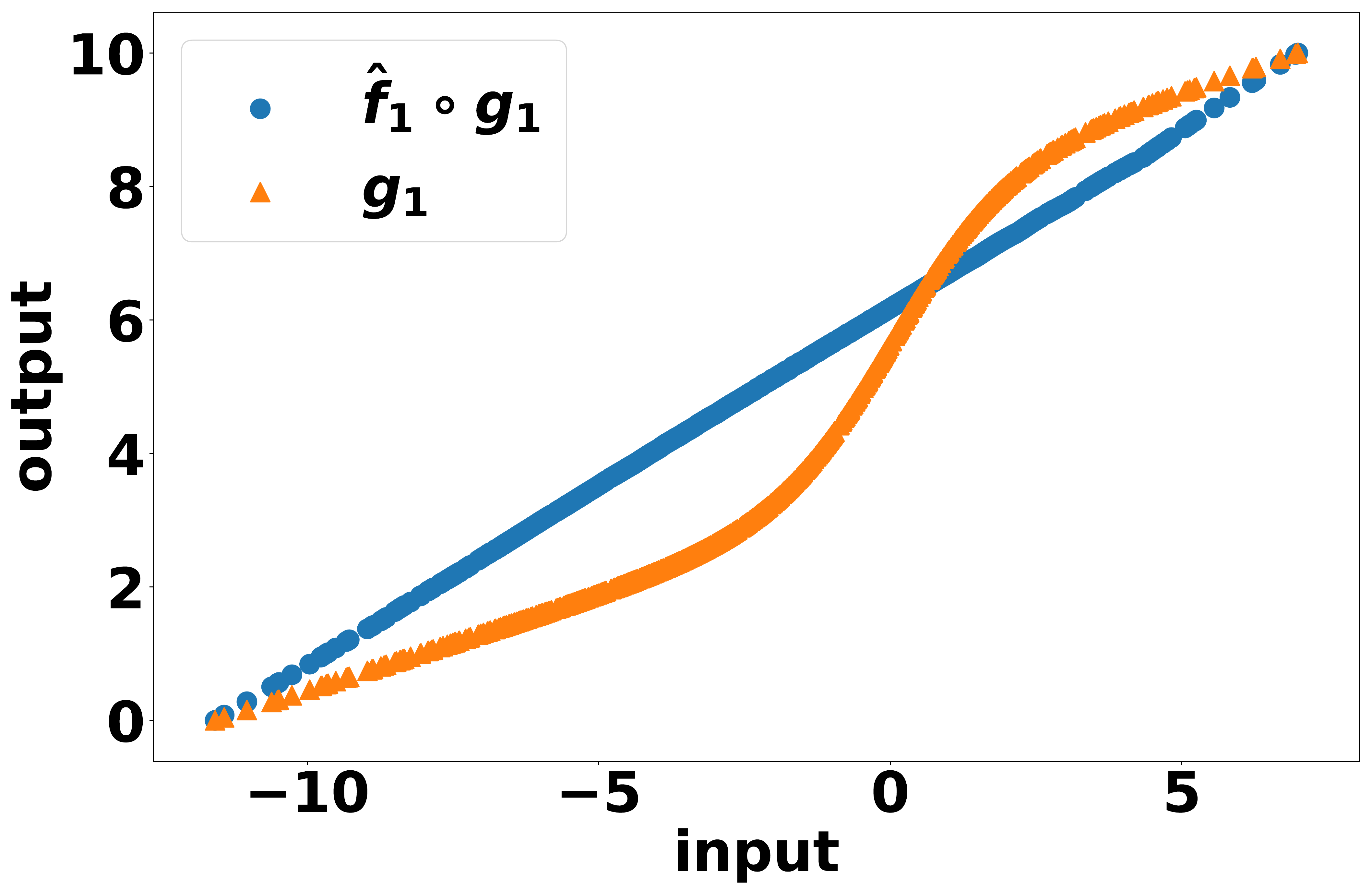}}
	\subfigure{\includegraphics[width=0.23\textwidth]{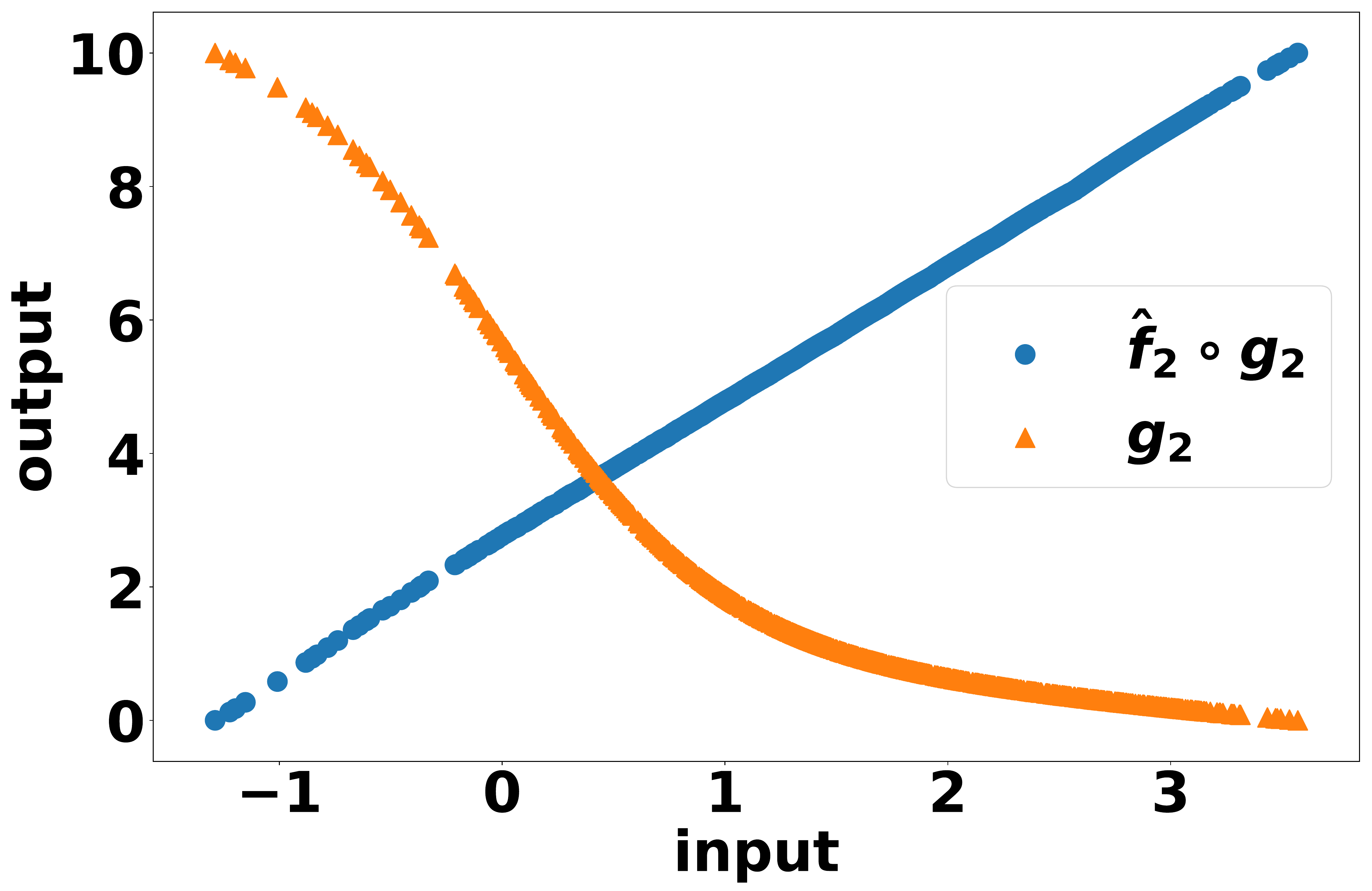}}
	\subfigure{\includegraphics[width=0.23\textwidth]{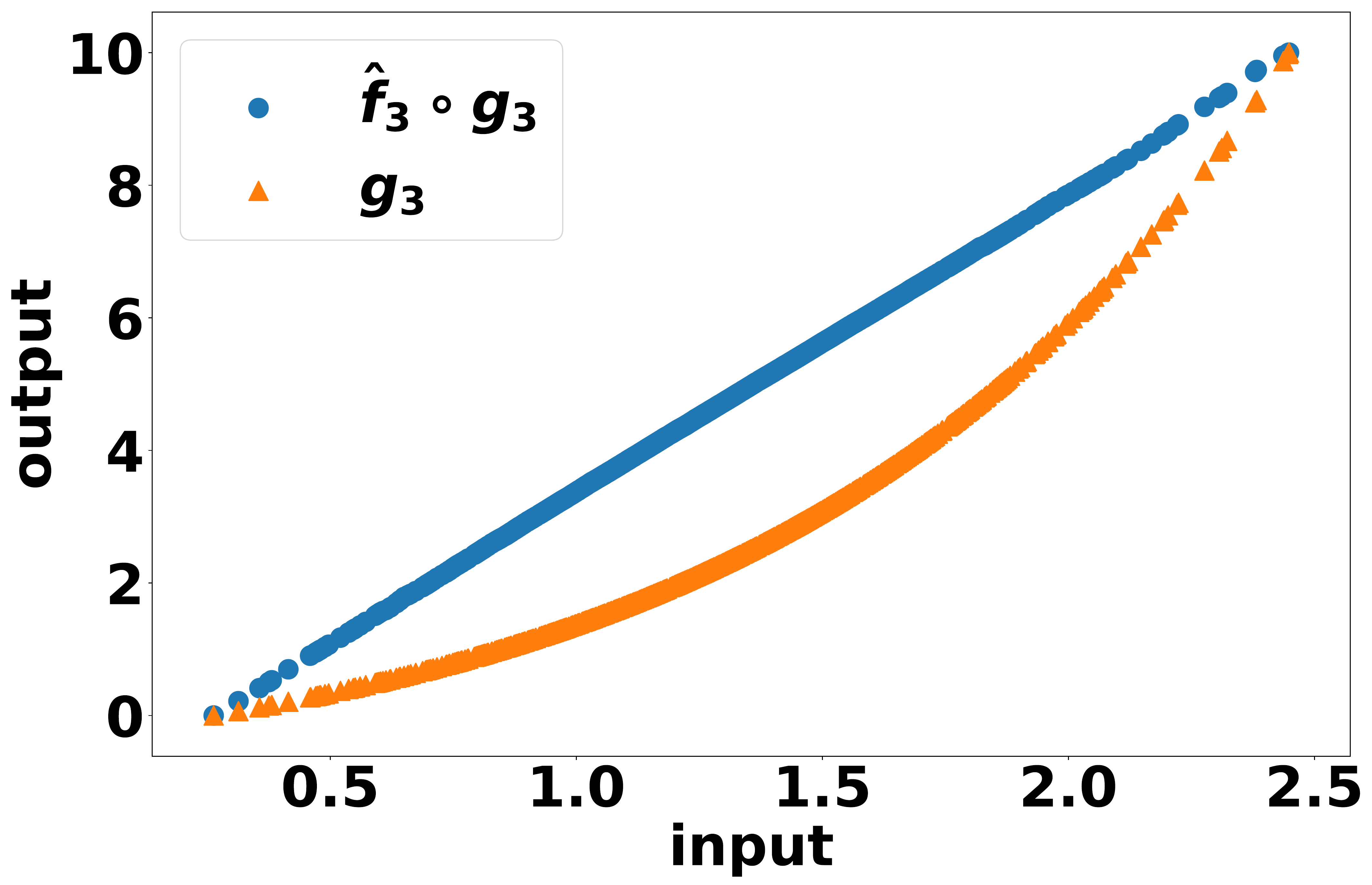}}
	\hspace{0.23\textwidth}}
	
	\colorbox{red!5}{\subfigure{\includegraphics[width=0.23\textwidth]{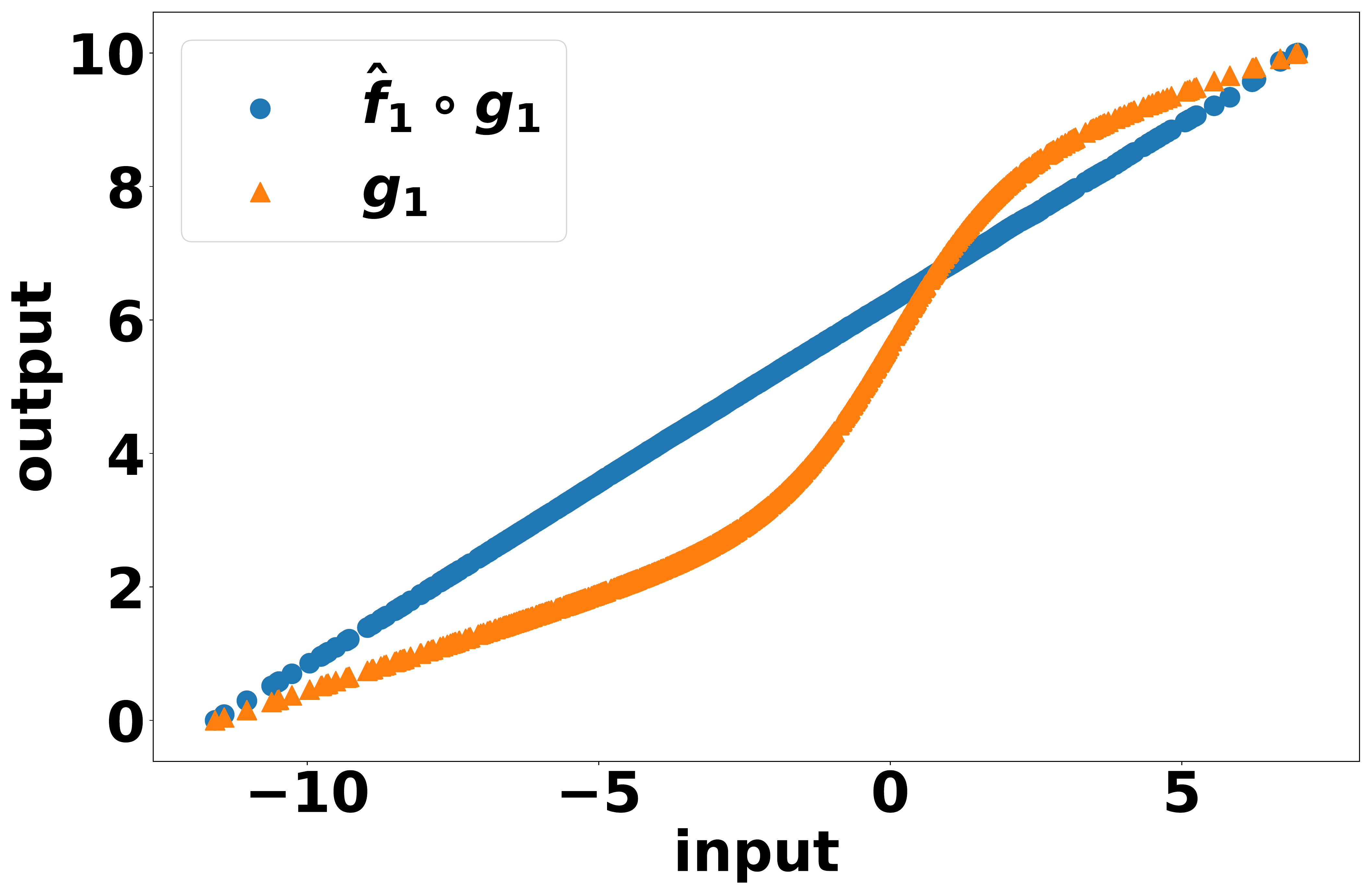}}
	\hspace{0.23\textwidth}
	\subfigure{\includegraphics[width=0.23\textwidth]{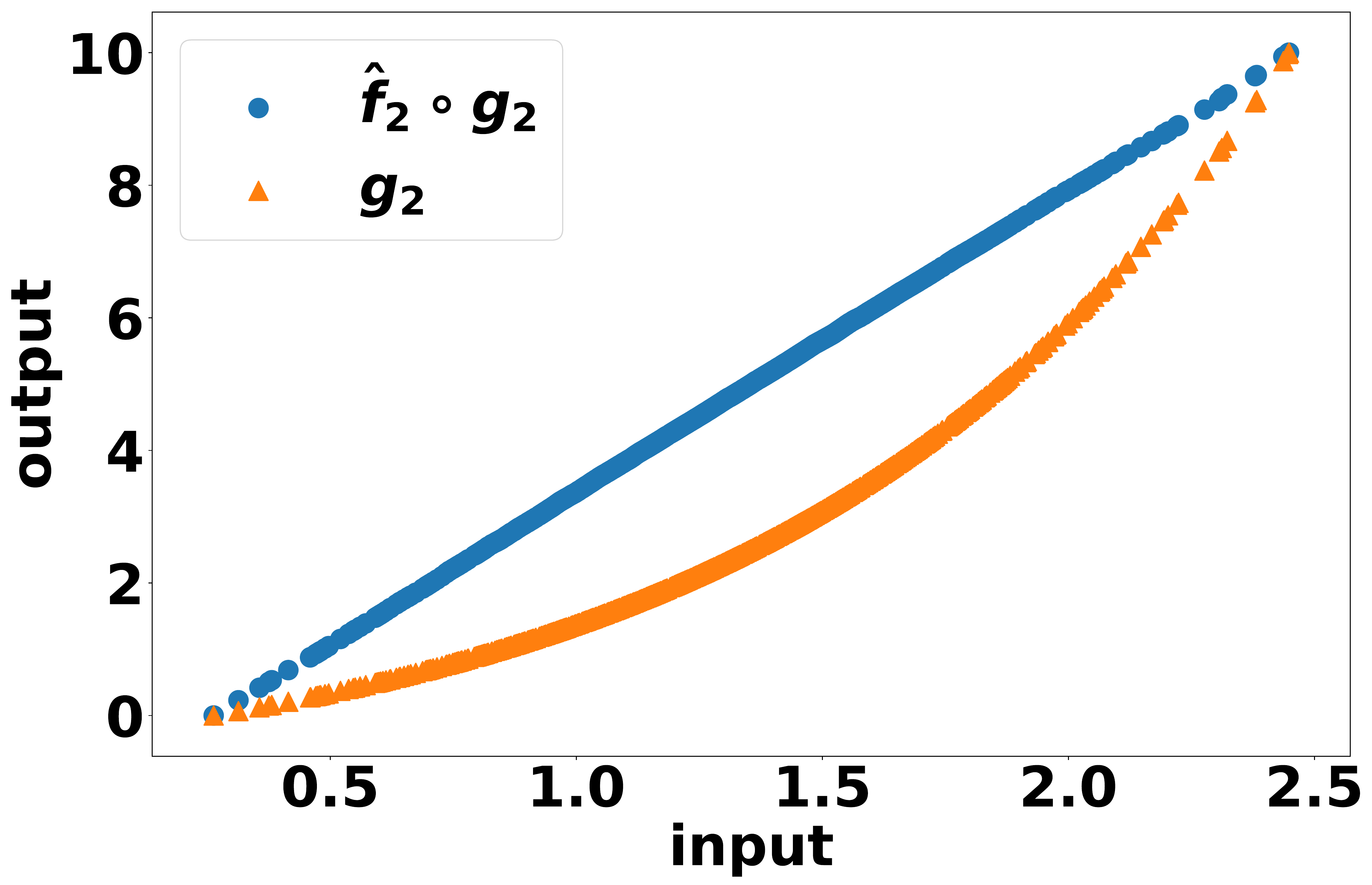}}
	\subfigure{\includegraphics[width=0.23\textwidth]{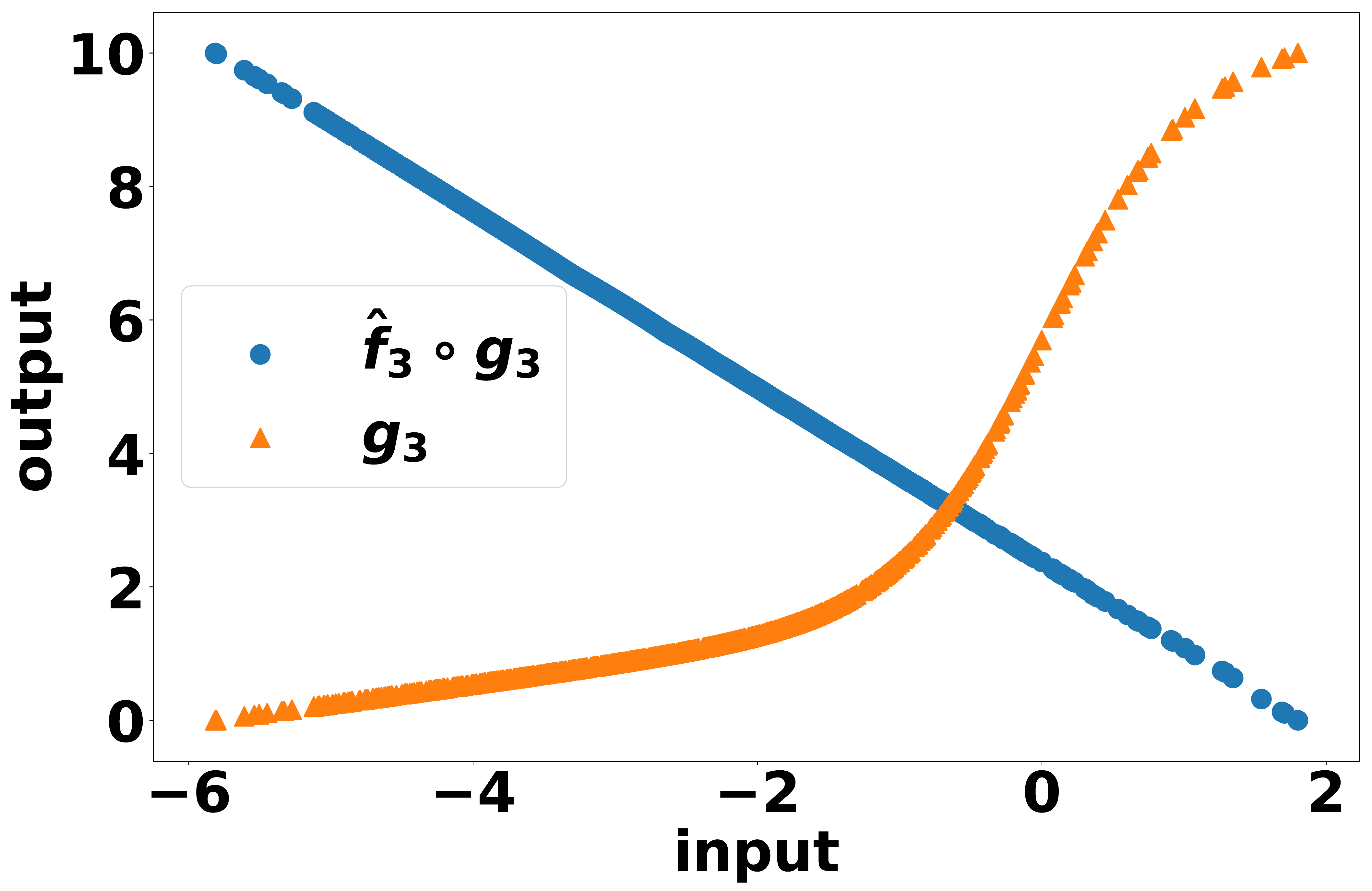}}}
	
	\caption{The results of the learned $\widehat{h}_m=\widehat{f}_m\circ g_m$'s under $(M,K)=(4,3)$. Top: using all 4 channels; Middle: using channel $x_1$, $x_2$, and $x_3$; Bottom: using channel $x_1$, $x_3$, and $x_4$.}\label{fig:necessary}
\end{figure*}

Fig.~\ref{fig:necessary} shows the {\it non-identifiability} of the proposed method when $M\leq  K(K-1)/2 $ is violated (cf. Theorem \ref{thm:necessary}). Specifically, we set $K=3$ and $M=4$. In this case, $M=4$ is larger than $\frac{K(K-1)}{2}=3$. 
The first row shows the result of applying \texttt{CNAE} on all of the four out channels, i.e., the entire column vector $\x_\ell$ for all $\ell$.
Apparently, the nonlinearity of every channel is not removed, which supports our claim in Theorem~\ref{thm:necessary} that $M\leq  K(K-1)/2 $ is not only sufficient, but also necessary if the criterion in \eqref{eq:population} is used. However, if one only selects three out of four channels and applies \texttt{CNAE} onto the selected channels [cf. Eq.~\ref{eq:selectchannel}], the nonlinear distortions can be successfully removed---see the second and the third rows in Fig.~\ref{fig:necessary}.

{To evaluate the impact of noise, we consider two different noisy models, i.e., \[\bm{x}_\ell=\bm{g}(\bm{As}_\ell)+\bm{v}_\ell,~  \bm{x}_\ell=\bm{g}(\bm{As}_\ell+\bm{v}_\ell),\] where $\bm{v}_\ell$ denotes the zero-mean white Gaussian noise. We define the signal-to-noise ratio (SNR) for the two models as 
\begin{align*}
    \text{SNR}_1 &= 10\log_{10}(\nicefrac{\sum_{\ell=1}^N \|\bm{g}(\bm{As}_\ell)\|_2^2}{\sum_{\ell=1}^N \|\bm{v}_\ell\|_2^2})\ \text{dB},\\
    \text{SNR}_2 &= 10\log_{10}(\nicefrac{\sum_{\ell=1}^N \|\bm{As}_\ell\|_2^2}{\sum_{\ell=1}^N \|\bm{v}_\ell\|_2^2})\ \text{dB},
\end{align*}
respectively. We test the noisy cases using $N=5,000$ samples under $K=4$ and $M=5$. In addition to the three nonlinear invertible functions used in the first simulation for generating data, $g_4(x)=-4\text{sigmoid}(x)-0.2x$ and $g_5(x)=5\text{tanh}(x)-0.3x$ are included. Other settings are the same as before.

Table \ref{tb:snr1} and Table \ref{tb:snr2} show the results under the two noisy models, respectively. One can see that the proposed method is robust to both noise models to a certain extent. 
In particular, when the SNR$_1$ and SNR$_2$ are larger than or equal to 20dB, the subspace distance between the
estimated ${\rm range}(\bm F^\T)$ and the ground truth ${\rm range}(\S^\T)$ is around 0.2. This is analogous to around only 10 degrees of misalignment between two vectors.
The best baseline's subspace distance is at least twice higher under the same noise level.
The MSEs of the estimated $\bm S$ also reflect a similar performance gap.

\begin{table}[t]
\centering
{
\caption{Subspace distance between $\bm F$ and $\bm S^\T$ and MSE between $\bm{S}$ and $\widehat{\bm{S}}$ under various $\text{SNR}_1$. Entries: dist/MSE.}
\begin{tabular}{c|cccc}\label{tb:snr1}
$\text{SNR}_1$     &  10 dB & 20 dB & 30 dB & 40 dB  \\ \hline \hline
\texttt{MVES}  & 0.997/$0.10$  & 0.998/$0.10$  & 0.998/$8.9e{-2}$ &  0.998/$0.10$ \\
\texttt{AE} & 0.591/$6.5e{-2}$  & 0.510/$6.2e{-2}$  & 0.470/$6.3e{-2}$ & 0.461/$6.6e{-2}$ \\
\texttt{NMFR} \cite{yang2020learning} & 0.701/$8.3e{-2}$ & 0.682/$8.1e{-2}$  & 0.395/$6.1e{-2}$ &  0.324/$2.2e{-2}$ \\
\texttt{CNAE} (Proposed) & 0.510/$6.7e{-2}$ & 0.218/$1.8e{-2}$  & 0.077/$3.9e{-3}$ & 0.055/$4.3e{-3}$
\end{tabular}
}
\end{table}

\begin{table}[t]
\centering
{
\caption{Subspace distance between $\bm F$ and $\bm S^\T$ and MSE between $\bm{S}$ and $\widehat{\bm{S}}$ under various $\text{SNR}_2$. Entries: dist/MSE.}
\begin{tabular}{c|cccc}\label{tb:snr2}
$\text{SNR}_2$     &  10 dB & 20 dB & 30 dB & 40 dB  \\ \hline \hline
\texttt{MVES}  & 0.998/$0.10$ & 0.997/$0.11$  & 0.997/$0.10$ & 0.998/$0.10$ \\
\texttt{AE} & 0.713/$8.6e{-2}$  &  0.501/$6.2e{-2}$  & 0.471/$5.9e{-2}$ & 0.469/$6.7e{-2}$ \\
\texttt{NMFR} \cite{yang2020learning} & 0.718/$8.9e{-2}$  & 0.438/$5.3e{-2}$  & 0.412/$1.6e{-2}$ &  0.267/$6.5e{-3}$  \\
\texttt{CNAE} (Proposed) & 0.482/$5.4e{-2}$ & 0.181/$9.1e{-3}$  & 0.067/$1.5e{-3}$  & 0.025/$7.4e{-4}$
\end{tabular}
}
\end{table}

}

\subsection{Real-Data Experiment: Nonlinear Hyperspectral Unmixing}
\begin{figure}[ht]
    \centering
    \subfigure{\includegraphics[clip, trim=4cm 1cm 4cm 0cm, width=0.35\textwidth]{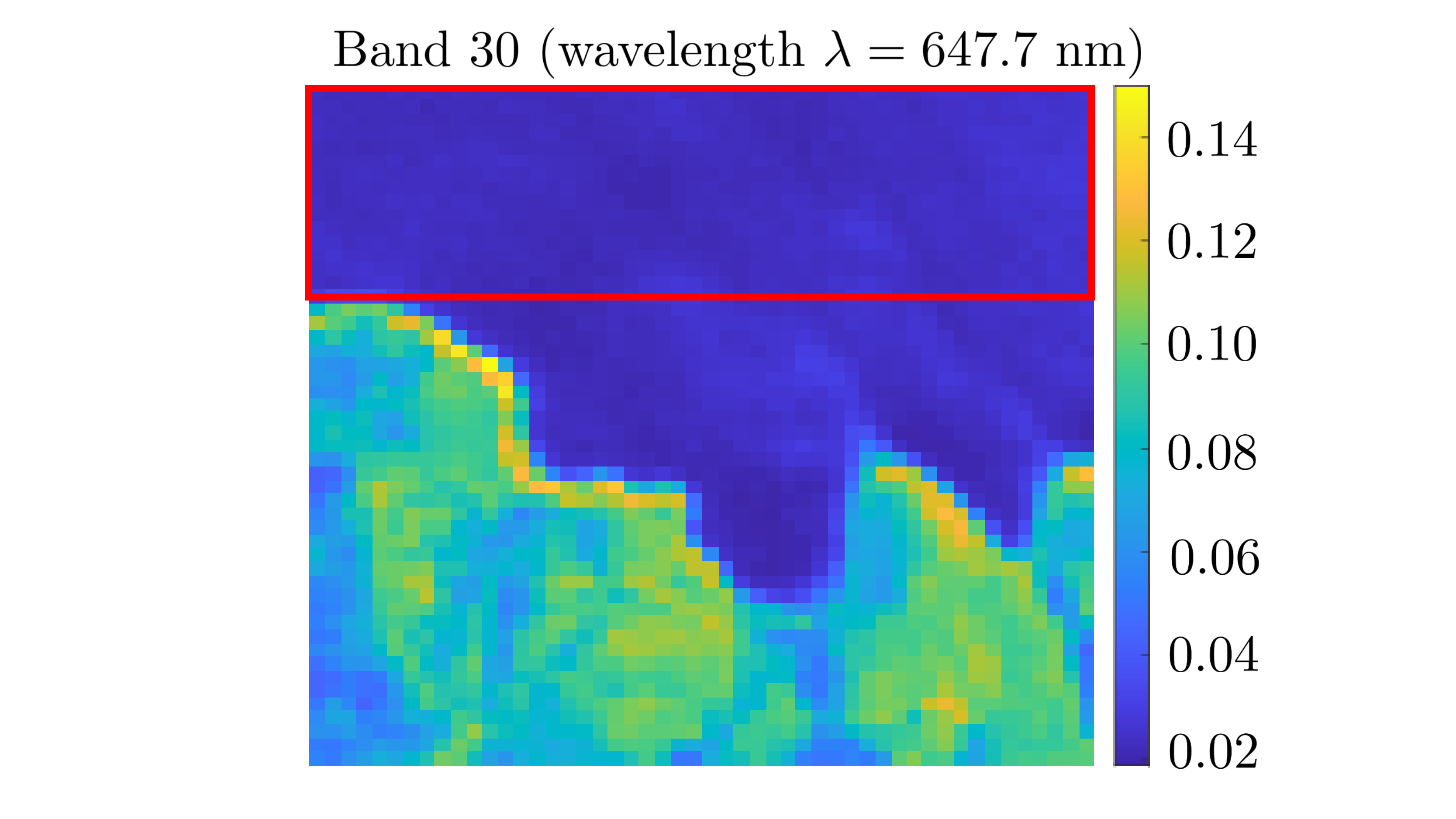}}
	\caption{The Moffett field data at band 30. The rectangle region corresponds to a lake. The water abundance is 1.0 in this region, while the abundances of other materials are zeros.}\label{fig:moffett}
\end{figure}

We consider the hyperspectral unmixing problem in the presence of nonlinearly mixed pixels. 
In this experiment, \texttt{CNAE} and \texttt{NMFR}
use the same neural network settings as before.
For \texttt{AE}, the encoder and decoder use identical (but mirrored) one-hidden-layer structure, where the number of neurons is $512$. 
{In addition to the baselines used in the simulations, we also consider another baseline that is specialized for hyperspectral unmixing, namely, {\it nonlinear hyperspectral unmixing based on deep autoencoder} (\texttt{NHUDAE}) \cite{wang2019nonlinear}. 
Our implementation follows the deep network architecture recommended in \cite{wang2019nonlinear}.
Note that the autoencoder used in \texttt{NHUDAE} is closer to a generic one, instead of the post-nonlinear model-driven structure used in \texttt{CNAE}.

Our experiment uses} the Moffett field data captured in California, which is known for the existence of nonlinearly distorted pixels \cite{fu2016robust,yang2020learning,dobigeon2014nonlinear}. The {considered} image has three major materials, {namely}, water, soil and vegetation.
The image consists of $50 \times 50$ pixels. Each pixel is represented by $M=198$ different spectral measurements. In other words, we have $\bm{x}_\ell\in\mathbb{R}^{198}$ and the sample size is $N=2,500$. Although no ground-truth is available, there are many pixels in the image that can be visually identified as purely water. {The water identity of these pixels is further verified by comparing with the previously recorded spectral signature of water in the literature \cite{li2005application,ji2015improving}}---and this information is used for evaluation; see Fig.~\ref{fig:moffett}, where a purely water region is highlighted using a red rectangle.

{Our evaluation strategy follows that in \cite{yang2020learning}.} To be specific, we compute the $\widehat{\bm s}_\ell$'s in this region using the methods under test, e.g., \texttt{CNAE}+\texttt{MVES}, and observe if the $\widehat{\bm s}_\ell$'s are unit vectors indicating that the region only contains one material (i.e., water).
Following this idea, we compute the average of $\widehat{s}_{k,\ell}$ over the pixels in the rectangle region. The ideal value should be 1 for the water and 0 for soil. The results are shown in Table~\ref{tab:weights}, where the $k$'s associated with `water' and `soil' are visually identified by plotting $\widehat{\S}_{k,:}$ and comparing to Fig.~\ref{fig:moffett}. Obviously, the SC-LMM based method \texttt{MVES} barely works on this data due to the existence of nonlinear distortions. {All of} the nonlinearity removal methods {except for \texttt{NHUDAE}} improve upon the result of \texttt{MVES}. In particular, both \texttt{NMFR} and \texttt{CNAE} perform better than the generic nonlinear learning tool \texttt{AE}, which shows that using latent component prior information in unsupervised nonlinear learning in an identifiability-guaranteed way is indeed helpful. Nevertheless, the proposed method outputs results that are the closest to the ideal values.

\begin{table}[ht]
\centering
{
\caption{Averaged estimated weights of the highlighted area in Fig.~\ref{fig:moffett} for the water and soil abundance map, respectively.}
\begin{tabular}{c|cc}\label{tab:weights}
material & \begin{tabular}[c]{@{}c@{}}water abundance\\ (true = 1.0)\end{tabular} & \begin{tabular}[c]{@{}c@{}}soil abundance\\ (true = 0.0)\end{tabular} \\ \hline\hline
\texttt{MVES}   & $0.657\pm0.003$  & $0.326\pm0.004$   \\ 
\texttt{AE}+\texttt{MVES}  & $0.741\pm0.145$ &  $0.202\pm0.160$ \\
{\texttt{NHUDAE} \cite{wang2019nonlinear}}  & $0.603\pm0.032$ &  $0.165\pm0.040$ \\
\texttt{NMFR} \cite{yang2020learning} +\texttt{MVES}  & $0.866\pm0.057$  &     $0.119\pm0.059$  \\ 
\texttt{CNAE} (Proposed)+\texttt{MVES} &     $0.943\pm0.030$    &      $0.025\pm0.029$  \\
\end{tabular}
}
\end{table}

\subsection{Real-Data Experiment: Image Representation Learning}
We also use the proposed \texttt{CNAE} method to serve as an unsupervised representation learner for image data. {Note that} the SC-LMM has been successful in image representation learning. {For example, the works in \cite{zhou2011minimum,lee1999learning,gillis2014robust} take the $\x_\ell=\A\s_\ell$ model for an image (e.g., a human face), where $\bm a_k$ for $k=1,\ldots,K$ are the constituents (e.g., nose, lips, and eyes) of a collection of such images, and $s_{k,\ell}$ is the weight of the $k$th constituent in image $\ell$; also see \cite{fu2018nonnegative} for illustration.}
Our hypothesis is that using SC-PNM is at least not harmful, and may exhibits some benefits since SC-PNM is a more general model relative to SC-LMM.

\begin{figure}[t!]
    \centering
	\subfigure{\includegraphics[clip, trim=6cm 6cm 6cm 5cm, width=0.48\textwidth]{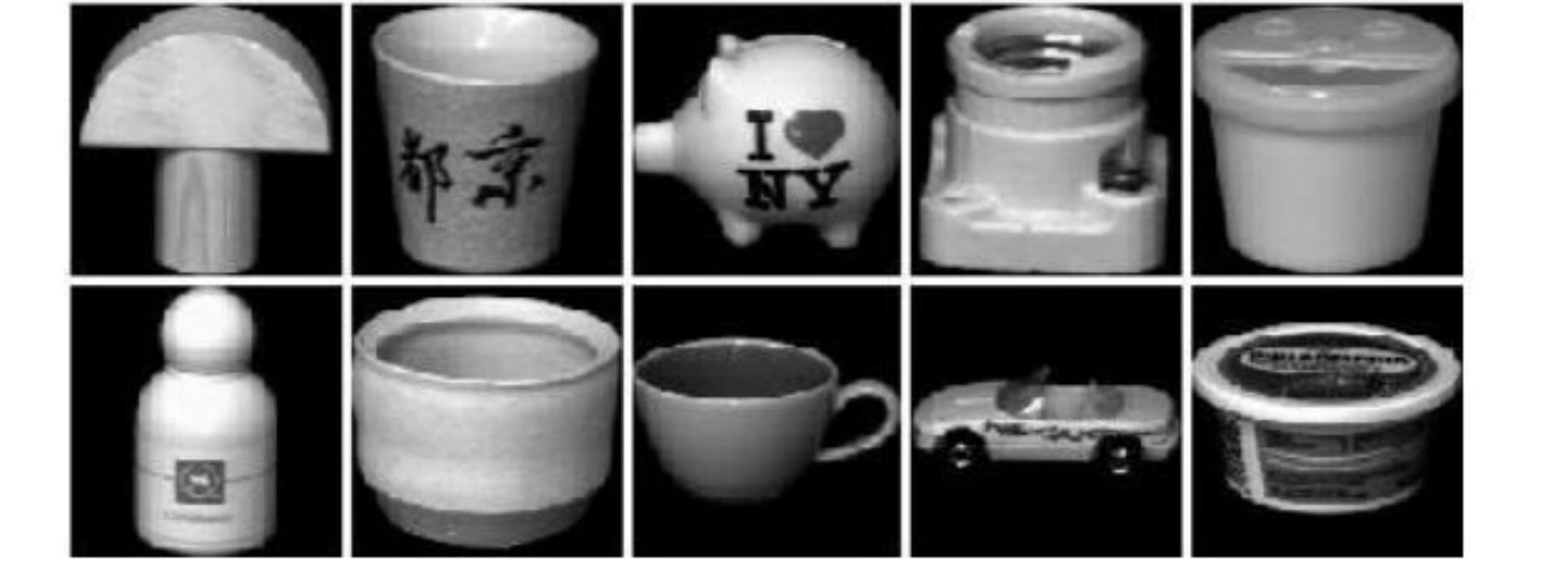}}
	\caption{Sample images from COIL dataset that are from 10 different classes.}\label{fig:coil}
\end{figure}

To verify our hypothesis, we test the methods on the COIL dataset \cite{sameer1996columbia}.
The COIL dataset has various classes of images, and each class admits 72 images. The images are represented by gray-level $32\times 32$ pixels. Some sample images are shown in Fig.~\ref{fig:coil}.

In our experiment, we test the algorithms using 5 and 10 classes of images, respectively.
Each class is split into training/validation/test sets following a $42/15/15$ ratio.
In order to show the effectiveness of different nonlinear dimensionality reduction (low-dimensional representation) methods (i.e., \texttt{CNAE}, \texttt{NMFR} and \texttt{AE}), we run $k$-means clustering algorithms followed the nonlinearity removal stage. 
We also use popular linear dimensionality reduction methods, i.e., \texttt{NMF} and \texttt{PCA}, to serve as our baselines.
For the $k$-means algorithm, we run clustering on the training set and then use the learned cluster centers to label the test set. 
For the proposed \texttt{CNAE}, the number of neurons for each $f_m$ is selected from $[32,64,128,256]$ and the latent dimension is chosen from $K\in[5,10,20,40,100,200]$.
These hyperparameters are selected using the validation set. 

\begin{table}[t!]
\centering
\caption{Kmeans clustering performance on COIL data.}
\begin{threeparttable}
\begin{tabular}{c|ccc|ccc}\label{tab:coil}
         & \multicolumn{3}{c|}{5 classes} & \multicolumn{3}{|c}{10 classes} \\ \hline
Methods  & ACC      & NMI      & ARI     & ACC      & NMI      & ARI      \\ \hline\hline
raw data &  0.829   &  0.851    &  0.746   &  0.698   & 0.799 & 0.611 \\ 
\texttt{PCA}      &  0.837 & 0.864  &  0.763  & 0.725 & 0.804 & 0.630 \\ 
\texttt{AE}      &  0.840    &  0.855 & 0.759 & 0.724 & 0.803 & 0.622 \\ 
\texttt{NMFR} \cite{yang2020learning}    & $\dagger$\tnote{1}   & $\dagger$       & $\dagger$       &     $\dagger$    &      $\dagger$    &    $\dagger$      \\ 
\texttt{MVES}     &  0.829 & 0.829 & 0.734 &    0.716  & 0.782 & 0.589  \\ 
\texttt{NMF}     &  0.728  & 0.767  & 0.603  & 0.590 &  0.737  & 0.480  \\
\texttt{CNAE} (Proposed)    & 0.896 & 0.888 & 0.824 & 0.753 & 0.823 & 0.662 \\ 
\end{tabular}
\begin{tablenotes}
    \item[1] Out of memory or too slow.
\end{tablenotes}
\end{threeparttable}
\end{table}

In Table~\ref{tab:coil},  the performance measures include clustering accuracy (ACC), normalized mutual information (NMI) and adjusted Rand index (ARI) \cite{yeung2001details}. ARI ranges
from $-1$ to $+1$, with one being the best and minus one the worst. ACC and NMI range from 0 to 1 with 1 being the best.
The results are averaged over 5 random selections of different classes. One can see that \texttt{CNAE} and \texttt{AE} in general outperform the baselines in terms of clustering performance. \texttt{CNAE} works even better compared to \texttt{AE}, perhaps because it enforces the sum-to-one condition in the nonlinearity reduction step that is consistent with the subsequent $k$-means model.

\section{Conclusions}
In this work, we addressed a number of challenges of the unsupervised SC-PNM learning problem. Specifically, we first advanced the understanding to the model identifiability of SC-PNM by offering largely relaxed identifiability conditions. In addition, we extended the population case based model identification analysis to the practical finite sample scenarios, and offered a number of sample complexity analyses. To our best knowledge, this is the first set of finite sample analyses for nonlinear mixture learning problems. Furthermore, we proposed a constrained neural autoencoder based formulation to realize the learning criterion, and showed that a set of feasible solutions of this formulation provably removes the unknown nonlinearity in SC-PNM. We also offered an
efficient Lagrangian multiplier based algorithmic framework to handle the optimization problem. We used synthetic and real-data experiments to show the effectiveness of our approach. Our approach is a combination of model-based mixture identification and data-driven nonlinear function learning. Our analyses may lead to better understanding for the effectiveness of neural autoencoders and more principled neural network design. 

{
There are also a number of limitations and potential future directions.
For example, under the current framework, every data dimension needs a neural autoencoder for nonlinearity removal, which may not be scalable (see Appendix~\ref{app:time_experiment} in the supplementary material). More judicious neural architecture design may be used to circumvent this challenge.
In addition, beyond the probability simplex-structured latent signals, it is also much meaningful to consider more general latent structures. Considering nonlinear distortions beyond the post-nonlinear case is another appealing direction---which may serve a wider range of applications.}

\section*{Acknowledgement}
Xiao Fu wishes to dedicate this work to Prof. Jos\'e Bioucas-Dias, whose work inspired him greatly in the last decade.

\ifplainver
    \section*{Appendix}
    \renewcommand{\thesubsection}{\Alph{subsection}}
\else
\appendices
\fi

\section{Proof of Theorem~\ref{thm:approx_err_finite_class}}
The proof is split into several steps.
\subsection{Useful Lemmas}
Note that there exists a constant $C\geq 0$ such that $(1-\bm 1^\T\bm f(\bm x))^2\leq C$ for all $f_m\in {\cal F}$ and $\x\in {\cal X}$ if Assumption~\ref{as:bound} holds.
We show the following lemma:
\begin{Lemma}\label{lem:hoeffding}
Assume that $N$ samples $\{\x_\ell\}_{\ell=1}^N$ are available, where $\x_\ell$'s are i.i.d. samples drawn from the domain ${\cal X}$ following distribution ${\cal D}$. Assume that Assumption~\ref{as:exact_finite} holds, and that we have $(1-\bm 1^\T\bm f(\bm x))^2\leq C$ for all $\x\in{\cal X}$.
Then, if Problem~\eqref{eq:sample} is solved with a solution $\widehat{\bm f}$ and
$$N\geq \frac{C^2\log(2d_{\cal F}/\delta)}{2\varepsilon^2},$$ the following holds with a probability at least $1-\delta$:
\begin{align}\label{eq:generalization_err_finite}
    \mathbb{E}\left[\left(1-\bm{1}^\top \widehat{\bm{f}}(\bm{x})\right)^2\right]&\leq \varepsilon,\quad \forall \x\in{\cal X},
\end{align}
\end{Lemma}

The proof can be found in Appendix~\ref{appdx:hoeffding}. 
We also show the following lemma:
\begin{Lemma}\label{lemma:bounded_4th}
    Assume that Assumption~\ref{as:bound} holds. Specifically, assume that the fourth-order derivatives of learned $\widehat{f}_i$ and the nonlinear distortion $g_i$ exist. In addition, $|f_i^{(n)}(\cdot)|\leq C_f$ and $|g_i^{(n)}(\cdot)|\leq C_g$, i.e., the $n$th-order derivatives are bounded for all $n\in\{1,\ldots,4\}$. Also assume that $\A$ has bounded elements, i.e., $|\A(i,j)|\leq C_a$. 
    Denote $${C}_{\phi}=16 M C_a^4 \left(C_f C_g^4+6 C_f C_g^3+3C_f C_g^2+4C_f C_g^2+C_f C_g\right).$$
    Then, the fourth-order derivative of $\phi(\s)=\bm{1}^\top\bm{h}(\bm{As})$ w.r.t. $s_i$ is bounded by 
    \begin{equation}
       \left|\frac{ \partial^4 \phi(\s)}{\partial s_i^4} \right|\leq { C}_{\phi}.
    \end{equation}
    In addition, the other cross-derivatives are also bounded by
    \begin{equation}
        \left| \frac{ \partial^4 \phi(\s)}{\partial s_i^3\partial s_j}  \right|\leq { C}_{\phi},\quad \left| \frac{ \partial^4 \phi(\s)}{\partial s_i^2\partial s_j^2}  \right|\leq { C}_{\phi}.
    \end{equation}
\end{Lemma}

The proof can be found in Appendix~\ref{appdx:bounded_4th}.

\subsection{Fitting Error Estimation}
With the lemmas above, we are ready to show the theorem. Our proof is constructive. 
Note that the fitting error estimation at any sample $\x_\ell\sim {\cal D}$ can be expressed as 
$
    (1-\bm{1}^\top \widehat{\bm{f}}(\bm{x}_\ell))^2 = \varepsilon_\ell.
$
If $\x_\ell$ is observed (i.e., $\x_\ell\in\{\x_1,\ldots,\x_N\}$) in the `training samples', then we have $\varepsilon_\ell=0$, otherwise we have  $\mathbb{E}\left[\varepsilon_\ell\right]\leq\varepsilon$ with high probability, as shown in Lemma~\ref{lem:hoeffding}. This is because $\varepsilon_\ell$ is a derived random variable of $\x_\ell$, and its expectation is, by the {\it fundamental theorem of expectation}, $\mathbb{E}[\varepsilon_\ell]=\mathbb{E}[ (1-\bm{1}^\top \widehat{\bm{f}}(\bm{x}_\ell))^2]$.

Using this notion, we will numerically `estimate' the second-order (cross-)derivatives of $\widehat{\bm{h}}=\widehat{\bm{f}}\circ\bm g$. 

\subsection{Estimating $\frac{\partial^2\phi(\s)}{\partial s_i^2}$} 
Recall that we have
\begin{align}\label{eq:derivative_base}
    \bm{1}^\top\widehat{\bm{h}}(\bm{As}_\ell)=1\pm \sqrt{\varepsilon_\ell}.
\end{align}

Define $\Delta \bm{s}_i=[0,\ldots,\Delta s_i, \ldots, 0, -\Delta s_i]^\top$
with 
\begin{equation}\label{eq:deltas_bound}
\Delta s_i\in {\cal S}_i = [0,\min\{s_{i,\ell},1-s_{i,\ell}\}),     
\end{equation}
for $i=1,\ldots,K-1$, and $\s_{\widehat{\ell}}=\s_\ell+\Delta \s_i$ and $\s_{\widetilde{\ell}}=\s_\ell-\Delta \s_i$. Note that $\s_\ell \pm \Delta \s_i \in {\rm int}\bm \varDelta_K$ still holds. Therefore, we have
\begin{equation}\label{eq:derivative_case1}
\begin{aligned}
    \bm{1}^\top\widehat{\bm{h}}(\bm{A}(\bm{s}_\ell+\Delta \bm{s}_i))&=1\pm \sqrt{\varepsilon_{\widehat{\ell}}},\\
    \bm{1}^\top\widehat{\bm{h}}(\bm{A}(\bm{s}_\ell-\Delta \bm{s}_i))&=1\pm \sqrt{\varepsilon_{\widetilde{\ell}}},
\end{aligned}
\end{equation}
where $\mathbb{E}[\varepsilon_{\widehat{\ell}}]\leq \varepsilon$ and $\mathbb{E}[\varepsilon_{\widetilde{\ell}}]\leq \varepsilon$ hold with high probability when $N$ is large.

For any continuous function $\omega(z)$ that admits non-vanishing 4th order derivatives, the second order derivative at $z$ can be estimated as follows \cite{morken2018numerical}:
\begin{align*}
    \omega''(z) =& \frac{\omega(z+\Delta z)-2\omega(z)+\omega(z-\Delta z)}{\Delta z^2}-\frac{\Delta z^2}{12}\omega^{(4)}(\xi),
\end{align*}
where $\xi\in(z-\Delta z,z+\Delta z)$.

Following this definition, one can see that
\begin{align*}
\frac{ \partial^2 \phi(\s)}{\partial s_i^2}
    &=\frac{\pm\sqrt{\varepsilon_{\widehat{\ell}}}\mp 2\sqrt{\varepsilon_\ell}\pm \sqrt{\varepsilon_{\widetilde{\ell}}}}{\Delta s_i^2}
    -\frac{\Delta s_i^2}{12}\phi^{(4)}(\bm \xi_i),
\end{align*}
where $\bm \xi_i\in(\bm s_{\widetilde{\ell}},\bm s_{\widehat{\ell}})$.
Consequently, we have the following inequalities:
\begin{align*}
    \left|\sum_{m=1}^M (a_{m,i}-a_{m,K})^2\widehat{h}''_m(\bm{As}_\ell)\right|
    &=\left|\frac{\pm\sqrt{\varepsilon_{\widehat{\ell}}}\mp 2\sqrt{\varepsilon_\ell}\pm \sqrt{\varepsilon_{\widetilde{\ell}}}}{\Delta s_i^2}-\frac{\Delta s_i^2}{12}\phi^{(4)}(\bm \xi_i)\right|\\
    &\leq \frac{\sqrt{\varepsilon_{\widehat{\ell}}}+2\sqrt{\varepsilon_\ell}+\sqrt{\varepsilon_{\widetilde{\ell}}}}{\Delta s_i^2}+\frac{\Delta s_i^2}{12}\left|\phi^{(4)}(\bm \xi_i)\right|.
\end{align*}

By taking expectation, we have the following holds with probability at least $1-\delta$:
\begin{align}\label{eq:hpp_bound}
    \mathbb{E}\left[\left|\sum_{m=1}^M (a_{m,i}-a_{m,K})^2\widehat{h}''_m(\bm{As}_\ell)\right|\right] &\leq \frac{\mathbb{E}[\sqrt{\varepsilon_{\widehat{\ell}}}]+2\mathbb{E}[\sqrt{\varepsilon_\ell}]+\mathbb{E}[\sqrt{\varepsilon_{\widetilde{\ell}}}]}{\Delta s_i^2}+\frac{\Delta s_i^2}{12}\left|\phi^{(4)}(\bm \xi_i)\right| \nonumber\\
    &\leq\frac{4\sqrt{\varepsilon}}{\Delta s_i^2}+\frac{\left|\phi^{(4)}(\bm \xi_i)\right|\Delta s_i^2}{12},
\end{align}
where the second inequality is by the Jensen's inequality
\[ \mathbb{E}[\sqrt{\varepsilon_\ell}]\leq \sqrt{\mathbb{E}[\varepsilon_\ell]}\leq\sqrt{\varepsilon}, \]
which holds by the concavity of $\sqrt{x}$ when $x\geq 0$.

Note that the bound in \eqref{eq:hpp_bound} holds for all $\Delta s_i$ that satisfy \eqref{eq:deltas_bound}. We are interested in finding the best upper bound, i.e.,
\begin{align}\label{eq:min_best_bound}
    \inf_{\Delta s_i \in {\cal S}_i } \frac{4\sqrt{\varepsilon}}{\Delta s_i^2}+\frac{\left|\phi^{(4)}(\bm \xi)\right|}{12} \Delta s_i^2.
\end{align}

Note that the function in \eqref{eq:min_best_bound} is convex and smooth when $\Delta s_i\in{\cal S}_i$.
Hence, it is straightforward to show that the infimum is attained by either the minimizer of the convex function or the boundary of ${\cal S}_i$, i.e.,:
$$\Delta s_i^\star\in \left\{\left(\frac{48\sqrt{\varepsilon}}{\left|\phi^{(4)}(\bm \xi_i)\right|}\right)^{1/4},\min\{s_{i,\ell},1-s_{i,\ell}\}\right\},$$ which gives the minimum as follows:
\begin{align}\label{eq:varepsilon_bound}
   \inf_{\Delta s_i} \frac{4\sqrt{\varepsilon}}{\Delta s_i^2}+\frac{\left|\phi^{(4)}(\bm \xi_i)\right|}{12} \Delta s_i^2
   \leq \min\left\{ \frac{2\sqrt{3\left|\phi^{(4)}(\bm \xi_i)\right|}\varepsilon^{1/4}}{3},\frac{4\sqrt{\varepsilon}}{\kappa_i^2}+\frac{\left|\phi^{(4)}(\bm \xi_i)\right|}{12} \kappa_i^2\right\},
\end{align}
where $\kappa_i=\min\{ s_{i,\ell},1-s_{i,\ell} \}$.
Note that we always have
$ \kappa_i \geq \gamma$.
Hence, if 
$ \left(\frac{48\sqrt{\varepsilon}}{\left|\phi^{(4)}(\bm \xi_i)\right|}\right)^{1/4} \leq \gamma, $
we can simplify the bound.
This means that, if we have
$
    \gamma \geq  \left(\frac{48\sqrt{\varepsilon}}{C_\phi}\right)^{1/4},$
then, \eqref{eq:varepsilon_bound} can be bounded by
\begin{equation*}
  \left|\mathbb{E}\left[\sum_{m=1}^M (a_{m,i}-a_{m,K})^2\widehat{h}''_m(\bm{As})\right]\right| \leq \frac{2\sqrt{3\left|\phi^{(4)}(\bm \xi_i)\right|}\varepsilon^{1/4}}{3}.
\end{equation*}
Given that $N$ is fixed, one may pick $\varepsilon = \sqrt{ \frac{C^2\log(2d_{\cal F}/\delta)}{2N}}$, which gives the conclusion that
if $\gamma \geq \left(\frac{48}{C_\phi}\right)^{1/4} \left(\frac{C^2\log\left(\frac{2d_{\cal F}}{\delta}\right)}{2N}\right)^{1/16},$ we have
\begin{equation}\label{ineq:second_derivative_bound}
\begin{aligned}
  \left|\mathbb{E}\left[\sum_{m=1}^M (a_{m,i}-a_{m,K})^2\widehat{h}''_m(\bm{As})\right]\right|
  \leq \frac{2\sqrt{3C_\phi}\left(\frac{C^2\log\left(\frac{2d_{\cal F}}{\delta}\right)}{2N}\right)^{1/8}}{3}.
  \end{aligned}
\end{equation}

\subsection{Estimating $\frac{\partial^2\phi(\s)}{\partial s_is_j}$}
In this subsection, we show a similar bound as in \eqref{ineq:second_derivative_bound} for the cross-derivatives $\frac{\partial^2\phi(\s)}{\partial s_is_j}$. The detailed proof is relegated to  Appendix~\ref{appdx:cross_derivative}.  
The bound for cross-derivatives is as follows.

Given a fixed $N$, pick $\varepsilon = \sqrt{ \frac{C^2\log(2d_{\cal F}/\delta)}{2N}}$. Then,
if $\gamma \geq (\frac{3}{C_\phi})^{1/4} \left(\nicefrac{C^2\log\left(\frac{2d_{\cal F}}{\delta}\right)}{2N}\right)^{1/16},$ the following holds for $i\neq j$:
\begin{equation}\label{ineq:cross_derivative_bound}
\begin{aligned}
    \left|\mathbb{E}\left[\sum_{m=1}^M (a_{m,i}-a_{m,K})(a_{m,j}-a_{m,K})\widehat{h}''_m(\bm{As}_\ell)\right]\right|
    \leq \frac{\sqrt{3 C_\phi}\left(\frac{C^2\log\left(\frac{2d_{\cal F}}{\delta}\right)}{2N}\right)^{1/8}}{6}.
\end{aligned}
\end{equation}

\subsection{Putting Together}
For both derivative estimations, by \eqref{ineq:second_derivative_bound} and \eqref{ineq:cross_derivative_bound}, we have the following bound since $\|\cdot\|_2$ is upper bounded by $\|\cdot\|_1$:
\begin{align*}
    \mathbb{E}\left[\left\|{\bm G}\widehat{\bm{h}}''(\A\s)\right\|_2^2\right] = O\left(C_\phi\sqrt{C}  \left(\frac{\log\left(\frac{2d_{\cal F}}{\delta}\right)}{N}\right)^{1/4} \right),
\end{align*}
which can be further simplified as:
\begin{align*}
    \mathbb{E}\left[\left\|\widehat{\bm{h}}''(\A\s)\right\|_2^2\right]&= O\left(\frac{C_\phi\sqrt{C}}{\sigma_{\min}^2({\bm G})}  \left(\frac{\log\left(\frac{2d_{\cal F}}{\delta}\right)}{N}\right)^{1/4} \right),
\end{align*}
where the above is because the expectation is taken over $\x\sim{\cal D}$ and ${\bm G}$ is not a function of $\x$.

\section{Proof of Theorem~\ref{thm:approx_err_nn_class}}\label{appdx:approx_err_nn_class}
The key of the proof is to establish a similar bound as in \eqref{eq:generalization_err_finite} for the NN-approximated case.
We start with the following lemma:
\begin{Lemma}\label{lem:rc_nn}
    Consider the function class
    \begin{equation}\label{eq:loss}
        {\cal H}=\left\{ l(\bm x) \left|  l(\bm x)=\left(1-\sum_{m=1}^M f_m(x_m)\right)^2  \right.\right\},
    \end{equation}
    where each $f_m(\cdot):\mathbb{R}\rightarrow \mathbb{R}\in \mathcal{F}$ which is defined in Assumption~\ref{as:nn}.
    Assume {that $\bm x_\ell$ for $\ell\in[N]$ are i.i.d. samples from ${\cal X}$ according to a certain distribution ${\cal D}$.}
    Then, the {\it Rademacher complexity} of class ${\cal H}$ is bounded by ${\mathfrak{R}_N}(\mathcal{H})\leq2M(MzB^2C_x+1){\mathfrak{R}_N}(\mathcal{F})$, where the bound of class $\mathcal{F}$ is
    $
        {\mathfrak{R}_N}(\mathcal{F})\leq 2zB^2C_x\sqrt{\frac{R}{N}}$, and $C_x$ is the {upper bound of $|x_m|$ for all $\bm{x}\in{\cal X}$.}
\end{Lemma}

The detailed proof of this lemma is in Appendix~\ref{appdx:rc_nn}.
The notion of {\it Rademacher complexity} (see Appendix~\ref{app:tools}) is commonly used in statistical machine learning to come up with generalization error bounds in terms of sample complexity, typically in supervised learning (see details in \cite{bartlett2002rademacher}). It also proves handy in proving sample complexity of our unsupervised problem.

Following Lemma~\ref{lem:rc_nn} and \cite[Theorem 3.3]{mohri2018foundations}
    we have that with probability of at least $1-\delta$, the following holds:
        \begin{align*}
            \mathbb{E}&[(1-\bm{1}^\top \widehat{\bm{f}}(\bm{x}))^2]\leq 
            \frac{1}{N}\sum_{\ell=1}^N \left(1-\bm{1}^\top \widehat{\bm{f}}(\bm{x}_\ell)\right)^2
            +2{\mathfrak{R}_N}(\mathcal{H})+(MzB^2C_x+1)^2\sqrt{\frac{\log(1/\delta)}{2N}}\\
            &\leq \frac{1}{N}\sum_{\ell=1}^N \left(1-\bm{1}^\top \widehat{\bm{f}}(\bm{x}_\ell)\right)^2+8MzB^2C_x (MzB^2C_x+1) \sqrt{\frac{R}{N}}
            +(MzB^2C_x+1)^2\sqrt{\frac{\log(1/\delta)}{2N}}.
        \end{align*}
Since we have assumed ${\cal G}^{-1}\subseteq {\cal F}$, the empirical error should be zero by learning $\widehat{\bm{f}}$ as an inverse of $\bm{g}$, i.e., $|1-\bm{1}^\top \widehat{\bm{f}}(\bm{x}_\ell)|=0$, which leads to the following:
\begin{align*}
    \mathbb{E}[(1-\bm{1}^\top \widehat{\bm{f}}(\bm{x}))^2]&\leq 8MzB^2C_x (MzB^2C_x+1) \sqrt{\frac{R}{N}}
    +(MzB^2C_x+1)^2\sqrt{\frac{\log(1/\delta)}{2N}}.
\end{align*}

By following the similar proof of Theorem \ref{thm:approx_err_finite_class} for the second-order derivative estimation and let
\begin{align}\label{eq:eps_nn}
    \varepsilon &= 8MzB^2C_x (MzB^2C_x+1) \sqrt{\frac{R}{N}}
    +(MzB^2C_x+1)^2\sqrt{\frac{\log(1/\delta)}{2N}},
\end{align}

Thus, when $\gamma = \Omega\left(\left(\frac{MzB^2C_x}{C_\phi}\right)^{1/4}\left(\sqrt{\frac{R}{N}}+\sqrt{\frac{\log(1/\delta)}{N}}\right)^{1/8}\right)$, we have
\begin{align*}
    \mathbb{E}\left[\left\|\widehat{\bm{h}}''(\A\s)\right\|_2^2\right]= {O}\left( \frac{C_\phi MzB^2C_x\left(\sqrt{R}+\sqrt{\log(1/\delta)}\right)^{1/2}}{\sigma_{\min}^2({\bm G}) N^{1/4}}  \right),
\end{align*}
for any $\bm s$ such that $1-\gamma\geq s_{i}\geq \gamma>0$ for all $i\in[K]$.
This completes the proof for the neural network case.

\putbib[bu1]
\end{bibunit}

\clearpage

\begin{bibunit}[IEEEtran]
\begin{center}
    {\bf \Large Supplemental Material of ``Identifiability-Guaranteed Simplex-Structured Post-Nonlinear Mixture Learning via Autoencoder'' by Qi Lyu and Xiao Fu}
\end{center}

\section{Proof of Theorem~\ref{thm:necessary}}
In \eqref{eq:linear_system}, $\bm{G}$ has a size of $\frac{K(K-1)}{2}\times M$. Note that when $M>\frac{K(K-1)}{2}$,  the null space of $\bm G$ is nontrivial. 
Hence, we have at least one non-zero solution $\bm{h}''=[\iota_1, \cdots, \iota_M]^\top$ such that $\bm{Gh}''=\bm{0}$ holds.
Our idea is to construct an invertible $h_m(\cdot)=\widehat{f}_m\circ g_m(\cdot)$'s where $\widehat{f}_m$'s are from a solution of \eqref{eq:population}, while some $h_m$'s are not affine. Note that since $\widehat{f}_m$ and $g_m$ are invertible, the constructed $h_m$ has to be invertible.

Assume that $\iota_m\neq 0$, we construct the corresponding $h_m$'s as follows:
\begin{align*}
    h_m(y_m) &= \iota_m (y_m+c_m)^2 +\beta_m
    =\iota_m y^2_m + 2\iota_m c_m y_m+ \iota_m c^2_m + \beta_m,
\end{align*}
where $c_m$ and $\beta_m$ are certain constants, $y_m=\sum_{k=1}^{K-1} b_{m,k}s_k+a_{m,K}$ (by denoting $s_K=1-s_1-\cdots-s_{K-1}$) with $b_{m,k}$ defined in Theorem~\ref{thm:main_population}.
When $\iota_m\neq 0$, we require each $c_m\leq\min_k a_{m,k}$ or $c_m\geq\max_k a_{m,k}$ for invertibility of $h_m$. 
For those $h_{m}(\cdot)$'s with $\iota_{m}=0$, we construct
$h_{m}(y_{m}) =  \omega_{m} y_m +\beta_m,$
where $\omega_{m}\neq 0$.
To have $\bm h''\in {\sf null}(\bm G)$, one only needs $\bm{\iota}=[\iota_1,\ldots,\iota_M]^\T\in {\sf null}(\bm G)$.

We now show that the above construction can always make $\bm 1^\T\widehat{\bm f}(\bm x)=1$ satisfied. Consider $\sum_{m=1}^M h_m(\cdot)$ over its domain. First, the quadratic terms can be expressed as follows:
\begin{align*}
    &\quad\sum_{m=1}^{M} \iota_m y_m^2=\sum_{m=1}^{M} \iota_m \left(\sum_{k=1}^{K-1} b_{m,k}s_k+a_{m,K}\right)^2\\
    &=\sum_{m=1}^{M} \iota_m \left(\left(\sum_{k=1}^{K-1} b_{m,k}s_k\right)^2 + 2a_{m,K}\left(\sum_{k=1}^{K-1} b_{m,k}s_k\right)+a^2_{m,K} \right)\\
    &=\sum_{m=1}^{M} \iota_m a_{m,K} \left(2y_m-a_{m,K}\right),
\end{align*}
where the second-order term is zero because it can be written as $\bm 1^\T \bm \varSigma \bm G\bm h''$ and $\bm G\bm h''=\bm 0$, in which $\bm \varSigma ={\rm Diag}(s_1^2,\ldots,s_{K-1}^2,s_1s_2,\ldots,s_{K-2}s_{K-1})$ has full rank for $\bm s\in {\rm int}\bm \varDelta_K$.

Let us define ${\cal M}_1=\{i_1,\ldots,i_{M_1}\}$ and ${\cal M}_2=\{j_1,\ldots,j_{M_2}\}$ as the index sets of the nonzero and zero $\iota_m$'s, respectively (not that ${\cal M}_1\cup{\cal M}_2=[M]$ and ${\cal M}_1\cap {\cal M}_2=\emptyset$). 
Our construction has to respect the constraints of \eqref{eq:population}, i.e.,
\begin{align}
    1&=\bm{1}^\T \bm{h}(\bm{As})=\sum_{m\in {\cal M}_1} [(2\iota_m a_{m,K}+2\iota_m c_m)y_m+\beta_m 
    -\iota_m a^2_{m,K}+c_m^2\iota_m] + \sum_{m\in {\cal M}_2} (\omega_{m} y_m +\beta_m). \label{eq:Az1}
\end{align}

By selecting $\beta_m=\iota_m a^2_{m,K}-c_m^2\iota_m$ for all $m$, the constant term of the above becomes zero. Then, we have
\begin{align}
    \sum_{m\in {\cal M}_1} (2\iota_m a_{m,K}+2\iota_m c_m)y_m+ \sum_{m\in {\cal M}_2} \omega_{m} y_m = 1,
\end{align}
which can be expressed as:
\begin{align}
    \bm{1}^\top\bm{\Sigma}'\bm{As}=1,
\end{align}
where 
\begin{align*}
    \bm{\Sigma}'=&\text{Diag}\left(2\iota_{i_1} a_{i_1,K}+2\iota_{i_1} c_{i_1},\ldots, 
    2\iota_{i_{M_1}} a_{i_{M_1},K}+2\iota_{i_{M_1}} c_{i_{M_1}}, \omega_{j_1}, \ldots, \omega_{j_{M_2}}\right),
\end{align*}
in which, without loss of generality (w.o.l.g.), we have assumed that $i_1\leq \ldots i_{M_1} \leq j_1\leq \ldots \leq j_{M_2}$. 
Since $\bm 1^\T\bm s=1$, our construction only needs to satisfy:
\begin{align}\label{eq:lsys}
    \bm{1}^\top\bm{\Sigma}'\bm{A}=\bm{1}^\top \iff \bm{A}^\top \bm z = \bm{1},
\end{align}
where $\bm z=  [2\iota_{i_1} ( a_{i_1,K}+ c_{i_1}),\ldots,2\iota_{i_{M_1}}( a_{i_{M_1},K}+ c_{i_{M_1}}),
\omega_{j_1},\ldots,\omega_{j_{M_2}}]^\T.$
By Lemma~\ref{lem:feas}, a dense solution of \eqref{eq:lsys} exists. Denote the dense solution as $\bm \tau=[\tau_1,\ldots,\tau_M]^\T$.
Then, we use the following construction to make $\bm z=\bm \tau$:
\begin{align*}
    2\iota_m( a_{m,K}+ c_m) = {\tau_m}, \text{ for } m \text{ with } \iota_m\neq 0,\\
    \omega_m = \tau_m, \text{ for } m \text{ with } \iota_m= 0,
\end{align*}
which gives us
\begin{align*}
    c_m &= \frac{\tau_m}{2\iota_m}-a_{m,K}, \text{ for } m \text{ with } \iota_m\neq 0,\\
    \omega_m &= \tau_m, \text{ for } m \text{ with } \iota_m= 0.
\end{align*}

It is obvious that if $\iota_m$ has a small enough magnitude, $c_m$ has a sufficiently large magnitude. Note that each $\iota_m$ can be scaled arbitrarily without violating $\bm G\bm h''=\bm 0$. As a result, it is guaranteed that there are $c_m$'s such that either $c_m\leq\min_k a_{m,k}$ or $c_m\geq\max_k a_{m,k}$ is satisfied. In addition, the affine $h_m(\cdot)$'s with nonzero $\omega_m=\tau_m$ are also invertible. 
Since $\widehat{f}_m$ can be any continuous invertible function, there always exists $\widehat{f}_m$ such that $h_m=\widehat{f}_m\circ g_m$ (e.g., by letting $\widehat{f}_m=h_m\circ g_m^{-1}$---which is invertible).   
Hence, by construction, we have shown that there exists an $\widehat{\bm f}$ that is a solution of \eqref{eq:population} while the elements of the corresponding $\bm h=\widehat{\bm f}\circ \bm g$ are not all affine.

\section{Proof of Theorem~\ref{thm:function_mismatch}}\label{appdx:function_mismatch}

\subsection{Finite Function Class}
By the assumption that for any $\widehat{f}_m\in{\cal F}$ we have:
    \begin{equation*}
        \sup_{\x\in{\cal X}}~|\widehat{f}_m(x_m)-\widehat{u}_m(x_m)|<\nu,~\forall m\in[M],~\exists \widehat{u}_m\in{\cal G}^{-1}.
    \end{equation*}

Hence, one can bound $1-\bm 1^\T\widehat{\bm f}(\bm{x}_\ell)$ for all $\ell$ by introducing $\widehat{\bm u}$, i.e., an inverse of $\bm{g}$ as defined in Definition~\ref{def:g_inverse}:
\begin{align*}
    \left|1-\bm 1^\T\widehat{\bm f}(\bm{x}_\ell)\right|&=\left|1-\bm 1^\T\widehat{\bm u}(\bm{x}_\ell)+\bm 1^\T\widehat{\bm u}(\bm{x}_\ell)-\bm 1^\T\widehat{\bm f}(\bm{x}_\ell)\right|\\
    &\leq \left|1-\bm 1^\T\widehat{\bm u}(\bm{x}_\ell)\right| + \|\bm{1}\|_1 \left\|\widehat{\bm u}(\bm{x}_\ell)-\widehat{\bm f}(\bm{x}_\ell)\right\|_\infty\\
    &= M\nu,
\end{align*}
where the second inequality is by the triangle inequality and H{\"o}lder's inequality, and the last equality is because $\widehat{\bm{u}}$ is an inverse of $\bm{g}$ such that we have $1-\bm 1^\T\widehat{\bm u}(\bm{x}_\ell)=0$ for all $\ell=[N]$.

Then, we have:
\begin{align*}
    \left(1-\bm 1^\T\widehat{\bm f}(\bm x)\right)^2\leq M^2\nu^2,
\end{align*}
which implies an approximate feasible solution $\widehat{\bm{f}}$.

Following Lemma \ref{lem:hoeffding} we have:
\begin{align*}
    {\sf Pr}\left[ \{\bm x_\ell\}_{\ell=1}^N: \forall \widehat{\bm f}: \left|{\cal P}_{\cal D}(\widehat{\bm f})-{\cal P}_N(\widehat{\bm f})\right|\leq\varepsilon \right]
    \geq 1-2d_{\cal F}\exp\left(\frac{-2N\varepsilon^2}{C^2}\right).
\end{align*}

By the definition ${\cal P}_{N}(\widehat{\bm f})$, we have
\begin{align*}
    {\cal P}_{N}(\widehat{\bm f}) &= \frac{1}{N}\sum_{\ell=1}^N\left(1-\bm 1^\T\widehat{\bm f}(\x_\ell)\right)^2\leq M^2\nu^2.
\end{align*}

Therefore, the above means that we have the following holds with probability of at least $1-\delta$:
\begin{align*}
    \mathbb{E}\left[\left(1-\bm 1^\T\widehat{\bm f}(\bm x)\right)^2\right]\leq \varepsilon+M^2\nu^2,
\end{align*}
when $N\geq \frac{C^2\log(2d_{\cal F}/\delta)}{2\varepsilon^2}$.

By applying the similar proof of Theorem \ref{thm:approx_err_finite_class}, note that the difference is that here we have an extra error term $M^2\nu^2$. So, for the finite function class case with \eqref{ineq:second_derivative_bound} and \eqref{ineq:cross_derivative_bound}, {when $\gamma = \Omega\left(\left(\frac{C^2\log(2d_{\mathcal{F}}/\delta)}{N C_\phi^4}\right)^{1/16}\right)$}, we have
\begin{align*}
    \mathbb{E}\left[\left\|\widehat{\bm{h}}''(\A\s)\right\|_2^2\right]= O\left(\frac{C_\phi}{\sigma_{\min}^2({\bm G})}  \left(\frac{C^2\log\left(\frac{2d_{\cal F}}{\delta}\right)}{N}+M^2\nu^2\right)^{1/4} \right),
\end{align*}
for any $\bm s$ such that $1-\gamma\geq s_{i}\geq \gamma>0$ for all $i\in[K]$.

\subsection{Neural Network Function Class}
For the case where $\mathcal{F}$ is the neural networks class defined in Assumption~\ref{as:nn}, similarly we have
\begin{align*}
    \frac{1}{N}\sum_{\ell=1}^N \left(1-\bm{1}^\top \widehat{\bm{f}}(\bm{x}_\ell)\right)^2\leq M^2\nu^2.
\end{align*}
Thus, by applying the Rademacher complexity-based generalization bound \cite[Theorem 3.3]{mohri2018foundations}, we have
\begin{align*}
    &\quad\mathbb{E}\left[\left\|{\bm G}\widehat{\bm{h}}''(\A\s)\right\|_2^2\right]= O\left( C_\phi \left(8MzB^2C_x (MzB^2C_x+1) \sqrt{\frac{R}{N}}\right.\right.\\
    &\quad\quad\quad\left.\left.+(MzB^2C_x+1)^2\sqrt{\frac{\log(1/\delta)}{2N}}+M^2\nu^2 \right)^{1/2} \right),
\end{align*}
when $\gamma = \Omega\left(\left(\frac{MzB^2C_x}{C_\phi}\right)^{1/4}\left(\sqrt{\frac{R}{N}}+\sqrt{\frac{\log(1/\delta)}{N}}\right)^{1/8}\right)$, which can be further simplified as (by inequality $(x+y)^{1/2}\leq x^{1/2}+y^{1/2}$ for $x,y>0$) \begin{align*}
    \mathbb{E}\left[\left\|\widehat{\bm{h}}''(\A\s)\right\|_2^2\right]= O\left( \frac{C_\phi{M}zB^2C_x\left(\sqrt{R}+\sqrt{\log(1/\delta)}\right)^{1/2}}{\sigma_{\min}^2({\bm G}) N^{1/4}}
    +\frac{{M\nu}}{\sigma_{\min}^2({\bm G})}  \right),
\end{align*}
for any $\bm s$ such that $1-\gamma\geq s_{i}\geq \gamma>0$ for all $i\in[K]$.

\section{Proof of Lemma~\ref{lem:hoeffding}}\label{appdx:hoeffding}
The proof follows the standard uniform convergence technique in statistical learning.
Let us define the following two terms:
\begin{subequations}
\begin{align}
    {\cal P}_N(\widehat{\bm f})&=\frac{1}{N}\sum_{\ell=1}^N(1-\bm 1^\T\widehat{\bm{f}}(\bm x_\ell))^2, \\
    {\cal P}_{\cal D}(\widehat{\bm f})& = \mathbb{E}[(1-\bm 1^\T\widehat{\bm{f}}(\bm x_\ell))^2 ].
\end{align}
\end{subequations}

By the Hoeffding's inequality (see Appendix~\ref{app:tools}), for any $\widehat{\bm f}$, we have
\begin{equation}
    {\sf Pr}\left[ \left|{\cal P}_{\cal D}(\widehat{\bm f})-{\cal P}_N(\widehat{\bm f})\right|>\varepsilon \right]\leq 2\exp(-2N\varepsilon^2/C^2).
\end{equation}

By the union bound, one can easily show that
\begin{align}
    {\sf Pr}\left[ \{\bm x_\ell\}_{\ell=1}^N: \exists \widehat{\bm f}: \left|{\cal P}_{\cal D}(\widehat{\bm f})-{\cal P}_N(\widehat{\bm f})\right|>\varepsilon \right]
    \leq 2d_{\cal F}\exp\left(\frac{-2N\varepsilon^2}{C^2}\right), \nonumber
\end{align}
Note that under our generative model and ${\cal F}\cup {\cal G}^{-1}\neq \emptyset$, there always exists a solution in ${\cal F}$ for \eqref{eq:sample}. Hence,
${\cal P}_N(\widehat{\bm f})=0$.

\section{Proof of Lemma~\ref{lemma:bounded_4th}}\label{appdx:bounded_4th}
It is readily seen that
\begin{align*}
    \frac{ \partial^4 \phi(\s)}{\partial s_i^4} =\sum_{m=1}^M (a_{m,i}-a_{m,K})^4 h^{(4)}_m \left(\bm{As}\right),~\forall i\in[K-1].
\end{align*}

Note that by rudimentary algebra, we have the following inequality:
\begin{align*}
    h^{(4)}&=\left(f\circ g\right)^{(4)}\\
    &=f^{(4)}(g)\cdot(g')^4+6f^{(3)}(g)\cdot(g')^2\cdot g''+3f''(g)\cdot(g'')^2
    +4f''(g)\cdot g'\cdot g^{(3)}+f'(g)\cdot g^{(4)}\\
    &\leq C_f C_g^4+6 C_f C_g^3+3C_f C_g^2+4C_f C_g^2+C_f C_g,
\end{align*}
where we have dropped the subscripts of $f,g$ and $h$ for notational simplicity.
By the definition of $C_\phi$, we have
\begin{align}
      \left|\frac{ \partial^4 \phi(\s)}{\partial s_i^4} \right|&\leq\left|\sum_{m=1}^M (a_{m,i}-a_{m,K})^4 h^{(4)}_m \left(\bm{As}\right)\right|
    \leq \left|\sum_{m=1}^M (2C_a)^4 h^{(4)}_m \left(\bm{As}\right)\right|\leq C_\phi.\label{eq:4thorderbound_1}
\end{align}

For the cross derivatives, the following holds:
\begin{align*}
   \frac{\partial^4\phi(\bm s)}{\partial s_i^3 \partial s_j}=\sum_{m=1}^M (a_{m,i}-a_{m,K})^3(a_{m,j}-a_{m,K}) h^{(4)}_m \left(\bm{As}\right),
\end{align*}
for all $i,j\in[K-1]$.
Using similar derivation as in \eqref{eq:4thorderbound_1}, one can attain the same bound for $\left|\frac{\partial^4\phi(\bm s)}{\partial s_i^3 \partial s_j}\right|$ and $\left|\frac{\partial^4\phi(\bm s)}{\partial s_i^2 \partial s_j^2}\right|$.

\section{Proof of Lemma~\ref{lem:rc_nn}}\label{appdx:rc_nn}
In this section, we show the Rademacher complexity of the function class that we use.
We first derive the Rademacher complexity of function class $\mathcal{F}$ which is defined in Assumption~\ref{as:nn}. 
By \cite[Theorem 43]{liang2016cs229t}, the Rademacher complexity of a bounded two-layer neural network $\mathcal{F}$ with $z$-Lipschitz activation is 
    \begin{align*}
        {\mathfrak{R}_N}(\mathcal{F})\leq 2zB^2C_x\sqrt{\frac{R}{N}}.
    \end{align*}
    
To further consider the Rademacher complexity of the loss function class defined in \eqref{eq:loss}, first note that the Rademacher complexity has the following property \cite{bartlett2002rademacher}: $$\mathfrak{R}_N\left(\mathcal{F}_1+\mathcal{F}_2\right)=\mathfrak{R}_N\left(\mathcal{F}_1\right)+\mathfrak{R}_N\left(\mathcal{F}_2\right),$$
where $f\in {\cal F}_1+{\cal F}_2$ means $f$ is a linear combination of functions from ${\cal F}_1$ and ${\cal F}_2$. Hence, the complexity of function $\sum_{m=1}^M f_m(\cdot)$ is bounded by   $M{\mathfrak{R}_N}(\mathcal{F})$.
    
Each function $f_m(\cdot)$ can be expressed as $f_m(x)=  \bm w_2^\T\bm \zeta(\bm w_1 x)$.
With the assumption that $\zeta(0)=0$, we have $f_m(0)=0$ and
\begin{align*}
    |f_m(x)-f_m(0)|&=  |\bm w_2^\T\bm \zeta(\bm w_1 x)-\bm w_2^\T \bm \zeta(\bm 0)|\\
    &\leq \|\bm w_2\|_2 \|\bm \zeta(\bm w_1 x)- \bm \zeta(\bm 0)\|_2\\
    &\leq zB\|\bm w_1 x-\bm 0\|_2\\
    &\leq zB^2C_x,\\
    \Longrightarrow |f_m(x)| & \leq  zB^2C_x, 
\end{align*}
where the second is by the Cauchy–Schwarz inequality, the third one is by the Lipschitz continuity of $\zeta(\cdot)$, and the last one is also by the Cauchy–Schwarz inequality.
Therefore, we claim that function $f_m(x_m)$ is bounded within $[-zB^2C_x,\ zB^2C_x]$.
        
Accordingly, $|1-\sum_{m=1}^M f_m(x_m)|$ can be bounded within $[0,MzB^2C_x+1]$. By the Lipschitz composition property of Rademacher complexity that ${\mathfrak{R}_N}(\phi\circ \mathcal{F})\leq L_\phi {\mathfrak{R}_N}( \mathcal{F})$ where $L_\phi$ denotes the Lipschitz constant of $\phi$. Here, the Lipschitz constant for the loss function is $L_\phi=2(MzB^2C_x+1)$. Thus we have
\begin{align*}
    {\mathfrak{R}_N}(\mathcal{H})\leq2(MzB^2C_x+1)M{\mathfrak{R}_N}(\mathcal{F}).
\end{align*}
        
By plugging in ${\mathfrak{R}_N}(\mathcal{F})$ we have
\begin{align*}
    {\mathfrak{R}_N}(\mathcal{H})&\leq2(MzB^2C_x+1)M{\mathfrak{R}_N}(\mathcal{F})\\
    &\leq 2M(MzB^2C_x+1)2zB^2C_x\sqrt{\frac{R}{N}}\\
    &=4 MzB^2C_x (MzB^2C_x+1) \sqrt{\frac{R}{N}}.
\end{align*}

\section{Estimation Error Bound for Second-Order Cross Derivatives}\label{appdx:cross_derivative}
First, we have the following Lemma for second-order cross derivatives.
\begin{Lemma}\label{lemma:cross_derivative}
    For the cross derivative of a continuous function $\psi(x,y)$ w.r.t. both of its arguments, we have the following numerical estimation
    \begin{align*}
        \frac{\partial^2\psi(x,y)}{\partial x \partial y}&=\frac{\psi(x+\Delta x,y+\Delta y)-\psi(x+\Delta x,y-\Delta y)}{4\Delta x \Delta y}
    -\frac{\psi(x-\Delta x,y+\Delta y)-\psi(x-\Delta x,y-\Delta y)}{4\Delta x \Delta y}\\
    & -\frac{\Delta x^2}{6}\frac{\partial^4 \psi(\xi'_{11}, \xi'_{21})}{\partial x^3\partial y}-\frac{\Delta y^2}{6}\frac{\partial^4 \psi(\xi'_{12}, \xi'_{22})}{\partial x\partial y^3}
    -\frac{\Delta x^3}{48\Delta y}\left(\frac{\partial^4 \psi(\xi'_{13},\xi'_{23})}{\partial x^4}-\frac{\partial^4 \psi(\xi'_{14},\xi'_{24})}{\partial x^4}\right)\\
    &-\frac{\Delta x \Delta y}{8}\left(\frac{\partial^4 \psi(\xi'_{15},\xi'_{25})}{\partial x^2\partial y^2}-\frac{\partial^4 \psi(\xi'_{16},\xi'_{26})}{\partial x^2\partial y^2}\right)
    -\frac{\Delta y^3}{48\Delta x}\left(\frac{\partial^4 \psi(\xi'_{17},\xi'_{27})}{\partial y^4}-\frac{\partial^4 \psi(\xi'_{18},\xi'_{28})}{\partial y^4}\right),
    \end{align*}
    where $\xi'_{1i}\in(x-\Delta x,x+\Delta x)$ and $\xi'_{2i}\in(y-\Delta y,y+\Delta y)$ for $i\in\{1,\cdots 8\}$.
\end{Lemma}

\noindent
{\bf Proof:}
For any two-dimensional functions whose higher-order derivatives exist, we have the following holds:
    \begin{align*}
        \psi(x+\Delta x,y+\Delta y)&=\psi(x,y)+\frac{\partial \psi(x,y)}{\partial x}\Delta x+\frac{\partial \psi(x,y)}{\partial y}\Delta y
        +\frac{\partial^2 \psi(x,y)}{\partial x^2}\frac{\Delta x^2}{2}+\frac{\partial^2 \psi(x,y)}{\partial x \partial y}\Delta x \Delta y\\
        &+\frac{\partial^2 \psi(x,y)}{\partial y^2}\frac{\Delta y^2}{2}
        +\frac{\partial^3 \psi(x,y)}{\partial x^3}\frac{\Delta x^3}{6}+\frac{\partial^3 \psi(x,y)}{\partial x^2\partial y}\frac{\Delta x^2 \Delta y}{2}+\frac{\partial^3 \psi(x,y)}{\partial x\partial y^2}\frac{\Delta x \Delta y^2 }{2}\\
        &+\frac{\partial^3 \psi(x,y)}{\partial y^3}\frac{\Delta y^3}{6}+\frac{\partial^4 \psi(\xi_{11},\xi_{21})}{\partial x^4}\frac{\Delta x^4}{24}
        +\frac{\partial^4 \psi(\xi_{11},\xi_{21})}{\partial x^3\partial y}\frac{\Delta x^3\Delta y}{6}\\
        &+\frac{\partial^4 \psi(\xi_{11},\xi_{21})}{\partial x^2\partial y^2}\frac{\Delta x^2\Delta y^2}{4}
        +\frac{\partial^4 \psi(\xi_{11},\xi_{21})}{\partial x\partial y^3}\frac{\Delta x\Delta y^3}{6}+\frac{\partial^4 \psi(\xi_{11},\xi_{21})}{\partial y^4}\frac{\Delta y^4}{24}.
\end{align*}

We have similar representations for $ \psi(x+\Delta x,y-\Delta y)$, $\psi(x-\Delta x,y+\Delta y)$, and $ \psi(x-\Delta x,y-\Delta y)$ following the basic rule of Taylor expansion (which are omitted for space).

By putting together, we have the following holds:
    \begin{align}
        &\quad\quad\psi(x+\Delta x,y+\Delta y)-\psi(x+\Delta x,y-\Delta y)
        -\psi(x-\Delta x,y+\Delta y)+\psi(x-\Delta x,y-\Delta y)\nonumber\\
        &=\frac{\partial^2 \psi(x,y)}{\partial x \partial y}4\Delta x \Delta y+\sum_{i=1}^4\frac{\partial^4 \psi(\xi_{1i},\xi_{2i})}{\partial x^3\partial y}\frac{\Delta x^3\Delta y}{6}
        +\sum_{i=1}^4\frac{\partial^4 \psi(\xi_{1i},\xi_{2i})}{\partial x\partial y^3}\frac{\Delta x\Delta y^3}{6}\nonumber\\
        &+\sum_{i=1}^4 (-1)^{i+1}\frac{\partial^4 \psi(\xi_{1i},\xi_{2i})}{\partial x^4}
        \frac{\Delta x^4}{24}
        +\sum_{i=1}^4(-1)^{i+1}\frac{\partial^4 \psi(\xi_{1i},\xi_{2i})}{\partial x^2\partial y^2}\frac{\Delta x^2\Delta y^2}{4}
        +\sum_{i=1}^4(-1)^{i+1}\frac{\partial^4 \psi(\xi_{1i},\xi_{2i})}{\partial y^4}\frac{\Delta y^4}{24}, \label{eq:cross_estimate}
    \end{align}
where $\xi_{11}\in(x,x+\Delta x)$, $\xi_{21}\in(y,y+\Delta y)$,  $\xi_{12}\in(x,x+\Delta x)$,   $\xi_{22}\in(y-\Delta y,y)$,  $\xi_{13}\in(x-\Delta x,x)$,  $\xi_{23}\in(y-\Delta y,y)$,  $\xi_{14}\in(x-\Delta x,x)$ and $\xi_{24}\in(y,y+\Delta y)$. 
    
    By the intermediate value theorem, there exists points $\xi'_{1i}\in(x-\Delta x,x+\Delta x)$ and $\xi'_{2i}\in(y-\Delta y,y+\Delta y)$ for $i\in\{1,\cdots,8\}$ such that 
    \begin{align*}
        \sum_{i=1}^4\frac{\partial^4 \psi(\xi_{1i},\xi_{2i})}{\partial x^3\partial y}=\frac{4\partial^4 \psi(\xi'_{11},\xi'_{21})}{\partial x^3\partial y},\quad
        \sum_{i=1}^4\frac{\partial^4 \psi(\xi_{1i},\xi_{2i})}{\partial x\partial y^3}=\frac{4\partial^4 \psi(\xi'_{12},\xi'_{22})}{\partial x\partial y^3},\\
    \end{align*}
    and
    \begin{align*}
        \sum_{i=1}^4(-1)^{i+1}\frac{\partial^4 \psi(\xi_{1i},\xi_{2i})}{\partial x^4}&=2\left(\frac{\partial^4 \psi(\xi'_{13},\xi'_{23})}{\partial x^4}-\frac{\partial^4 \psi(\xi'_{14},\xi'_{24})}{\partial x^4}\right),\\
        \sum_{i=1}^4(-1)^{i+1}\frac{\partial^4 \psi(\xi_{1i},\xi_{2i})}{\partial x^2\partial y^2}&=2\left(\frac{\partial^4 \psi(\xi'_{15},\xi'_{25})}{\partial x^2\partial y^2}-\frac{\partial^4 \psi(\xi'_{16},\xi'_{26})}{\partial x^2\partial y^2}\right),\\
        \sum_{i=1}^4(-1)^{i+1}\frac{\partial^4 \psi(\xi_{1i},\xi_{2i})}{\partial y^4}&=2\left(\frac{\partial^4 \psi(\xi'_{17},\xi'_{27})}{\partial y^4}-\frac{\partial^4 \psi(\xi'_{18},\xi'_{28})}{\partial y^4}\right).
    \end{align*}
    
    By combining the above and dividing Eq.~\eqref{eq:cross_estimate} by $4\Delta x\Delta y$ we have (for the right hand side)
    \begin{align*}
        &\frac{\partial^2 \psi(x,y)}{\partial x \partial y}+\frac{\partial^4 \psi(\xi'_{11},\xi'_{21})}{\partial x^3\partial y}\frac{\Delta x^2}{6}+\frac{\partial^4 \psi(\xi'_{12},\xi'_{22})}{\partial x\partial y^3}\frac{\Delta y^2}{6}
        +\left(\frac{\partial^4 \psi(\xi'_{13},\xi'_{23})}{\partial x^4}-\frac{\partial^4 \psi(\xi'_{14},\xi'_{24})}{\partial x^4}\right)\frac{\Delta x^3}{48\Delta y}\\
        &+\left(\frac{\partial^4 \psi(\xi'_{15},\xi'_{25})}{\partial x^2\partial y^2}-\frac{\partial^4 \psi(\xi'_{16},\xi'_{26})}{\partial x^2\partial y^2}\right)\frac{\Delta x \Delta y}{8}
        +\left(\frac{\partial^4 \psi(\xi'_{17},\xi'_{27})}{\partial y^4}-\frac{\partial^4 \psi(\xi'_{18},\xi'_{28})}{\partial y^4}\right)\frac{\Delta y^3}{48\Delta x},
    \end{align*}
    where the first term is what we aim to estimate, with the rest as error terms. \hfill $\square$

\bigskip

To further show the bound, we define:
\begin{align*}
    \Delta \bm{s}_{ij}^{++}&=[\bm 0,\ldots,+\Delta s_i,\ldots, \bm 0,\ldots,+\Delta s_j,\ldots, \bm 0, -\Delta s_i-\Delta s_j]^\top,\\
    \Delta \bm{s}_{ij}^{+-}&=[\bm 0,\ldots,+\Delta s_i,\ldots, \bm 0,\ldots,-\Delta s_j,\ldots, \bm 0, -\Delta s_i+\Delta s_j]^\top,\\
    \Delta \bm{s}_{ij}^{-+}&=[\bm 0,\ldots,-\Delta s_i,\ldots, \bm 0,\ldots,+\Delta s_j,\ldots, \bm 0, +\Delta s_i-\Delta s_j]^\top,\\
    \Delta \bm{s}_{ij}^{--}&=[\bm 0,\ldots,-\Delta s_i,\ldots, \bm 0,\ldots,-\Delta s_j,\ldots, \bm 0, +\Delta s_i+\Delta s_j]^\top,
\end{align*}
with $\Delta s_i>0$ and $\Delta s_j>0$ for any $i,j\in[K-1]$ with $i< j$ and 
\begin{align*}
\Delta s_i\in {\cal S}_i = [0,\min\{s_{i,\ell},1-s_{i,\ell}\}),\quad
\Delta s_j\in {\cal S}_j = [0,\min\{s_{j,\ell},1-s_{j,\ell}\}).
\end{align*}

Define $\s_{\widehat{\ell}}=\s_\ell + \Delta \s_{ij}^{++}$, $\s_{\widetilde{\ell}}=\s_\ell + \Delta \s_{ij}^{+-}$, $\s_{\overline{\ell}}=\s_\ell + \Delta \s_{ij}^{-+}$, and $\s_{{\ell}'}=\s_\ell + \Delta \s_{ij}^{--}$.
Then, we have
\begin{equation}\label{eq:derivative_case2}
\begin{aligned}
    \bm{1}^\top\widehat{\bm{h}}(\bm{A}(\bm{s}_\ell+\Delta \bm{s}_{ij}^{++}))&=1\pm \sqrt{\varepsilon_{\widehat{\ell}}},\quad
    \bm{1}^\top\widehat{\bm{h}}(\bm{A}(\bm{s}_\ell+\Delta \bm{s}_{ij}^{+-}))=1\pm \sqrt{\varepsilon_{\widetilde{\ell}}},\\
    \bm{1}^\top\widehat{\bm{h}}(\bm{A}(\bm{s}_\ell+\Delta \bm{s}_{ij}^{-+}))&=1\pm \sqrt{\varepsilon_{\overline{\ell}}},\quad
    \bm{1}^\top\widehat{\bm{h}}(\bm{A}(\bm{s}_\ell+\Delta \bm{s}_{ij}^{--}))=1\pm \sqrt{\varepsilon_{{\ell}'}}.
\end{aligned}
\end{equation}

For any continuous function $\psi(x,y)$ which has non-vanishing fourth-order partial derivatives, the second-order cross derivatives can be expressed using the following formula (see Lemma~\ref{lemma:cross_derivative}):
\begin{equation}\label{eq:derivative_cross}
\begin{aligned}
    \frac{\partial^2\psi(x,y)}{\partial x \partial y}&=\frac{\psi(x+\Delta x,y+\Delta y)-\psi(x+\Delta x,y-\Delta y)}{4\Delta x \Delta y}
    -\frac{\psi(x-\Delta x,y+\Delta y)-\psi(x-\Delta x,y-\Delta y)}{4\Delta x \Delta y}\nonumber\\
    & -\frac{\Delta x^2}{6}\frac{\partial^4 \psi(\xi_{11}, \xi_{21})}{\partial x^3\partial y}-\frac{\Delta y^2}{6}\frac{\partial^4 \psi(\xi_{12}, \xi_{22})}{\partial x\partial y^3}
    -\frac{\Delta x^3}{48\Delta y}\left(\frac{\partial^4 \psi(\xi_{13},\xi_{23})}{\partial x^4}-\frac{\partial^4 \psi(\xi_{14},\xi_{24})}{\partial x^4}\right)\\
    &-\frac{\Delta x \Delta y}{8}\left(\frac{\partial^4 \psi(\xi_{15},\xi_{25})}{\partial x^2\partial y^2}-\frac{\partial^4 \psi(\xi_{16},\xi_{26})}{\partial x^2\partial y^2}\right)
    -\frac{\Delta y^3}{48\Delta x}\left(\frac{\partial^4 \psi(\xi_{17},\xi_{27})}{\partial y^4}-\frac{\partial^4 \psi(\xi_{18},\xi_{28})}{\partial y^4}\right),
\end{aligned}
\end{equation}
where $\xi_{1i}\in(x-\Delta x,x+\Delta x)$ and $\xi_{2i}\in(y-\Delta y,y+\Delta y)$ for $i\in\{1,\cdots,8\}$.

Using the above formula, one can express $\frac{\partial^2\phi(\s)}{\partial s_i\partial s_j}$ as follows:
\begin{align*}
    \frac{\partial^2\phi(\s)}{\partial s_i\partial s_j}
    &=\frac{\pm\sqrt{\varepsilon_{\widehat{\ell}}}\mp \sqrt{\varepsilon_{\widetilde{\ell}}}\mp \sqrt{\varepsilon_{\overline{\ell}}}\pm \sqrt{\varepsilon_{\ell'}}}{4\Delta s_i \Delta s_j}
    -\frac{\Delta s_i^2}{6}\frac{\partial^4 \phi(\bm \xi^{(1)}_{ij})}{\partial s_i^3\partial s_j}-\frac{\Delta s_j^2}{6}\frac{\partial^4 \phi(\bm \xi^{(2)}_{ij})}{\partial s_i\partial s_j^3}
    -\frac{\Delta s_i^3}{48\Delta s_j}\left(\frac{\partial^4 \phi(\bm \xi^{(3)}_{ij})}{\partial s_i^4}-\frac{\partial^4 \phi(\bm \xi^{(4)}_{ij})}{\partial s_i^4}\right)\\
    &-\frac{\Delta s_i \Delta s_j}{8}\left(\frac{\partial^4 \phi(\bm \xi^{(5)}_{ij})}{\partial s_i^2\partial s_j^2}-\frac{\partial^4 \phi(\bm \xi^{(6)}_{ij})}{\partial s_i^2\partial s_j^2}\right)
    -\frac{\Delta s_j^3}{48\Delta s_i}\left(\frac{\partial^4 \phi(\bm \xi^{(7)}_{ij})}{\partial s_j^4}-\frac{\partial^4 \phi(\bm \xi^{(8)}_{ij})}{\partial s_j^4}\right),
\end{align*}
where $\bm \xi^{(k)}_{ij}$'s are vectors satisfying
\[ \bm \xi^{(k)}_{ij} = \theta^{(k)} \s_{\widehat{\ell}}+(1-\theta^{(k)})\s_{\ell'}\in {\rm int}\Delta, k\in\{1,\cdots,8\}, \]
where $\theta^{(k)}\in(0,1)$, is a vector such that $[\bm \xi^{(k)}_{ij}]_i\in (s_{i,\ell}-\Delta s_i,s_{i,\ell}+\Delta s_i)$ and $[\bm \xi^{(k)}_{ij}]_j\in (s_{j,\ell}-\Delta s_j,s_{j,\ell}+\Delta s_j)$.

Therefore, the following holds:
\begin{align*}
    \left| \frac{\partial^2\phi(\s)}{\partial s_i\partial s_j}\right|&\leq
     \frac{\sqrt{\varepsilon_{\widehat{\ell}}}+ \sqrt{\varepsilon_{\widetilde{\ell}}}+ \sqrt{\varepsilon_{\overline{\ell}}}+ \sqrt{\varepsilon_{\ell'}}}{4\Delta s_i \Delta s_j}
     +\frac{\Delta s_i^2}{6}\left|\frac{\partial^4 \phi(\bm \xi^{(1)}_{ij})}{\partial s_i^3\partial s_j}\right|+\frac{\Delta s_j^2}{6}\left|\frac{\partial^4 \phi(\bm \xi^{(2)}_{ij})}{\partial s_i\partial s_j^3}\right|\\
     &\quad+\frac{\Delta s_i^3}{48\Delta s_j}\left(\left|\frac{\partial^4 \phi(\bm \xi^{(3)}_{ij})}{\partial s_i^4}\right|+\left|\frac{\partial^4 \phi(\bm \xi^{(4)}_{ij})}{\partial s_i^4}\right|\right)
    +\frac{\Delta s_i \Delta s_j}{8}\left(\left|\frac{\partial^4 \phi(\bm \xi^{(5)}_{ij})}{\partial s_i^2\partial s_j^2}\right|+\left|\frac{\partial^4 \phi(\bm \xi^{(6)}_{ij})}{\partial s_i^2\partial s_j^2}\right|\right)\\
    &\quad+\frac{\Delta s_j^3}{48\Delta s_i}\left(\left|\frac{\partial^4 \phi(\bm \xi^{(7)}_{ij})}{\partial s_j^4}\right|+\left|\frac{\partial^4 \phi(\bm \xi^{(8)}_{ij})}{\partial s_j^4}\right|\right).
\end{align*}

By taking expectation and using Jensen's inequality,
\begin{align*}
    &\left|\mathbb{E}\left[\sum_{m=1}^M (a_{m,i}-a_{m,K})(a_{m,j}-a_{m,K})\widehat{h}''_m(\bm{As}_\ell)\right]\right|
    \leq\frac{\sqrt{\varepsilon}}{\Delta s_i\Delta s_j}+\frac{\Delta s_i^2}{6}\left|\frac{\partial^4 \phi(\bm \xi^{(1)}_{ij})}{\partial s_i^3\partial s_j}\right|+\frac{\Delta s_j^2}{6}\left|\frac{\partial^4 \phi(\bm \xi^{(2)}_{ij})}{\partial s_i\partial s_j^3}\right|\\
    &\quad +\frac{\Delta s_i^3}{48\Delta s_j}\left(\left|\frac{\partial^4 \phi(\bm \xi^{(3)}_{ij})}{\partial s_i^4}\right|+\left|\frac{\partial^4 \phi(\bm \xi^{(4)}_{ij})}{\partial s_i^4}\right|\right)
    +\frac{\Delta s_i \Delta s_j}{8}\left(\left|\frac{\partial^4 \phi(\bm \xi^{(5)}_{ij})}{\partial s_i^2\partial s_j^2}\right|+\left|\frac{\partial^4 \phi(\bm \xi^{(6)}_{ij})}{\partial s_i^2\partial s_j^2}\right|\right)\\
    &\quad +\frac{\Delta s_j^3}{48\Delta s_i}\left(\left|\frac{\partial^4 \phi(\bm \xi^{(7)}_{ij})}{\partial s_j^4}\right|+\left|\frac{\partial^4 \phi(\bm \xi^{(8)}_{ij})}{\partial s_j^4}\right|\right),
\end{align*}
where $\varepsilon=\varepsilon(N,\delta)$ which will be specified later.  

We are interested in finding the optimal upper bound 
\begin{align*}
    \inf_{\Delta s_i,\Delta s_j} &\frac{\sqrt{\varepsilon_0}}{\Delta s_i\Delta s_j}+\frac{\Delta s_i^2}{6}\left|\frac{\partial^4 \phi(\bm \xi^{(1)}_{ij})}{\partial s_i^3\partial s_j}\right|+\frac{\Delta s_j^2}{6}\left|\frac{\partial^4 \phi(\bm{\xi}^{(2)}_{ij})}{\partial s_i\partial s_j^3}\right|
    +\frac{\Delta s_i^3}{48\Delta s_j}\left(\left|\frac{\partial^4 \phi(\bm \xi^{(3)}_{ij})}{\partial s_i^4}\right|+\left|\frac{\partial^4 \phi(\bm \xi^{(4)}_{ij})}{\partial s_i^4}\right|\right)\\
    &\quad +\frac{\Delta s_i \Delta s_j}{8}\left(\left|\frac{\partial^4 \phi(\bm \xi^{(5)}_{ij})}{\partial s_i^2\partial s_j^2}\right|+\left|\frac{\partial^4 \phi(\bm \xi^{(6)}_{ij})}{\partial s_i^2\partial s_j^2}\right|\right)
    +\frac{\Delta s_j^3}{48\Delta s_i}\left(\left|\frac{\partial^4 \phi(\bm \xi^{(7)}_{ij})}{\partial s_j^4}\right|+\left|\frac{\partial^4 \phi(\bm \xi^{(8)}_{ij})}{\partial s_j^4}\right|\right).
\end{align*}

Without loss of generality, we assume that $\Delta s=\Delta s_i=\Delta s_j$, with its feasible domain
\begin{align*}
    \Delta s\in [0,\min\{s_{i,\ell},s_{i,\ell},1-s_{i,\ell},1-s_{j,\ell}\}),
\end{align*}
and we have the following
\begin{align*}
    \inf_{\Delta s} &\frac{\sqrt{\varepsilon_0}}{\Delta s^2}+\frac{\Delta s^2}{6}\left(\left|\frac{\partial^4 \phi(\bm \xi^{(1)}_{ij})}{\partial s_i^3\partial s_j}\right|+\left|\frac{\partial^4 \phi(\bm \xi^{(2)}_{ij})}{\partial s_i\partial s_j^3}\right|\right)
    +\frac{\Delta s^2}{48}\left(\left|\frac{\partial^4 \phi(\bm \xi^{(3)}_{ij})}{\partial s_i^4}\right|+\left|\frac{\partial^4 \phi(\bm \xi^{(4)}_{ij})}{\partial s_i^4}\right|\right)\\
    &\quad +\frac{\Delta s^2}{8}\left(\left|\frac{\partial^4 \phi(\bm \xi^{(5)}_{ij})}{\partial s_i^2\partial s_j^2}\right|+\left|\frac{\partial^4 \phi(\bm \xi^{(6)}_{ij})}{\partial s_i^2\partial s_j^2}\right|\right)
    +\frac{\Delta s^2}{48}\left(\left|\frac{\partial^4 \phi(\bm \xi^{(7)}_{ij})}{\partial s_j^4}\right|+\left|\frac{\partial^4 \phi(\bm \xi^{(8)}_{ij})}{\partial s_j^4}\right|\right).
\end{align*}

Define $\tau$ as
\begin{align*}
    \tau&:=8\left(\left|\frac{\partial^4 \phi(\bm \xi^{(1)}_{ij})}{\partial s_i^3\partial s_j}\right|+\left|\frac{\partial^4 \phi(\bm \xi^{(2)}_{ij})}{\partial s_i\partial s_j^3}\right|\right)
    +\left(\left|\frac{\partial^4 \phi(\bm \xi^{(3)}_{ij})}{\partial s_i^4}\right|+\left|\frac{\partial^4 \phi(\bm \xi^{(4)}_{ij})}{\partial s_i^4}\right|\right)\\
    &\quad+6\left(\left|\frac{\partial^4 \phi(\bm \xi^{(5)}_{ij})}{\partial s_i^2\partial s_j^2}\right|+\left|\frac{\partial^4 \phi(\bm \xi^{(6)}_{ij})}{\partial s_i^2\partial s_j^2}\right|\right)
    +\left(\left|\frac{\partial^4 \phi(\bm \xi^{(7)}_{ij})}{\partial s_j^4}\right|+\left|\frac{\partial^4 \phi(\bm \xi^{(8)}_{ij})}{\partial s_j^4}\right|\right).
\end{align*}

So the minimum is obtained at
\begin{align*}
    \Delta s^* \in \left\{ \left(\frac{48\sqrt{\varepsilon_0}}{\tau}\right)^{1/4},\ \min\{s_{i,\ell},s_{j,\ell},1-s_{i,\ell},1-s_{j,\ell}\} \right\},
\end{align*}
which gives the following minimum
\begin{align}
    \inf_{\Delta s} \frac{\sqrt{\varepsilon_0}}{\Delta s^2}+\frac{\tau\Delta s^2}{48}
    \leq\min\left\{\frac{\sqrt{3}\varepsilon_0^{1/4}\sqrt{\tau}}{6},\frac{\sqrt{\varepsilon_0}}{\kappa^2}+\frac{\kappa^2\tau}{6}\right\},\label{ineq:cross_bound}
\end{align}
where $\kappa=\min\{s_{i,\ell},s_{j,\ell},1-s_{i,\ell},1-s_{j,\ell}\}$. Similarly we have
\[ \kappa \geq \gamma. \]
Hence, if 
\[ \left(\frac{48\sqrt{\varepsilon_0}}{\tau}\right)^{1/4} \leq \gamma, \]
the bound can be further simplified. If we have
\begin{align*}
    \left(\frac{48\sqrt{\varepsilon_0}}{16C_\phi}\right)^{1/4} \leq \gamma,
\end{align*}
then \eqref{ineq:cross_bound} can be bounded by
\begin{align*}
    \left|\mathbb{E}\left[\sum_{m=1}^M (a_{m,i}-a_{m,K})(a_{m,j}-a_{m,K})\widehat{h}''_m(\bm{As}_\ell)\right]\right|
    \leq \frac{\sqrt{3C_\phi}\varepsilon_0^{1/4}}{6}.
\end{align*}

Given that $N$ is fixed, one may pick $\varepsilon = \sqrt{ \frac{C^2\log(2d_{\cal F}/\delta)}{2N}}$, which gives the conclusion that
if $$\gamma \geq \left(\frac{3}{C_\phi}\right)^{1/4} \left(\frac{C^2\log\left(\frac{2d_{\cal F}}{\delta}\right)}{2N}\right)^{1/16},$$ we have
\begin{equation}\label{ineq:cross_derivative_bound2}
\begin{aligned}
    \left|\mathbb{E}\left[\sum_{m=1}^M (a_{m,i}-a_{m,K})(a_{m,j}-a_{m,K})\widehat{h}''_m(\bm{As}_\ell)\right]\right|
    \leq \frac{\sqrt{3 C_\phi}\left(\frac{C^2\log\left(\frac{2d_{\cal F}}{\delta}\right)}{2N}\right)^{1/8}}{6}.
\end{aligned}
\end{equation}

\section{Proof of Proposition~\ref{prop:samplekkt}}\label{appdx:samplekkt}
    First, define ${\cal C}_{\bm{\theta}^\ast}(\bm x_\ell)=\bm 1^\T\widehat{\bm{f}}(\bm x_\ell)-1$ where $\widehat{\bm{f}}$ is a solution of \eqref{eq:auto}.
    For neural network $\mathcal{F}$ defined in Assumption~\ref{as:nn}, by selecting $\epsilon$ as in Eq.~\eqref{eq:eps_nn}, when 
    \begin{align*}
        N= \Omega\left(\frac{M^4z^4B^8C^4_x\left(\sqrt{R}+\sqrt{{\log(1/\delta)}}\right)^2}{\epsilon^2}\right),
    \end{align*}
    we have $\left| \mathbb{E}[{\cal C}_{\bm \theta^\ast}(\x)] \right|^2 \leq \varepsilon$. 
    
    To bound $v_{\infty,\bm \theta^\ast}$, we need first to bound the Rademacher complexity of function $\bm{q}\circ\bm{f}$ with ${q}_m, f_m\in\mathcal{F}$ which is defined in Assumption~\ref{as:nn}. 
    The Rademacher complexity of the function class $\mathcal{K}=\{q\circ f(x)\ |\ q, f\in\mathcal{F}\}$ could be bounded by
    \begin{align*}
        {\mathfrak{R}_N}(\mathcal{K})&\leq zB^2{\mathfrak{R}_N}(\mathcal{F})\leq 2z^2B^4C_x\sqrt{\frac{R}{N}},
    \end{align*}
    where the first inequality is by the function composition property of Rademacher complexity \cite{bartlett2002rademacher} and any function in $\mathcal{F}$ is $zB^2$-Lipschitz continuous.
    
    Next, the Rademacher complexity of the function class ${\cal K}'$ {such that $\mathcal{K}'=\{q\circ f(x)-x\ |\ q, f\in\mathcal{F}\}$, where $\mathcal{F}$ is defined in Assumption~\ref{as:nn},} is bounded by
    
    \begin{align*}
        {\mathfrak{R}_N}(\mathcal{K}')={\mathfrak{R}_N}(\mathcal{K})+0\leq 2z^2B^4C_x\sqrt{\frac{R}{N}},
    \end{align*}
    due to the linearity of Rademacher complexity \cite{bartlett2002rademacher} and the fact that the function $-x$ is a singleton function, whose Rademacher complexity is 0.
    
    Following similar proof of Lemma~\ref{lem:rc_nn}, the Rademacher complexity for the {function class $$\mathcal{K}''=\left\{\left\|\bm q\left( \bm f(\x)\right) -\x \right\|_2^2\ |\ q_m, f_m\in\mathcal{F},\ \forall m=[M]\right\}$$ can be derived as follows}. First, the function $|q_m\circ f_m(x_m)-x_m|$ is upper bounded by $z^2B^4C_x+C_x$.
    Thus we have
    \begin{align*}
        {\mathfrak{R}_N}(\mathcal{K}'')&\leq2M(z^2B^4C_x+C_x){\mathfrak{R}_N}(\mathcal{K}')
        \leq 4Mz^2B^4C_x(z^2B^4C_x+C_x)\sqrt{\frac{R}{N}}.
    \end{align*}
    
    Consequently, 
    \begin{align*}
        \mathbb{E}&\left[\left\|\bm{q}(\bm{f}(\bm{x}))-\bm{x}\right\|_2^2\right]\leq \frac{1}{N}\sum_{\ell=1}^N \|\bm{q}(\bm{f}(\bm{x}_\ell))-\bm{x}_\ell\|_2^2
        +2\mathfrak{R}_N(\mathcal{K}'') +M(z^2B^4C_x+C_x)^2\sqrt{\frac{\log(1/\delta)}{2N}}.
    \end{align*}
    
    Let $\epsilon=2\mathfrak{R}_N(\mathcal{K}'')+M(z^2B^4C_x+C_x)^2\sqrt{\frac{\log(1/\delta)}{2N}}$, we have
    \begin{align*}
        N= \Omega\left(\frac{M^2z^8B^{16}C_x^4\left(\sqrt{R}+\sqrt{\log(1/\delta)}\right)^2}{\epsilon^2}\right).
    \end{align*}

    For the finite function class, by assuming that $\left\|\bm q\left( \bm f(\x)\right) -\x \right\|_2^2\leq C_r$, we have \eqref{eq:sample_kkt} hold when $N\geq \max\left\{\frac{C^2\log(2d_{\cal F}/\delta)}{2\varepsilon^2}, \frac{C_r^2\log(2d_{\cal F}/\delta)}{2\varepsilon^2}\right\}$.

\section{Analytical Tools}\label{app:tools}
In this section, we introduce two analytical tools used in our proof, namely, the Hoeffding's inequality and the Rademacher complexity---depending on what kind of nonlinear function $\bm f\in{\cal F}$ that is used in the learning formulation \eqref{eq:sample}. The Hoeffding's concentration theorem is as follows \cite{shalev2014understanding}:
\begin{Theorem} 
    Let $Z_1,\cdots,Z_N$ be sequence of i.i.d. random variables. Assume that $\mathbb{E}[Z_i]=\mu$ and $\mathbb{P}[a\leq Z_i \leq b]=1$ for all $i$. Then for any $\epsilon>0$
    \begin{align*}
        \mathbb{P}\left[\left|\frac{1}{N}\sum_{i=1}^N Z_i-\mu\right|>\epsilon\right]\leq 2 \exp\left(-2N\epsilon^2/(b-a)^2\right).
    \end{align*}
\end{Theorem}
This will be used when we consider $f_m\in{\cal F}$ where $|{\cal F}|<\infty$.

We will also consider function classes where $|{\cal F}|=\infty$, in particular, neural networks. For this class, we will leverage the Rademacher complexity.
Let $\mathcal{G}$ be a family of functions $\bm g: \mathcal{X}\in\mathbb{R}^{M}\rightarrow y\in\mathbb{R}$ where $y\in[a, b]$. Denote $\{\bm{x}_\ell\}_{\ell=1}^N$ a fixed i.i.d. sample of size $N$ sampled from ${\cal X}$ following a certain distribution ${\cal D}$. Then, the {empirical Rademacher complexity \cite{bartlett2002rademacher}} of function class $\mathcal{G}$ with respect to $\{\bm{x}_\ell\}_{\ell=1}^N$ is defined as
 \begin{align*}
            \widehat{\mathfrak{R}}_{\X}(\mathcal{G})=\mathbb{E}_{\bm \sigma}\left[ \sup_{\bm{g}\in \mathcal{G}}\frac{1}{N}\sum_{\ell=1}^N \sigma_\ell\bm{g}(\bm{x}_\ell) \right],
\end{align*}
where $\X=[\x_1,\ldots,\x_N]$, $\sigma_i$ is independent uniform random variables taking $\{-1,+1\}$. Intuitively, for more complex families $\mathcal{G}$, the vectors $\bm{g}(\bm{x}_1),\ldots,\bm{g}(\bm{x}_N)$ are less similar to each other, and thus $\widehat{\mathfrak{R}}_{\bm X}(\mathcal{G})$ is expected to be larger---since the chance that $+\bm g(\x_\ell)$ and $-\bm g(\x_{\ell'})$ for $\ell\neq \ell'$ cancel each other is smaller.

The {Rademacher complexity} of $\mathcal{G}$ is the expectation of the empirical Rademacher complexity over all sample sets of size $N$ following the same distribution:
\begin{align*}
        {\mathfrak{R}_N}(\mathcal{G})=\mathbb{E}_{\X\sim{\cal D}^N}\left[\widehat{\mathfrak{R}}_{\X}(\mathcal{G}) \right],
\end{align*}
where ${\cal D}^N$ denotes the joint distribution of $N$ samples.

{
\section{Additional Experiment}\label{app:time_experiment}
In Fig.~\ref{fig:time_vs_M} shows the runtime of the proposed method is plotted as $M$ increases. In the experiment, we generate $5,000$ samples and the nonlinear functions for each dimension are randomly selected from $g_m(x)=\alpha\cdot\text{sigmoid}(x)+\beta x$ or $g_m(x)=\alpha\cdot\text{tanh}(x)+\beta x$ with $\alpha$ and $\beta$ drawn from the normal distribution and uniform distribution $[-0.5,0.5]$, respectively. 
We use a one-hidden-layer network with $32$, $64$ and $128$ neurons to model each dimension's $f_m(\cdot)$. We run the following settings with $(K,M)$ being $(5,5)$, $(5,10)$, $(10,20)$ and $(20,30)$, respectively.

One can see that as $M$ becomes larger, the runtime to reach $\frac{1}{N}\sum_{\ell=1}^N |\bm 1^\T\bm f(\bm x_\ell)-1|^2<10^{-5}$
increases rapidly when $R=256$. The time increase is more moderate when $R=128$ and $R=64$. This shows a tradeoff between the expressiveness of the employed neural network (i.e., larger $R$ means that the corresponding neural network is more expressive) and the computational cost. How to better balance these two aspects is a meaningful future direction.

\begin{figure}[ht]
    \centering
    {
    \subfigure{\includegraphics[width=0.55\linewidth]{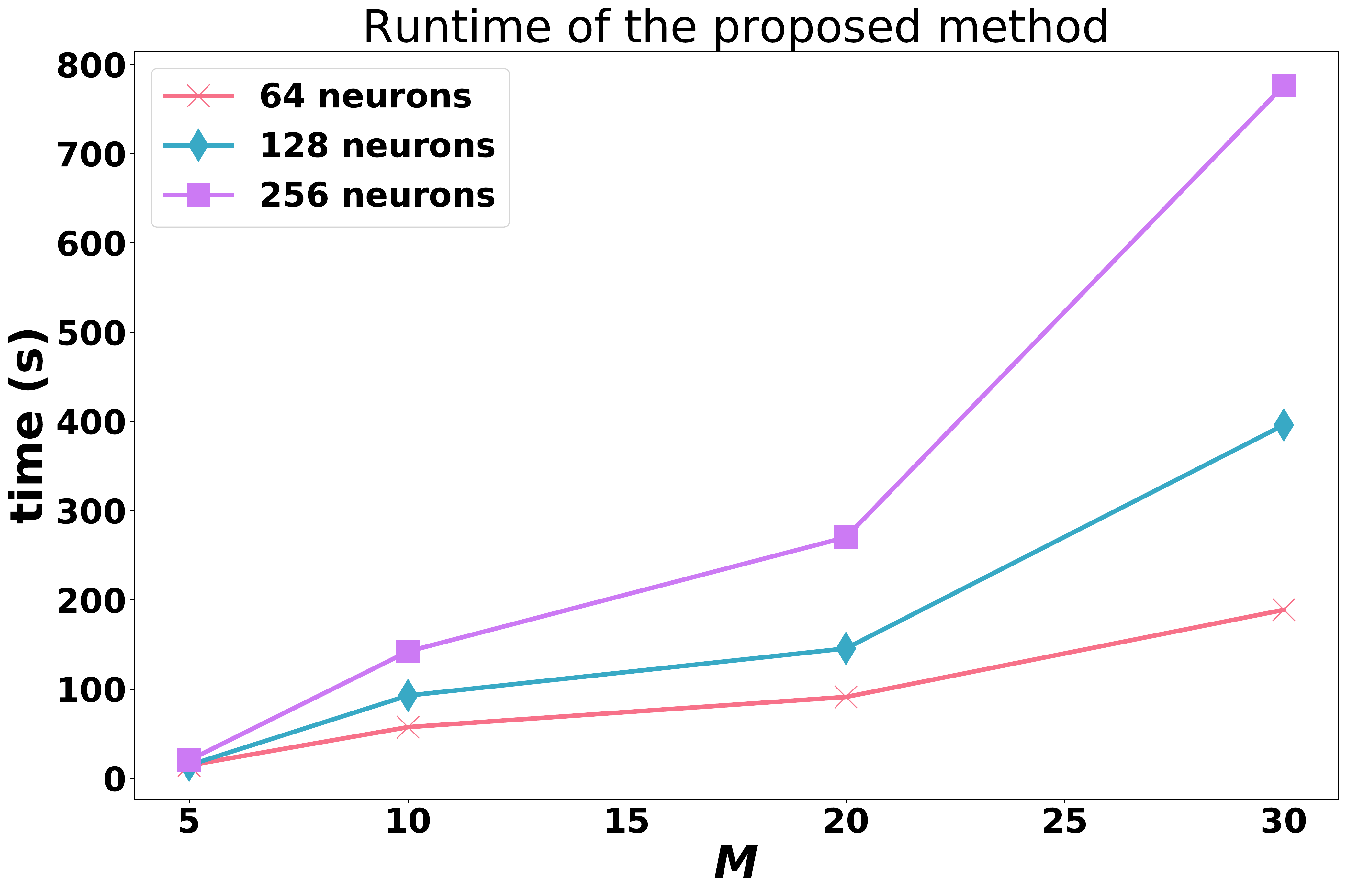}}
    \caption{Runtime of the proposed method as $M$ increases.}\label{fig:time_vs_M}
    }
\end{figure}

}

\putbib[bu2]
\end{bibunit}

\end{document}